\documentclass[a4paper]{article}
\usepackage[margin=1in]{geometry}
\usepackage[english]{babel}
\usepackage[utf8]{inputenc}
\usepackage{amsmath}
\usepackage{graphicx}
\usepackage[colorinlistoftodos]{todonotes}
\usepackage{covington}
\usepackage{booktabs}
\usepackage{pgf, tikz}
\usepackage{pgfplots}
\usepackage{mathdots}
\usepackage{mathtools}
\pgfplotsset{compat=1.12}
\usetikzlibrary{arrows}
\usepackage{braket}
\usepackage[normalem]{ulem}  
\usepackage{colortbl}
\usepackage{multicol}
\usepackage{multirow}
\usepackage{blkarray} 
\usepackage{tipa}
\usepackage{natbib}
\usepackage{pstricks,pst-vue3d,pst-3dplot}

\bibpunct{(}{)}{;}{a}{}{,}

\newcommand{\Blue}[1]{\color{blue}{#1}}
\newcommand{\Red}[1]{\color{red}{#1}}
\newcommand{\Green}[1]{\color{green}{#1}}

\newcommand{\scc}[1]{\textsc{#1}}
\newcommand{\nul}{\ensuremath{\emptyset}}

\newcommand{\BB}{\ensuremath{\mathcal{B}{}}}

\newcommand{\PHI}{\ensuremath{\Phi{}}}

\newcommand{\Bi}{\textipa{1}} 
\newcommand{\Feature}[1]{\textsc{#1}}
\newcommand{\FV}[1]{\textit{\Blue{#1}}}
\newcommand{\Gl}[1]{`\textsf{#1}'}

\usepackage[framemethod=tikz]{mdframed}
\definecolor{box_color}{rgb}{0.122, 0.435, 0.698}
\newmdenv[innerlinewidth=0.5pt, roundcorner=4pt,linecolor=box_color,innerleftmargin=6pt,
innerrightmargin=6pt,innertopmargin=6pt,innerbottommargin=6pt]{mybox}

\title{Geometrical morphology}

\usepackage{authblk}
\author[1]{John Goldsmith}
\author[2]{Eric Rosen}
\affil[1]{University of Chicago\\goldsmith@uchicago.edu}
\affil[2]{University of British Columbia\\errosen@mail.ubc.ca}
\date{\today}

\begin{document}
\maketitle

\begin{abstract}
We explore inflectional morphology as an example of the relationship of the discrete and the continuous in linguistics. The grammar {\em requests} a form of a lexeme by specifying a set of feature values, which corresponds to a corner $\ket{M}$ of a hypercube in feature value space. The morphology {\em responds} to that request by providing a morpheme, or a set of morphemes, whose vector sum is geometrically closest to the corner $M$. In short, the chosen morpheme $\hat{\bra{\mu}}$ is the morpheme which maximizes the inner product of $\bra{\mu}$ and $\ket{M}$.

\end{abstract}
\tableofcontents

\section*{Preface}
This is a working paper, describing work that the two of us have done over the course of 2016 and the beginning of 2017, developing in some detail an analysis of some aspects of inflectional morphology.\footnote{This is version 0.2. A slightly earlier version of this paper appeared as technical report TR-2017-02 of the Department of Computer Science at the University of Chicago.} In most respects this work continues the development of work in this area over the past fifty or more years, but in one or two important ways our approach is different. We believe that a  {\em geometric} interpretation of inflectional morphology is possible, and that it simplifies our understanding of the formal aspects of the syntax-morphology interface. It simplifies our understanding in that it renders unnecessary a large part of the derivational mechanics that have been part and parcel of most formalisms in this area. Readers for whom the mathematics below is unfamiliar are unlikely to agree with us that this is a simplification, but we will do our best to show that the mathematics and the geometry that we exploit is quite helpful in expressing our ideas about how languages work. 

Here is a list of some central ideas, to get us started:

\begin{enumerate}
\item All linguists know that inflectional paradigms in natural language morphology are natural objects to think of as arrays in as few as 2 to as many as 10 or so dimensions.  Each dimension corresponds to a linguistic feature, and features have a small number of possible values. The feature \Feature{Tense} may have past/present as values, \Feature{person} may have \FV{1st/2nd/3rd} as values. But we have found that this is not the organization that is most useful for us. 

\item It is sensible to view the dimensions of an inflectional paradigm as being built up out of a small set of privative features (features that take on the value \FV{yes} or \FV{no}, or $0$/$1$). These privative features are the feature values mentioned just above, like \FV{1st person}, or \FV{past tense}. Those features lead to feature-value space, of somewhat larger dimensionality than the arrays that linguists fashion for paradigms. 

\item The set of morphosyntactic feature specifications that is specified by a grammar for a particular position in a sentence can be modeled as corners in feature-value space (i.e., points whose coordinates are all 0s and 1s). For example, a 3rd person singular present verb corresponds to a particular corner of a hypercube, whose coordinates are 1 for the dimensions \FV{3rd-person}, \FV{present}, and \FV{singular}, but 0 for the four other dimensions. Not all corners in that space are meaningful, however: a point which takes the value 1 for both past tense and present tense will not be meaningful. 

\item The morphemes that realize inflectional morphology can be analyzed as vectors in feature value space.

\item When we do so, we can say that the correct morpheme for any morphosyntactic feature specification (a {\em corner position} of a hypercube) is the morpheme that is {\em geometrically closest to it.}

\item When a word is composed of several morphemes, the proper choice of morphemes is the set of morphemes (each taken to be a vector) whose vector sum is geometrically closest to the given morphosyntactic feature specification. 

\item Providing a learning mechanism by which the language-specific information about the grammar can be deduced or inferred from observed data is an essential aspect of a linguistic theory. Linguists' concerns about restricting the languages covered by linguistic theory are actually displaced (and misplaced) concerns about learnability.

\item Inflection patterns are sets of vectors in feature value space. Two related inflectional patterns are related to each by a rotation (that is, by a linear transformation that preserves inner product).

\item There is a (for us, surprising) similarity between the way this leads us to talk about morphology and the way in which quantum mechanics is described.   And as a measurement in physics corresponds to a morpheme realization, we have no reason to view that as a probabilistic operation; instead, a prepared system is realized by the morpheme that is closest to it, not probabilistically by all morphemes onto which it could project.

\end{enumerate}

\section{Introduction} \label{sec:intro}

In this paper we explore a {\em geometric} way of understanding inflectional morphology.\footnote{This paper develops a proposal lightly sketched in \citet[95-113]{Goldsmith:1994}. \citet{Stump:2016} is a good presentation of our starting point; we find his arguments clarifying the role of paradigms in morphology convincing, and if there is a current framework into which we imagine our work to be integrated, it is a framework such as the one that Stump develops there.  
We are grateful to discussions that have led to this work with a number of colleagues, including Stephen Fitz, Jackson Lee, Matt Goldrick, Doug Pulleyblank, Bryan Gick, Gunnar Hansson, Fred Chong, Risi Kondor, Brandon Rhodes, and Karlos Arregi.} The guiding idea is that many aspects of morphology can best be understood by thinking of features, morphemes, paradigms, and many other notions in morphology, through the lens of geometry. In a word: we can {\em visualize} morphology if properly conceived. At least we can if we can visualize spaces with quite a few dimensions to them. This geometrization of linguistic questions is similar in some regards to the advances achieved in phonology  from better understanding the geometrical structure of phonological representations, and it is also motivated by the notion that grammatical explanation can be best achieved by showing that the correct forms generated by a grammar are those that maximize (or minimize) a (largely continuous) function defined over a high-dimensional space; they can therefore be understood in most cases as seeking a representation that maximizes or minimizes a particular function on representations. 


One of the first characteristics of this approach is that we formalize the morphosyntactic feature specifications for a given inflected form not as a feature bundle, but as a point in a vector space. We will look at several different vector spaces, of which the two most important are paradigm space, which is a slight variant on our traditional understanding of paradigms; feature-value space, which takes paradigm space apart into a larger space whose dimensions correspond to feature values; and morpheme space, in which the basis vectors are the morphemes (something that will become clear in what follows). Most of our work will take place in feature-value space. 

In the cases we are interested in, the morphology serves as an interface between the morphosyntax and the morphological spelling-out. The morphosyntax specifies a corner point on a hypercube, and the morphology choses a vector (or, later, a set of vectors) that is as close to that corner point as possible.

\subsection{Paradigm space}

As we just observed, linguists often think about an inflectional paradigm as a multidimensional array, where each dimension takes on anywhere from 2 to 20 or so values (though it is uncommon for there to be more than 6 or so values for any given feature). One dimension might be number, with values singular and plural, and another might be person, with values 1st, 2nd and 3rd. The paradigm of the weak finite verb of English can be presented as in (\ref{ex:englishverb-paradigmspace}) (the presence of morpheme boundaries is not important at this point).

\begin{example}
\label{ex:englishverb-paradigmspace}
\begin{minipage}{.45\textwidth}
\begin{tabular}{lllll} \toprule
\Feature{person} & \FV{singular}  & \FV{plural} \\ \midrule 
\multicolumn{3}{c}{\FV{past}} \\
\FV{1st}	& jump+ed &jump+ed \\
\FV{2nd}	&  jump+ed &jump+ed \\
\FV{3rd}	& jump+ed &jump+ed\\ \bottomrule 
\end{tabular}
\end{minipage}
\begin{minipage}{.45\textwidth}
\begin{tabular}{lllll}\toprule
\Feature{person} & \FV{singular}  & \FV{plural} \\ \midrule 
\multicolumn{3}{c}{\FV{present}} \\
\FV{1st}	& jump  &jump \\
\FV{2nd}	&  jump &jump \\
\FV{3rd}	&  jump+s &jump\\ \bottomrule 
\end{tabular}
\end{minipage}
\end{example}

As a multidimensional array, each box or location in the array can be uniquely identified by a sequence of specifications in each dimension.\footnote{\citet[10-11]{Stump:2016} discusses some of the pros and cons of viewing a paradigm as a set of such boxes (or {\em cells}, as he calls them). Our view is that these cells are epiphenomena, even though the notion of a paradigm is not, and we explain our reasons over the course of this paper. Stump uses the phrase {\em inflectional category} where we use the more traditional term {\em feature}, and our {\em feature value} corresponds to his {\em morphosyntactic property}. We use what seems to us to be traditional terminology.} The box which contains {\em jump+s} can be identified as the position (Present, 3rd, singular). More generally, each position can be labeled as in (\ref{ex:englishverb-paradigmspace2}).

\begin{example}
\label{ex:englishverb-paradigmspace2}
\begin{tabular}{lllll} \toprule
\Feature{person} & \FV{singular}  & \FV{plural} \\ \midrule 
\textsc{past} \\
\FV{1st}	& (\FV{past, 1st, singular})   & (\FV{past, 1st, plural})  \\
\FV{2nd}	& (\FV{past, 2nd, singular}) &(\FV{past, 2nd, plural}) \\
\FV{3rd}	& (\FV{past, 3rd, singular})&(\FV{past, 3rd, plural})\\ \bottomrule 
\textsc{present} \\
\FV{1st}	& (\FV{present, 1st, singular})   & (\FV{present, 1st, plural})  \\
\FV{2nd}	& (\FV{present, 2nd, singular}) &(\FV{present, 2nd, plural}) \\
\FV{3rd}	& (\FV{present, 3rd, singular})&(\FV{present, 3rd, plural})\\ \bottomrule 
\end{tabular}
\end{example}

However, while this representation is familiar to linguists, we will often find it more useful to express this information in a list-like way, as in (\ref{ex:englishverb-paradigmspace3}), where we list the positions in the paradigm by ordering the dimensions of the paradigm in a fixed, conventional way (here, \Feature{tense/person/number}). This is a representation in {paradigm space}. It is not a vector space, and not a space at all in a geometrical sense; the dimensions here are features (which are functions) that take on only one of a discrete set of values, and those values are not ordered among each other. (Thus despite the fact that we typically order 1st, 2nd, 3rd person in that order, the order is not reflected in the structure of the model.)

\begin{example}
\label{ex:englishverb-paradigmspace3}
\begin{tabular}{lllll}   \toprule
 \multicolumn{2}{c}{Paradigm space}\\
 feature combination & verb & suffix \\  \midrule
\FV{(past, 1st, singular)}  & jump+ed  & ed  \\ 
\FV{(past, 2nd, plural)}  & jump+ed & ed \\
\FV{(past, 3rd , singular)} & jump+ed& ed \\
\FV{(past, 1st, plural)} & jump+ed & ed \\
\FV{(past, 2nd, singular)} & jump+ed & ed\\
\FV{(past, 3rd, plural)} & jump+ed& ed \\  
\FV{(present, 1st, singular)} & jump & $\nul$\\    
\FV{(present, 2nd, plural)} & jump & $\nul$  \\
\FV{(present, 3rd, singular)} & jump+s & s\\
\FV{(present, 1st, plural)} & jump & $\nul$ \\
\FV{(present, 2nd, singular)} & jump& $\nul$ \\
\FV{(present, 3rd, plural)} & jump& $\nul$ \\   \bottomrule
\end{tabular}
\end{example}

The principal role for paradigm space is as the interface between (morpho-)syntax and morphology. The syntax is concerned with specifying each syntactic position as one of the positions in paradigm space.

We can now construct a matrix which spells out the functions of each affix in the paradigm; we call this the {\em Total Paradigm Matrix}, or TPM. For the English weak verb, this is as in (\ref{english-TPM}). The columns of TPM describe the positions in the paradigm in which each morpheme occurs, a set of positions in paradigm space; we indicate that vector with a hat over the morpheme, as in the following table.

\begin{example} 
\label{english-TPM}
{\sc Total Paradigm Matrix}
\[
\begin{blockarray}{lccc}
&  \hat{\nul}   &  \hat{s}  &  \hat{ed}   \\
\begin{block} {l(ccc)}
\FV{(past, 1st, singular)} & 0  & 0 & 1\\ 
 \FV{(past, 2nd, plural)}  & 0  & 0 & 1\\
\FV{(past, 3rd , singular)} &  0  & 0 & 1\\ 
 \FV{(past, 1st, plural)} & 0  & 0 & 1\\
\FV{(past, 2nd, singular)} &  0  & 0 & 1\\
 \FV{(past, 3rd, plural)} & 0  & 0 & 1\\
\FV{(present, 1st, singular)} &   1 & 0 & 0 \\ 
\FV{(present, 2nd, plural)} &  1 & 0 & 0 \\
\FV{(present, 3rd, singular)} &   0 & 1 & 0 \\ 
 \FV{(present, 1st, plural)} &  1 & 0 & 0 \\
 \FV{(present, 2nd, singular)} & 1 & 0 & 0 \\
\FV{(present, 3rd, plural)} &  1 & 0 & 0 \\
\end{block}
\end{blockarray}
 \]
\end{example}

\subsection{Feature-value space}
Feature-value space has a different structure from paradigm space: it is the space in which most of our work in geometrical morphology takes place.  In feature-value space, each dimension corresponds to a value taken on by one of the features in the paradigm, and for the present, we will consider only the values 0 and 1 that may be assigned to a coordinate. In the case of the English verb system, the example we will return to often initially, the feature values are \FV{past, present, 1st person, 2nd person, 3rd person, singular}, and \FV{plural}. Thus in this example, feature-value space has 7 dimensions. A vector in that space is represented by a vector with 7 coordinates.  

The dimensions are ordered arbitrarily; we adopt the convention that the feature values are ordered: (\FV{Past, Present, 1st person, 2nd person, 3rd person, sg., pl.}). In  (\ref{ex:englishverb-featurevalue-of-paradigm}), we give the feature-value space representation for each position in the verbal paradigm of English:

\vspace{.15in}
 
\begin{example}
\label{ex:englishverb-featurevalue-of-paradigm}
\begin{tabular}{lllll} \toprule
\multicolumn{3}{c}{Feature value space} \\
\Feature{person} & \FV{singular}  & \FV{plural} \\ \midrule 
\textsc{past} \\
\FV{1st}	& (1,0,1,0,0,1,0) &(1,0,1,0,0,0,1) \\
\FV{2nd}	&  (1,0,0,1,0,1,0) &(1,0,0,1,0,0,1) \\
\FV{3rd}	&  (1,0,0,0,1,1,0) &(1,0,0,0,1,0,1)\\ \bottomrule 
\Feature{present} \\
\FV{1st}	& (0,1,1,0,0,1,0) &(0,1,1,0,0,0,1) \\
\FV{2nd}	&  (0,1,0,1,0,1,0) &(0,1,0,1,0,0,1) \\
\FV{3rd}	&  (0,1,0,0,1,1,0) &(0,1,0,0,1,0,1)\\ \bottomrule 
\end{tabular}
\end{example}
\vspace{.15in}

Those 12 corner points on a 7-dimensional cube could be thought of as messages coming from the syntax, or morphosyntax, if one likes a dynamic metaphor. In any event, this metaphorical ``message'' is limited to these corner points on the hypercube: only values of 0 and 1 for each dimension. This structure reflects the discrete aspect of inflectional morphology.

In (\ref{bigphi-english}) we present  a binary matrix that shows the relationship between the positions in the paradigm and the coordinates in feature value space, and we call it \PHI, as we indicate there. Each row is a possible message to be expressed; each column corresponds to a particular feature value. 

\begin{example}
\label{bigphi-english}
$
\begin{blockarray}{lccccccc}
 & \FV{past}  & \FV{present}  & \FV{1st}  & \FV{2nd}  & \FV{3rd}  & \FV{singular}  & \FV{plural}  \\
\begin{block}{l(ccccccc)}   
\FV{Past,1st, sg}   & 1 & 0 &  1 &  0 &  0 &  1 &  0 \\
\FV{Past,2nd, sg}  & 1  & 0 &  0 &  1 &  0 &  1 &  0\\
\FV{Past,3rd, sg}  & 1  & 0 &  0 &  0 &  1 &  1 &  0 \\
\FV{Past,1st, pl}  & 1  & 0 & 1 &  0 &  0 &  0 &  1 \\ 
\FV{Past,2nd, pl}  & 1  & 0 &  0 &  1 &  0 &  0 &  1 \\
\FV{Past,3rd, pl}  & 1 &  0 &   0 &  0 &  1 &  0 &  1 \\  
\FV{Present,1st, sg}  & 0  & 1 &  1 &  0 &  0 &  1 &  0 \\
\FV{Present,2nd, sg}  & 0  & 1 &  0 &  1 &  0 &  1 &  0 \\
\FV{Present,3rd, sg}  & 0  & 1 &  0 &  0 &  1 &  1 &  0 \\
\FV{Present,1st, pl}   & 0  & 1 &  1 &  0 &  0 &  0 &  1 \\
\FV{Present,2nd, pl}  & 0 &  1 &  0 &  1 &  0 &  0 &  1 \\
\FV{Present,3rd, pl}  & 0 &  1 &  0 &  0 &  1 &  0 &  1 \\  
\end{block}
\end{blockarray}
=\PHI
$
\end{example}

The columns (which express the feature values for each morpheme) divide up into sets which have the property that any two columns in the same set are orthogonal to each other, that is, that the inner product of any two is zero. In the case here, these sets are the first two columns, the next three columns, and the last two columns. For example, the inner product of the column vector for past and that for present is zero, since for each row, one vector or the other has the value 0. In addition, the sum of all of the vectors in each subset is the identity vector $1^7$. These subsets correspond to the linguist's conception of a {\em feature}, and there are 3 here: \Feature{tense, person}, and \Feature{number}.\footnote{See \citet{gr} for discussion of how such featural systems are learned, in a study of gender and case in German.} 
 
Another way to think about $\Phi$ involves the eigenmaps on the graph that represents the relationship between features and feature values. A graph whose nodes are feature values, and which contains edges between any two nodes whose feature values are values of the same feature, looks like this:

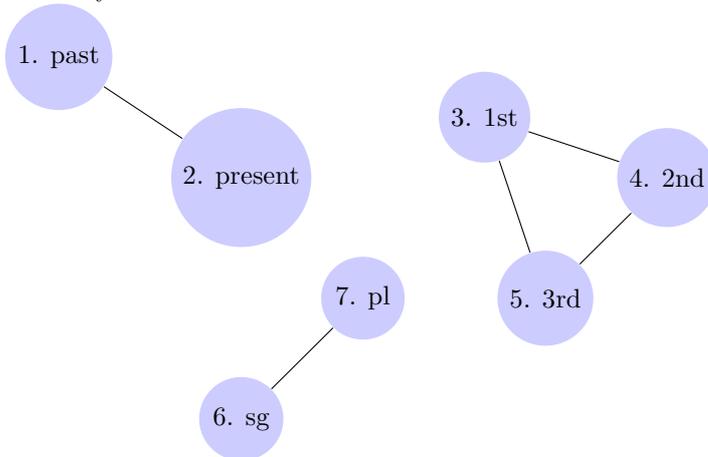
\begin{tikzpicture}
  [scale=.8,auto=left,every node/.style={circle,fill=blue!20}]
  \node (past) at (1,10) {1. past};
  \node (present) at (4,8)  {2. present};
  \node (p1) at (8,9)  {3. 1st};
  \node (p2) at (11,8) {4. 2nd};
  \node (p3) at (9,6)  {5. 3rd};
  \node (sg) at (4,4)  {6. sg};
  \node (pl) at (6,6) {7. pl};
  \foreach \from/\to in {past/present,p1/p2,p2/p3,p1/p3,sg/pl}
    \draw (\from) -- (\to);

\end{tikzpicture}

The eigenmaps on this graph are (1,1,0,0,0,0,0), (0,0,1,1,1,0,0), and (0,0,0,0,0,1,1), each corresponding to a linguistic features (tense, person and number, respectively).\footnote{An eigenmap on a graph is a function $f$ from nodes to reals  which is mapped to a multiple of $f$ by the laplacian of the graph. The laplacian of a graph is an operator on functions on graphs (i.e., on maps from nodes to reals) which can be thought of as a rigorous account of how an activation on a set of nodes would diffuse through the graph, if edge weight were understood as a measurement of channel width. If a graph has disconnected subgraphs, then for each disconnected subgraph there will be at least one eigenmap that are zero on all the other disconnected subgraphs.} 

The two objects for the reader to remember now are the Total Paradigm Matrix (TPM), and \PHI. The TPM expresses information about the paradigm, while \PHI{} expresses the relationship between paradigm space and feature-value space.

\subsection{Corners and the unit sphere in feature-value space}
 
 \subsubsection{Corners in feature-value space}
As we have observed, in feature-value space, there are certain points that are special, because they represent positions in the paradigm; they are the points whose coordinates appear as rows in the matrix \PHI. In the case of the English verb, these {\em corners} are the points that have exactly one 1 in the first two coordinates, exactly one 1 in the next three coordinates, and exactly one 1 in the final two coordinates (all other coordinates are 0). We will often use the letter $M$ to represent such a corner ($M$ for {\em message}).

In addition, each inflectional morpheme is represented by a vector in feature-value space. As we will see, the three verbal suffixes {\em ed}, {\em s}, and $\emptyset$ each correspond to vectors there. The problem of morpheme selection thus can be restated as: for each corner, find the appropriate morpheme. Our hypothesis is that 

\begin{mybox}
\centering{For each corner, the correct morpheme is the one that is closest to that corner.}
\end{mybox}

\subsubsection{The unit sphere in feature value space}

One of the central ideas of this approach is that the morphological characteristics of a morpheme are best understood in the feature-value space. Each morpheme is modeled by a vector from the origin to a position in the feature-value space. Furthermore, we take all morphemes to be associated with vectors of unit length, i.e., the length $|\mu|$ of a morpheme is equal to $\sqrt{\sum_i \mu_i^2}$; these vectors reach from the origin to a point on the surface of a hypersphere of unit radius.\footnote{It is very often the case that when we consider vectors in a high-dimensional space, when we speak of how close two vectors are, we do not mean the distance from the tip of one to the tip of the other, but care rather about the angle between the vectors. In such cases, it is common to say that we {\em normalize} the vectors, which is simply to shrink them to length 1 while maintaining their direction. Bear in mind that when we speak of {\em normalization}, we assume a particular notion of length or dimensionality. When we are dealing with probabilities, normalization generally refers to finding a scaling factor so that the sum of the values (here, the coordinates) sum to 1.0. This employs what is called the $L_1$ norm. When we think geometrically, we employ the $L_2$ norm, which means that normalization divides each term by a factor so that the sum of the squares of those normalized terms sum to 1.0}

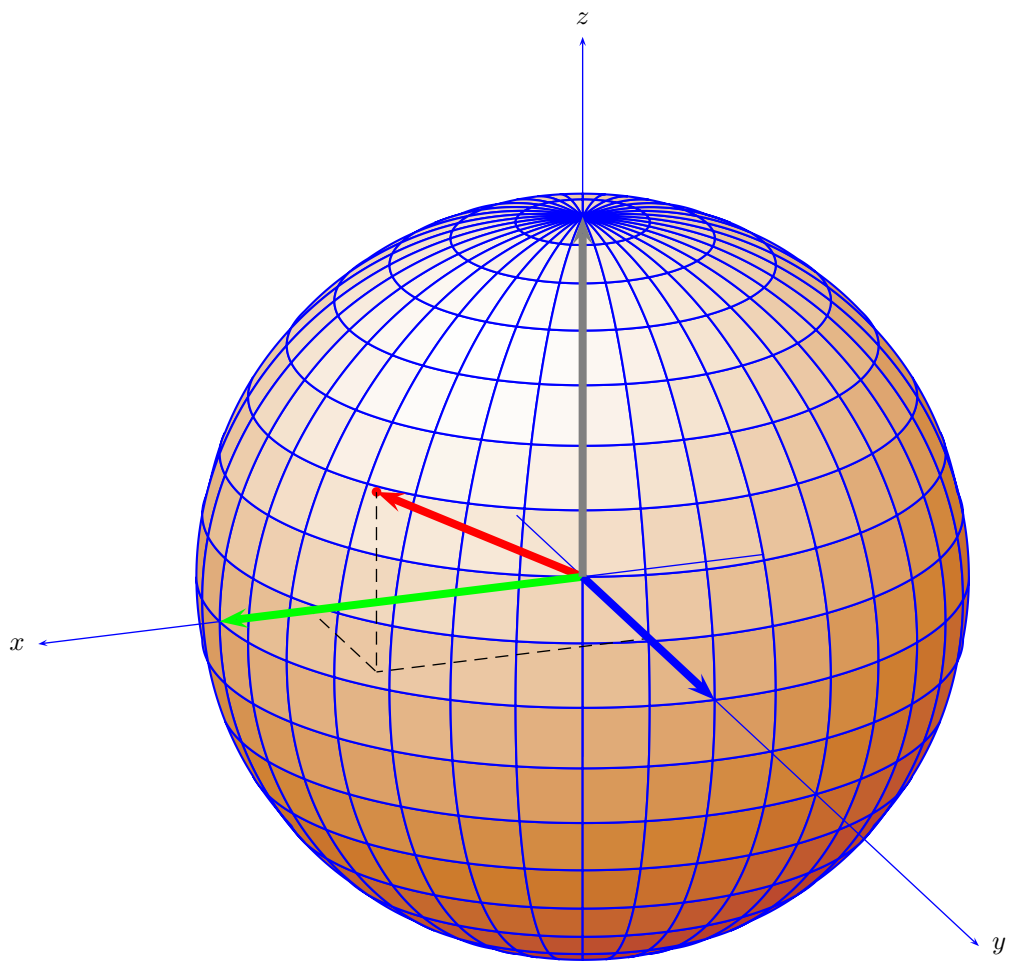
\begin{figure}
 \psset{unit=1in,Alpha=20,Beta=20}
\begin{pspicture}(-4,-2)(2,4)
\pstThreeDSphere[linecolor=blue](0,0,0){2}
\pstThreeDCoor[linecolor=blue,xMin=-1,xMax=3,yMin=-1,yMax=6,zMin=-1,zMax=3]
\psset{dotstyle=*,linecolor=red,drawCoor=true}
\pstThreeDDot(1.5,1,1)
\pstThreeDLine[linewidth=3pt, linecolor=red, arrows=-> ](0,0,0)(1.5,1,1)

\pstThreeDLine[linewidth=3pt, linecolor=green, arrows=-> ](0,0,0)(2,0,0)
\pstThreeDLine[linewidth=3pt, linecolor=blue, arrows=-> ](0,0,0)(0,2,0)
\pstThreeDLine[linewidth=3pt, linecolor=gray, arrows=-> ](0,0,0)(0,0,2)
\end{pspicture}
\caption{A sphere of radius 1, with four vectors of length 1.0}
\end{figure}

Thus (1,0,0,0,1,0) is not a possible vector for a morpheme, because its length is $\sqrt{2}$.  A morpheme that represents the feature-value ``singular'' can be represented as (1,0,0,0,0,0) since its length is 1.0. A morpheme which is specified as {\em nominative singular} can be represented as ($\frac{\sqrt{2}}{2}, 0,\frac{\sqrt{2}}{2},0,0$). The crucial point is that {\em morphosyntactic positions} are corners, that is, points whose coordinates are 0s and 1s (subject to constraints), while {\em morphemes} are vectors that  typically are {\em not} composed of only 0s and 1s: they are of unit length, however. Typically they take on positive values for those feature-values that they realize, for reasons that will emerge in our discussion below. The coordinates may take on negative values as well.\footnote{As we were writing this, we by good fortune came across what Scott Aaronson wrote in {\em Quantum Computing Since Democritus}: ``Now what happens if you try to come up with a theory that's {\em like} probability theory, but based on the 2-norm instead of on the 1-norm? I'm going to try to convince you that quantum mechanics is what inevitably results.'' \citep[112]{aaronson}}
 
\vspace{.15in}

The correct morpheme for each position is the morpheme which is {\em closest} to the position, in the geometric sense, in the feature-value space. We define closeness as the inner product of the two vectors (one for the position in the paradigm, one for the morpheme candidate).

These two vectors play different sorts of roles. The vectors representing morphemes are normalized to unit length, and are directly linked to observable characteristics of the word. The positions in a paradigm, that is to say the specification of the morphosyntactic features, correspond to corners of a hypercube, and a vector to a corner is typically not of unit length, and is in a certain sense hidden from the linguist. 

\begin{mybox}
We will see below that it is not always a {\em single} vector that we are looking for, one which is as close as possible to the target corner; it will in general be a set of vectors that we seek, whose vector sum is as close as possible to the corner specified by the grammar. When a word consists of several morphemes, the vector corresponding to that multimorphemic word is the (vector) sum of the vectors corresponding to each morpheme. We will explore this in section 3. Figure 2 illustrates vector addition in three dimensions, where the vector $\vec{\mu}$ is the sum of the three vectors $\vec{u}, \vec{v},$ and  $\vec{w}$.

\end{mybox}

We frequently compute the inner product of two such vectors, and in order to emphasize the difference between the two, we will make use of a notation in which the observable morpheme is placed thusly:  $\bra{\mu}$ and the morphosyntactic specification is written thusly: $\ket{M}$ (this notation is known as Dirac's bracket notation). The inner product of the two is written $\bra{\mu} \ket{M}$. The central formal idea that we develop in this paper employs the following choice procedure:\footnote{The bracket notation is intended to remind the reader of treating this as a collapse of the wave function, where M represents the state of a system, and the morphemes essentially {\em are} observation operations, and rather than assigning probability based on the inner product, the choice is deterministic; the morpheme (observation) with the largest inner product (hence, smallest angle) is predicted. Seeking the largest inner product, which we speak of {\em maximizing}, is often equivalent to choosing the vector with the smallest angle compared to some particular fixed vector, and so sometimes we will speak of minimizing an angle, which in the cases we consider is the same thing as maximizing the inner product. It is also sometimes convenient to think of a product $\bra{\mu} \ket{M}$ not as the result of a product on a pair of vectors, but rather as the result of evaluating a linear functional $\bra{\mu}$ evaluated on the vector $\ket{M}$, and we know that the space of linear functionals on a vector space is homomorphic to the space itself.
}

\begin{mybox}
\[ \hat{\mu}= argmax_i \bra{\mu_i} \ket{M}    \]
\end{mybox}

\noindent where  $\hat{\mu}$  is the morpheme selected by the grammar. 

We will hold in abeyance for just a moment how the vector that is associated with a given morpheme is determined; for the moment, consider the following matrix, whose column vectors specify the vectors for each affix in the English weak verb conjugation (\ref{englishverb-Bmatrix}). We refer to this matrix as \BB, and the reader can easily see that the column vectors are of unit length (the sum of the squares of their coordinates sum to 1.0).

\begin{example}
\label{englishverb-Bmatrix}
$
\begin{blockarray}{lccccccc}
 & \emptyset & -s & -ed   \\
\begin{block}{l(ccccccc)}
\FV{past} 		& 0 			& 0			&   \frac{6}{\sqrt{66}}  \\
\FV{present}		&  \frac{5}{\sqrt{47}}  &  \frac{1}{\sqrt{3}}  	& 0  \\
\FV{1st}		&  \frac{2}{\sqrt{47}} 	&  0 			&  \frac{2}{\sqrt{66}}    \\ 
\FV{2nd}		&  \frac{2}{\sqrt{47}}  &  0 			&   \frac{2}{\sqrt{66}}  \\
\FV{3rd}		&  \frac{1}{\sqrt{47}}  &  \frac{1}{\sqrt{3}}   &  \frac{2}{\sqrt{66}} \\
\FV{sg}		&  \frac{2}{\sqrt{47}}  &  \frac{1}{\sqrt{3}}   &   \frac{3}{\sqrt{66}} \\
\FV{pl}		&  \frac{3}{\sqrt{47}}  & 0			&   \frac{3}{\sqrt{66}} \\
\end{block}
\end{blockarray}
  = \BB$
\end{example}

which is approximately: 

\begin{example}
\vspace{.15in}
$\BB \approx
\begin{bmatrix}
0 & 0 &  .738 \\
.739 & .577  & 0  \\
.292 & 0.0 & 0.246 \\
.292 & 0  & 0.246 \\
.146 & .577 & .246 \\
.292 & .577 & .369 \\
.438 & 0  & .369 \\
\end{bmatrix}
$
\end{example}

\begin{figure}

\begin{pspicture}(-2,-1.25)(3,4.25)
\psset{Alpha=30,Beta=30}
\pstThreeDCoor[xMin=-3,xMax=1,yMin=-1,yMax=2,zMin=-1,zMax=4,nameX=3rd person,nameY=singular,nameZ=present tense]
\pstThreeDDot[drawCoor=true](-1,1,2)
\psset{arrows=->,arrowsize=0.2}
\pstThreeDLine[linecolor=green](0,0,0)(-1,1,2)
\uput[0](0.45,1.0){$\vec{\mu}$}
\uput[0](-0.5,.8){$\vec{u}$}
\uput[90](0.5,-1.0){$\vec{v}$}
\uput[45](0.65,0.2){$\vec{w}$}
\pstThreeDLine[linecolor=blue](0,0,0)(0,0,2)
\pstThreeDLine[linecolor=blue](0,0,0)(0,1,0)
\pstThreeDLine[linecolor=blue](0,0,0)(-1,0,0)
\end{pspicture}

\caption{A vector sum}
\end{figure}
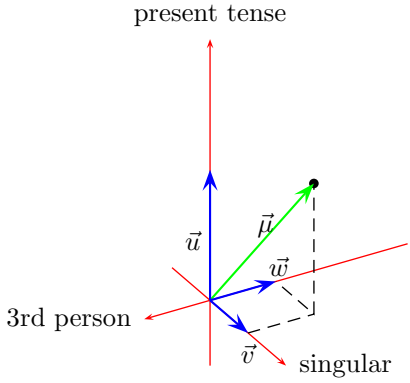

\subsection{From $\mathcal{B}$ to morpheme selection}

Let's see how this plays out for selection of the suffix in the present tense. We calculate the product   \PHI $\times$ \BB, which in effect gives us the inner product of each morpheme's vector (in (\ref{englishverb-Bmatrix})) with its position vector (in (\ref{bigphi-english})); we show only two decimal places because with even three, the table becomes visually unreadable.

\vspace{.1in}
 
\newpage
\begin{table}[h!]
\resizebox{\textwidth}{!}{%

\begin{tabular}{lcccll}  
  & $\emptyset$ & -s & -ed \\ \midrule
\FV{past 1st sg} & (1,0, 1,0,0, 1,0)(0,.73,  .29,.29,.15, .29,.44) & (1,0, 1,0,0, 1,0)(0,.58, 0,0,.58, .58,0) & (1,0, 1,0,0, 1,0)(.74,0, .25,,25,.25, .37,.37) \\ 
\FV{past 2nd sg}&  (1,0, 0,1,0, 1,0)(0,.73,  .29,.29,.15, .29,.44) & (1,0, 0,1,0, 1,0)(0,.58, 0,0,.58, .58,0) & (1,0, 0,1,0, 1,0)(.74,0, .25,,25,.25, .37,.37) \\
\FV{past 3rd sg} & (1,0, 0,0,1, 1,0)(0,.73,  .29,.29,.15, .29,.44) & (1,0, 0,0,1, 1,0)(0,.58, 0,0,.58, .58,0) & (1,0, 0,0,1, 1,0)(.74,0, .25,,25,.25, .37,.37)\\ 
\FV{past 1st pl} & (1,0, 1,0,0, 0,1)(0,.73,  .29,.29,.15, .29,.44) & (1,0, 1,0,0, 0,1)(0,.58, 0,0,.58, .58,0) & (1,0, 1,0,0, 0,1)(.74,0, .25,,25,.25, .37,.37) \\
\FV{past 2nd pl} & (1,0, 0,1,0, 0,1)(0,.73,  .29,.29,.15, .29,.44) & (1,0, 0,1,0, 0,1)(0,.58, 0,0,.58, .58,0) & (1,0, 0,1,0, 0,1)(.74,0, .25,,25,.25, .37,.37)\\
\FV{past 3rd pl} & (1,0, 0,0,1, 0,1)(0,.73,  .29,.29,.15, .29,.44) & (1,0, 0,0,1, 0,1)(0,.58, 0,0,.58, .58,0) & (1,0, 0,0,1, 0,1)(.74,0, .25,,25,.25, .37,.37)          \\
\FV{present 1st sg} & (0,1, 1,0,0, 1,0)(0,.73,  .29,.29,.15, .29,.44) & (0,1, 1,0,0, 1,0)(0,.58, 0,0,.58, .58,0) & (0,1, 1,0,0, 1,0)(.74,0, .25,,25,.25, .37,.37) \\ 
\FV{present 2nd sg}&  (0,1, 0,1,0, 1,0)(0,.73,  .29,.29,.15, .29,.44) & (0,1, 0,1,0, 1,0)(0,.58, 0,0,.58, .58,0) & (0,1, 0,1,0, 1,0)(.74,0, .25,,25,.25, .37,.37) \\
\FV{present 3rd sg} & (0,1, 0,0,1, 1,0)(0,.73,  .29,.29,.15, .29,.44) & (0,1, 0,0,1, 1,0)(0,.58, 0,0,.58, .58,0) & (0,1, 0,0,1, 1,0)(.74,0, .25,,25,.25, .37,.37)\\ 
\FV{present 1st pl} & (0,1, 1,0,0, 0,1)(0,.73,  .29,.29,.15, .29,.44) & (0,1, 1,0,0, 0,1)(0,.58, 0,0,.58, .58,0) & (0,1, 1,0,0, 0,1)(.74,0, .25,,25,.25, .37,.37) \\
\FV{present 2nd pl} & (0,1, 0,1,0, 0,1)(0,.73,  .29,.29,.15, .29,.44) & (0,1, 0,1,0, 0,1)(0,.58, 0,0,.58, .58,0) & (0,1, 0,1,0, 0,1)(.74,0, .25,,25,.25, .37,.37)\\
\FV{present 3rd pl} & (0,1, 0,0,1, 0,1)(0,.73,  .29,.29,.15, .29,.44) & (0,1, 0,0,1, 0,1)(0,.58, 0,0,.58, .58,0) & (0,1, 0,0,1, 0,1)(.74,0, .25,,25,.25, .37,.37) \\ \bottomrule
 
\end{tabular}}
\caption{\sc \PHI $\times$ \BB, Weak verb} 
 
\end{table}

\vspace{0.2in}
\noindent which equals the following (where we have put in blue the largest value in each row, marking the winning suffix):

\begin{example} 
 $\PHI \times \BB$=
\begin{tabular}{ccccl} 
suffix & \nul & -s & -ed \\ \midrule
\FV{1st sg past} & 0.584 & 0.577 & \Blue{1.353}\\ 
\FV{2nd sg past} & 0.584 & 0.577 & \Blue{1.353}\\
\FV{3rd sg past} & 0.438 & 1.154 & \Blue{1.353}\\ 
\FV{1st pl past} & 0.730 & 0     & \Blue{1.353}\\
\FV{2nd pl past} & 0.730 & 0     & \Blue{1.353}\\
\FV{3rd pl past} & 0.584 & 0.577 & \Blue{1.353}\\
 
\FV{1st sg pres} & \Blue{1.313} & 1.154 & 0.615\\ 
\FV{2nd sg pres} & \Blue{1.313} & 1.154 & 0.615\\
\FV{3rd sg pres} & 1.167 & \Blue{1.731} & 0.615\\ 
\FV{1st pl pres} & \Blue{1.459} & 0.577 & 0.615\\
\FV{2nd pl pres} & \Blue{1.459} & 0.577 & 0.615\\
\FV{3rd pl pres} & \Blue{1.313} & 1.154 & 0.615\\
\end{tabular}
\end{example}

We have already indicated that choosing the element with the highest value in each row will be a central part of our analysis, and so it is convenient to be able to refer to that element. We use a notation $Max_{rows}$(M,i,j) which we define as:

\[
	Max_{rows}(M,i,j) = 
	\begin{cases*}
		1  & if  for all  $k \ne j, M(i,j)>  M(i,k)$  \\
		0  & otherwise.	
	\end{cases*}
\]

and more generally, extending that function to the whole matrix, so that we may write $Max_{rows}(M)$:

\[
	Max_{rows}(M)(i,j) = Max_{rows}(M,i,j)
\]

With this notation, our colored, winning entries in \PHI $\times$ \BB \ take the value $1$ in $Min_{row}(\PHI \times \BB)$. This is what we have already labeled the {\em Total Paradigm Matrix} (TPM). What we have just shown is that the TPM can be computed from \PHI{} and \BB{}; this is {\em not} a mathematical certainty (we will see cases where a value of \BB{} is maximal and makes an incorrect prediction); there is always an empirical test as to whether a particular set of values for the morphemes (\BB) correctly leads to the TPM.\footnote{We use the term ``empirical'' here as it is often used in linguistics (though it would be hard to justify this usage outside linguistics) whereby a statement is empirical iff it could be shown to be wrong.} 

\begin{example} 
TPM: $Max_{rows}(\PHI \times \BB) =
\begin{bmatrix}  
  0  & 0 & 1\\ 
  0  & 0 & 1\\
  0  & 0 & 1\\ 
  0  & 0 & 1\\
  0  & 0 & 1\\
  0  & 0 & 1\\
 
  1 & 0 & 0 \\ 
  1 & 0 & 0 \\
  0 & 1 & 0 \\ 
  1 & 0 & 0 \\
  1 & 0 & 0 \\
  1 & 0 & 0 \\
\end{bmatrix}
$
\end{example}

\subsection{Smart initialization}
 
One of the most important aspects of our approach is its connection to a theory of learning. It is our goal to develop a model of morphology which stands or falls on its ability to project, or induce, a satisfactory analysis from data.  As before, we take it for granted that we are in possession of the morphosyntactic features of the system, and that the principal problem is how we assign a vector to each of the suffixes.

Let's return to the table in (\ref{bigphi-english}), which we repeat here as (\ref{bigphi-english-2}), the matrix $\Phi$ that has NumParaPos rows, and NumFeaVal columns. Each row in $\Phi$ corresponds to a position in the paradigm. A morpheme can be represented initially as a (column) vector with a 1 in each position which it represents in the paradigm. The whole paradigm is represented as a matrix, which we will call $\Phi$. As we will see shortly, $\Phi$ is a linear transformation that maps a paradigm representation (see below) to an element in feature-value space.

\begin{example}
\label{bigphi-english-2}
$
\begin{blockarray}{lccccccc}
 & \FV{past} & \FV{present} & \FV{1st} & \FV{2nd} & \FV{3rd} & \FV{singular} & \FV{plural} \\
\begin{block}{l(ccccccc)}   
\FV{Past,1st, sg}  & 1 & 0 &  1 &  0 &  0 &  1 &  0 \\
\FV{Past,2nd, sg}   & 1  & 0 &  0 &  1 &  0 &  1 &  0\\
\FV{Past,3rd, sg}  & 1  & 0 &  0 &  0 &  1 &  1 &  0 \\
\FV{Past,1st, pl}  & 1  & 0 & 1 &  0 &  0 &  0 &  1 \\ 
\FV{Past,2nd, pl}  & 1  & 0 &  0 &  1 &  0 &  0 &  1 \\
\FV{Past,3rd, pl}  & 1 &  0 &   0 &  0 &  1 &  0 &  1 \\  
\FV{Present,1st, sg}  & 0  & 1 &  1 &  0 &  0 &  1 &  0 \\
\FV{Present,2nd, sg}  & 0  & 1 &  0 &  1 &  0 &  1 &  0 \\
\FV{Present,3rd, sg}  & 0  & 1 &  0 &  0 &  1 &  1 &  0 \\
\FV{Present,1st, pl}   & 0  & 1 &  1 &  0 &  0 &  0 &  1 \\
\FV{Present,2nd, pl}  & 0 &  1 &  0 &  1 &  0 &  0 &  1 \\
\FV{Present,3rd, pl}  & 0 &  1 &  0 &  0 &  1 &  0 &  1 \\  
\end{block}
\end{blockarray}
=\PHI
$
\end{example}

The three morphemes $\emptyset$, s, and ed can be represented as vectors in this way. Here we are representing them as the points in the paradigm that they represent, so there are 12 rows, one for each position in the paradigm. We call these {\em paradigm representations}, and mark them with a hat:  $\hat{f}$.

\begin{example}
$\hat{\emptyset} = 
\begin{bmatrix}
0 \\
0 \\
0 \\
0 \\
0 \\
0 \\
1 \\
1 \\
0 \\
1 \\
1 \\
1 \\
\end{bmatrix}
$
$ \hat{s} = \begin{bmatrix}
0 \\
0 \\
0 \\
0 \\
0 \\
0 \\
0 \\
0 \\
1 \\
0 \\
0 \\
0 \\
\end{bmatrix}
$
$ \hat{ed} = \begin{bmatrix}
1\\
1 \\
1 \\
1 \\
1 \\
1 \\
0 \\
0 \\
0 \\
0 \\
0 \\
0 \\

\end{bmatrix}
$
\end{example}
\medskip

\label{pg:paradigmrep}

If we take the matrix product of $\Phi^t$ ($\Phi$ transpose) and each paradigm representation, we get a summary of each morpheme's expression of each feature-value. Here are the values that result. The vector that results exists in  feature-value space. For example, the first product shows that the null morpheme has a 0 value in the \FV{past}, 5 in the \FV{present}, 2 in \FV{1st person}, 2 in \FV{2nd person}, 1 in \FV{3rd person}, 2 in the \FV{singular}, and 3 in the \FV{plural}. In short: $\Phi^t$ maps from {\em PS} (paradigm space) to {\em FV} (feature value space).

\setcounter{MaxMatrixCols}{20}
\begin{example}
$ \bra{\Phi}\ket{\emptyset} = \Phi^t \hat{\emptyset}$ =
$
\begin{bmatrix} 
1 & 1 & 1 & 1 & 1 & 1 & 0 & 0 & 0 & 0 & 0 & 0 \\
0 & 0 & 0 & 0 & 0 & 0 & 1 & 1 & 1 & 1 & 1 & 1   \\
1 & 0 & 0 & 1 & 0 & 0 & 1 & 0 & 0 & 1 & 0 & 0\\
0 & 1 & 0 & 0 & 1 & 0 & 0 & 1 & 0 & 0 & 1 & 0\\
0 & 0 & 1 & 0 & 0 & 1 & 0 & 0 & 1 & 0 & 0 & 1\\
1 & 1 & 1 & 0 & 0 & 0 & 1 & 1 & 1 & 0 & 0 & 0\\
0 & 0 & 0 & 1 & 1 & 1 & 0 & 0 & 0 & 1 & 1 & 1\\
\end{bmatrix}   $
$
\begin{bmatrix}
0 \\
0 \\
0 \\
0 \\
0 \\
0 \\
1 \\
1 \\
0 \\
1 \\
1 \\
1 \\
\end{bmatrix}
$
=
$
\begin{bmatrix}
0 \\
5 \\
2 \\
2 \\
1 \\
2 \\
3 \\
\end{bmatrix}
$
\end{example}
\vspace{.15in}

Similarly,
 
\begin{example}
$ \bra{\Phi}\ket{s} = \Phi^t \hat{s} =
\begin{bmatrix} 
1 & 1 & 1 & 1 & 1 & 1 & 0 & 0 & 0 & 0 & 0 & 0 \\
0 & 0 & 0 & 0 & 0 & 0 & 1 & 1 & 1 & 1 & 1 & 1   \\
1 & 0 & 0 & 1 & 0 & 0 & 1 & 0 & 0 & 1 & 0 & 0\\
0 & 1 & 0 & 0 & 1 & 0 & 0 & 1 & 0 & 0 & 1 & 0\\
0 & 0 & 1 & 0 & 0 & 1 & 0 & 0 & 1 & 0 & 0 & 1\\
1 & 1 & 1 & 0 & 0 & 0 & 1 & 1 & 1 & 0 & 0 & 0\\
0 & 0 & 0 & 1 & 1 & 1 & 0 & 0 & 0 & 1 & 1 & 1\\
\end{bmatrix}   
\begin{bmatrix}
0 \\
0 \\
0 \\
0 \\
0 \\
0 \\
0 \\
0 \\
1 \\
0 \\
0 \\
0 \\
\end{bmatrix}
=
\begin{bmatrix}
0 \\
1 \\
0 \\
0 \\
1 \\
1 \\
0 \\
\end{bmatrix}
$
\end{example}

$s$ has a positive value in present, 3rd person, and singular.

\vspace{.15in} 
\begin{example}
$ \bra{\Phi}\ket{ed} = \Phi^t \hat{ed} =
\begin{bmatrix} 
1 & 1 & 1 & 1 & 1 & 1 & 0 & 0 & 0 & 0 & 0 & 0 \\
0 & 0 & 0 & 0 & 0 & 0 & 1 & 1 & 1 & 1 & 1 & 1   \\
1 & 0 & 0 & 1 & 0 & 0 & 1 & 0 & 0 & 1 & 0 & 0\\
0 & 1 & 0 & 0 & 1 & 0 & 0 & 1 & 0 & 0 & 1 & 0\\
0 & 0 & 1 & 0 & 0 & 1 & 0 & 0 & 1 & 0 & 0 & 1\\
1 & 1 & 1 & 0 & 0 & 0 & 1 & 1 & 1 & 0 & 0 & 0\\
0 & 0 & 0 & 1 & 1 & 1 & 0 & 0 & 0 & 1 & 1 & 1\\
\end{bmatrix}   
\begin{bmatrix}
1 \\
1 \\
1 \\
1 \\
1 \\
1 \\
1 \\
0 \\
0 \\
0 \\
0 \\
0 \\
0 \\
\end{bmatrix}
=
\begin{bmatrix}
6 \\
0 \\
2 \\
2 \\ 
2 \\
3 \\
3 \\
\end{bmatrix}
$
\end{example} 

\medskip

If matrix operations are not familiar to the reader, this operation corresponds to this: count the number of times each morphosyntactic features occurs with the realization of each suffix:

\begin{example}\label{ex:english-feature-counts}
\begin{tabular}{llll} \toprule
``Count array''		 & \nul & -s & -ed  \\ \midrule
\FV{past} 	     & 0    & 0 & 6\\
\FV{present}	 & 5    & 1 & 0 \\
\FV{1st}		 & 2    & 0 & 2  \\
\FV{2nd}         & 2    & 0 & 2\\
\FV{3rd}         & 1    & 1 & 2 \\
\FV{singular}    & 2    & 1 & 3 \\
\FV{plural}      & 3    & 0 & 3\\ \bottomrule
\end{tabular}
\end{example}

\vspace{.1in}


Now we normalize each column vector (one for each morpheme), making it of unit length, and we arrive at the tables given just above in (\ref{englishverb-Bmatrix}).  Having done that, we can use the {\em bra} $\bra{.}$ notation.

\vspace{.2in}

\begin{example}
\label{englishverb-Bmatrix-2}
$
\begin{blockarray}{lccccccc}
 & \bra{\emptyset} & \bra{-s} & \bra{-ed}   \\
\begin{block}{l(ccccccc)}
\FV{past} 		& 0 			& 0			&   \frac{6}{\sqrt{66}}  \\
\FV{present}		&  \frac{5}{\sqrt{47}}  &  \frac{1}{\sqrt{3}}  	& 0  \\
\FV{1st}		&  \frac{2}{\sqrt{47}} 	&  0 			&  \frac{2}{\sqrt{66}}    \\ 
\FV{2nd}		&  \frac{2}{\sqrt{47}}  &  0 			&   \frac{2}{\sqrt{66}}  \\
\FV{3rd}		&  \frac{1}{\sqrt{47}}  &  \frac{1}{\sqrt{3}}   &  \frac{2}{\sqrt{66}} \\
\FV{sg}		&  \frac{2}{\sqrt{47}}  &  \frac{1}{\sqrt{3}}   &   \frac{3}{\sqrt{66}} \\
\FV{pl}		&  \frac{3}{\sqrt{47}}  & 0			&   \frac{3}{\sqrt{66}} \\
\end{block}
\end{blockarray}
  = \mathcal{B}$
\end{example}

These values represent hypothesized settings for the morphemes derived directly from one occurrence of each point in the paradigm. The same operation can be performed with even an incomplete set of data from the paradigm. In either case, there is no guarantee that the morpheme-vectors correctly generate the appropriate forms for each point in the paradigm, which is why we refer to this process as {\em initialization}. We will explore shortly another learning process that we employ to deal with the cases in which smart initialization is not enough, a traditional learning rule called the Delta Rule. 

Let us summarize what we have done. We have used a vector/matrix notation to describe part of a relatively simple inflectional paradigm, and we have begun by assuming that we know the correct morpheme for each position in the paradigm. On the basis of that knowledge, we infer a vectorial representation for the three morphemes. Each box in the traditional paradigm corresponds to a corner of a hypercube in feature value space, and the morpheme that is closest to that corner is the correct morpheme for that position in the paradigm.

\vspace{.4in}
 
\subsection{Syncretism and blocking}

One of the characteristics of the present model which we find the most striking is the naturalness of what has traditionally been called {\em syncretism}, which is the use of the same inflectional morpheme to represent more than one position (Stump's {\em cell}) in the inflectional paradigm. This phenomenon has a natural interpretation in the present context: a particular morpheme is {\em used} to express a position in an inflectional paradigm if and only if it is the closest one to that position in the geometrical space that we are exploring. In the examples from Nuer that follow, we will be able to look at a system with a great deal of syncretism, and see how our system works in that context.

The normalization of morpheme vectors---taking them to be of unit length---is both mathematically natural and linguistically significant. One of the central observations in inflectional morphology flows from this characteristic: the effect that is often referred to as {\em blocking}. It will frequently be found that one morpheme is used to express a part of a paradigm (for example, the verbal suffix {\em -$\emptyset$} marks present tense in weak verbs), and a different morpheme is used to express a subpart of that part (for example, $-s$ marks the 3rd person singular present). The morpheme that indicates the subpart takes precedence over the morpheme that indicates the larger domain. Thus, from the grammar's point of view, the null verbal suffix is not specified as {\em avoiding} the 3rd person singular in any sense, but it does not in fact appear in 3rd person singular forms because it is superseded by the more specific morpheme. This effect is a result of the geometry of the system.\footnote{In section 3, we will return to the particular observation that in a broad range of cases, one finds a strong tendency for feature-values not to be realized on two different morphemes, intuitively speaking. In the framework we develop here, this is the result of a decomposition of the M vector into two parts that is effected by each morpheme realization.}

Let's look at a few examples to illustrate these ideas.

 \subsection{German verb}
\label{subsection:germanverb} 
Our first example will deal with the person and number suffixes for the German present tense verb. In this initial illustration, we will consider only the the suffix which marks person and number.

\begin{example}
\label{ex:german-paradigm1-affixes}
\begin{tabular}{lllll}  \toprule
& \multicolumn{2}{c}{\textsc{Number}} \\
 & \FV{singular}  & \FV{plural} \\ \midrule 
\Feature{person} \\
\FV{1st}	& sing + e & sing + en\\
\FV{2nd}	& sing + st & sing + t\\
\FV{3rd}	& sing + t & sing + en\\ \bottomrule 
\end{tabular}
\end{example}

\begin{example}
\label{german-FVs}
\begin{tabular}{llllll}
                    & e       &st      &en      &t       \\ \toprule
\FV{present, 1, sg} &     1  &    0   &    0   &    0  \\ 
\FV{present, 2, sg} &    0   &     1  &    0   &    0   \\ 
\FV{present, 3, sg} &    0   &    0   &    0   &     1  \\ 
\FV{present, 1, pl} &    0   &    0   &     1  &    0   \\ 
\FV{present, 2, pl} &    0   &    0   &    0   &     1  \\ 
\FV{present, 3, pl} &    0   &    0   &     1  &    0   \\ 
\end{tabular}
\vspace{.2in}
\end{example}
The same matrices express the relationship between representations as we saw for the English weak verb, but here in German we have four suffixes instead of three: \{-e, -st, -en, -t\}. We follow the same steps as we did for English and begin by counting the occurrences of morphosyntactic features for each suffix; see (\ref{ex:germansuffixcounts}), which is just like (\ref{ex:english-feature-counts}) above.

\begin{example}\label{ex:germansuffixcounts}

\begin{tabular}{llllllll}\toprule
 		    & e & st & en & t   \\ \midrule
\FV{past}   & 0  & 0 & 0  & 0  \\         
\FV{present}& 1  & 1 & 2  & 2  \\
\FV{1st}    & 1  & 0 & 1  & 0 \\
\FV{2nd}    & 0  & 1 & 0  & 1 \\
\FV{3rd}    & 0  & 0 & 1  & 1 \\
\FV{sg}     & 1  & 1 & 0  & 1 \\
\FV{pl}     & 0  & 0 & 2  & 1  \\ \bottomrule
\end{tabular}
\end{example}

Again we normalize each column, giving us the following values.  

\begin{example}\label{ex:simplemethod}
\begin{tabular}{llllllll}
 		& $\bra{e}$  & $\bra{st}$	& $\bra{en}$ 			& $\bra{t}$    \\ \midrule
\FV{past} & $\frac{0}{\sqrt{3}}$  	& $\frac{0}{\sqrt{3}}$ &   $\frac{0}{ \sqrt{10}}$ & $\frac{0}{\sqrt{8}}$  \\ 
\FV{present} & $\frac{1}{\sqrt{3}}$ & $\frac{1}{\sqrt{3}}$ & $\frac{2}{ \sqrt{10}}$ & $\frac{1}{\sqrt{2}}$ \\ 
\FV{1st} & $\frac{1}{\sqrt{3}}$	 	& $\frac{0}{\sqrt{3}}$ & $\frac{1}{ \sqrt{10}}$ &$\frac{0}{\sqrt{8}}$ \\ 
\FV{2nd} &  $\frac{0}{\sqrt{3}}$ 	& $\frac{1}{\sqrt{3}}$ &  $\frac{0}{ \sqrt{10}}$ &$\frac{1}{\sqrt{9}}$  \\ 
\FV{3rd} & $\frac{0}{\sqrt{3}}$ 	& $\frac{0}{\sqrt{3}}$ &  $\frac{1}{ \sqrt{10}}$ &  $\frac{1}{\sqrt{8}}$         \\ 
\FV{sg} &   $\frac{1}{\sqrt{3}}$ 	& $\frac{1}{\sqrt{3}}$ &  $\frac{0}{ \sqrt{10}}$ &$\frac{1}{\sqrt{8}}$  \\
\FV{pl} &  $\frac{0}{\sqrt{3}}$ 	& $\frac{0}{\sqrt{3}}$ &  $\frac{2}{ \sqrt{10}}$ & $\frac{1}{\sqrt{8}}$ \\  
\end{tabular}
\end{example}

 \vspace{.1in}

We can express this numerically as follows, where we simplify typographic clutter by replacing `0' by hyphen:

 \vspace{.1in}
\begin{example}\label{german-present-tense}
\BB =
\begin{tabular}{llllll} \toprule
& $\bra{e}$  & $\bra{st}$	& $\bra{en}$ 			& $\bra{t}$    \\ \midrule
\FV{past}         & -  &  -  &  -  &  - \\ 
\FV{present}         &  0.58  &  0.58  &  0.63  &  0.71  \\ 
\FV{1}               &  0.58  &    -   &  0.32  &    -   \\ 
\FV{2}               &    -   &  0.58  &    -   &  0.35  \\ 
\FV{3}               &    -   &    -   &  0.32  &  0.35  \\ 
\FV{sg}              &  0.58  &  0.58  &    -   &  0.35  \\ 
\FV{pl}              &    -   &    -   &  0.63  &  0.35  \\ \bottomrule
\end{tabular}
\end{example}

\vspace{.2in}

Limiting to the case of the present tense forms:

 \vspace{.2in}

\begin{example}
\begin{tabular}{llllllll}\toprule
                &present &1       &2       &3       &sg      &pl      \\  \midrule
\FV{present, 1, sg} &     1  &     1  &    0   &    0   &     1  &    0   \\ 
\FV{present, 2, sg} &     1  &    0   &     1  &    0   &     1  &    0   \\ 
\FV{present, 3, sg} &     1  &    0   &    0   &     1  &     1  &    0   \\ 
\FV{present, 1, pl} &     1  &     1  &    0   &    0   &    0   &     1  \\ 
\FV{present, 2, pl} &     1  &    0   &     1  &    0   &    0   &     1  \\ 
\FV{present, 3, pl} &     1  &    0   &    0   &     1  &    0   &     1  \\  \bottomrule
\end{tabular} 
\end{example}
\vspace{.2in}

Competition matrix:
 
\begin{example} 
\PHI $\times$ \BB =
\begin{tabular}{llllll} \toprule
  & $\bra{e}$       & $\bra{st}$       & $\bra{en}$       & $\bra{t}$        \\ \midrule
\FV{present, 1, sg}  & 1.73*  &  1.15  &  0.95  &  1.06  \\ 
\FV{present, 2, sg}  &  1.15  & 1.73*  &  0.63  &  1.41  \\ 
\FV{present, 3, sg}  &  1.15  &  1.15  &  0.95  & 1.41*  \\ 
\FV{present, 1, pl}  &  1.15  &  0.58  & 1.58*  &  1.06  \\ 
\FV{present, 2, pl}  &  0.58  &  1.15  &  1.26  & 1.41*  \\ 
\FV{present, 3, pl}  &  0.58  &  0.58  & 1.58*  &  1.41  \\  \bottomrule
\end{tabular}
\end{example}
\vspace{.2in}

This example is much like what we saw with the  English weak verb, but here we have four suffixes instead of three: \{-e, -st, -en, -t\}. We follow the same steps as we did for English ({\em smart initialization}), and begin by counting the occurrences of morphosyntactic features for each suffix; see (\ref{ex:germansuffixcounts}), which is just like (\ref{ex:english-feature-counts}) above. We can see that for this paradigm, smart initialization resulted in the correct choice of affixes for each paradigm position.

 \subsection{Latin adjectives}
We next look at how smart initialization fares for Latin adjectives. Here are the suffixes that occur on Latin adjectives according to their case, number and gender.

\bigskip

\begin{table}[ht]
\begin{tabular}{llll|lll} \hline
\Feature{Number}&\multicolumn{3}{c}{\FV{Singular}} &\multicolumn{3}{c}{\FV{Plural}} \\  
\Feature{Gender} & \FV{masculine} & \FV{feminine} & \FV{neuter} & \FV{masculine} & \FV{feminine} & \FV{neuter} \\ \hline
\Feature{Case:}  \\  
\FV{Nominative} & us & a & um & i & ae & a \\
\FV{Genitive} & i & ae & i & orum & arum & orum \\
\FV{Dative} & o & ae & o & is & is & is \\
\FV{Accusative} & um & am & um & os & as & a \\
\FV{Ablative} & o & a & o & is & is & is \\
\FV{Vocative} & e & a & um & i & ae & a \\ \hline
\end{tabular}
\caption{Latin adjectives}
\end{table}

\bigskip

When we apply smart initialization to this paradigm we get the following counts:

\bigskip

\begin{example}
\begin{tabular}{llllllllllllll}\toprule
& $us$ & $i$ & $ o$ & $um$ & $e$ & $a$ & $ae$ & $am$ & $ orum$ & $ is$ & $as$ & $arum$ & $os$ \\ \hline
\FV{sg} & 1 & 2 & 4 & 4 & 1 & 3 & 2 & 1 & 0 & 0 & 0 & 0 & 0  \\
\FV{pl}   & 0 & 2 & 0 & 0 & 0 & 3 & 2 & 0 & 2 & 6 & 1 & 1 & 1 \\
\FV{masc} & 1 & 3 & 2 & 1 & 1 & 0 & 0 & 0 & 1 & 2 & 0 & 0 & 1 \\
\FV{fem}  & 0 & 0 & 0 & 0 & 0 & 3 & 4 & 1 & 0 & 2 & 1 & 1 & 0 \\
\FV{neu}  & 0 & 1 & 2 & 3 & 0 & 3 & 0 & 0 & 1 & 2 & 0 & 0 & 0 \\
\FV{nom}  & 1 & 1 & 0 & 1 & 0 & 2 & 1 & 0 & 0 & 0 & 0 & 0 & 0 \\
\FV{gen}  & 0 & 2 & 0 & 0 & 0 & 0 & 1 & 0 & 2 & 0 & 0 & 1 & 0 \\
\FV{dat}  & 0 & 0 & 2 & 0 & 0 & 0 & 1 & 0 & 0 & 3 & 0 & 0 & 0 \\
\FV{acc}  & 0 & 0 & 0 & 2 & 0 & 1 & 0 & 1 & 0 & 0 & 1 & 0 & 1 \\
\FV{abl}  & 0 & 0 & 2 & 0 & 0 & 1 & 0 & 0 & 0 & 3 & 0 & 0 & 0 \\
\FV{voc}  & 0 & 1 & 0 & 1 & 1 & 2 & 1 & 0 & 0 & 0 & 0 & 0 & 0 \\ \bottomrule
\end{tabular}
\end{example}

Matrix $\mathcal{B}$ after normalizing:

\bigskip

\begin{example}
\(\mathcal{B}=\left[\begin{array}{llllllllllllll} 
&\bra{us} &    \bra{i} &   \bra{o} &   \bra{um} &   \bra{e} &   \bra{a} &   \bra{ae} &   \bra{am} &   \bra{orum} &   \bra{is} &   \bra{as} &   \bra{arum} &   \bra{os} \\ 
\FV{sg}&0.577  & 0.408  & 0.707  & 0.707  & 0.577  & 0.442  & 0.378  & 0.577  & 0  & 0  & 0  & 0  & 0 \\
\FV{pl}&0  & 0.408  & 0  & 0  & 0  & 0.442  & 0.378  & 0  & 0.632  & 0.739  & 0.577  & 0.577  & 0.577 \\
\FV{masc}& 0.577  & 0.612  & 0.354  & 0.177  & 0.577  & 0  & 0  & 0  & 0.316  & 0.246  & 0  & 0  & 0.577 \\
\FV{fem}&0  & 0  & 0  & 0  & 0  & 0.442  & 0.756  & 0.577  & 0  & 0.246  & 0.577  & 0.577  & 0 \\
\FV{neu}&0  & 0.204  & 0.354  & 0.530  & 0  & 0.442  & 0  & 0  & 0.316  & 0.246  & 0  & 0  & 0 \\
\FV{nom}&0.577  & 0.204  & 0  & 0.177  & 0  & 0.295  & 0.189  & 0  & 0  & 0  & 0  & 0  & 0 \\
\FV{gen}&0  & 0.408  & 0  & 0  & 0  & 0  & 0.189  & 0  & 0.632  & 0  & 0  & 0.577  & 0 \\
\FV{dat}&0  & 0  & 0.354  & 0  & 0  & 0  & 0.189  & 0  & 0  & 0.369  & 0  & 0  & 0 \\
\FV{acc}&0  & 0  & 0  & 0.354  & 0  & 0.147  & 0  & 0.577  & 0  & 0  & 0.577  & 0  & 0.577 \\
\FV{abl}&0  & 0  & 0.354  & 0  & 0  & 0.147  & 0  & 0  & 0  & 0.369  & 0  & 0  & 0 \\
\FV{voc}&0  & 0.204  & 0  & 0.177  & 0.577  & 0.295  & 0.189  & 0  & 0  & 0  & 0  & 0  & 0 \\
\end{array}\right]\)
\end{example}

Of the thirty cases in the array, 26 are correct from smart initialization, and 4 are not correct, and must be specifically learned after smart initialization.  In these the cases, the incorrect prediction is shown in red and the real winner in green. In \S\ref{sec:deltarule} we shall look at a further step beyond smart initialization that enables a speaker to modify feature values in order to obtain the correct morpheme for each paradigm position in all cases.

\newpage
\begin{example}\label{ex:LatinActivations}
\(\Phi\times\mathcal{B}=\left[\begin{array}{lllllllllllllll}
&&\mathrm{us} &    \mathrm{i} &   \mathrm{o} &   \mathrm{um} &   \mathrm{e} &   \mathrm{a} &   \mathrm{ae} &   \mathrm{am} &   \mathrm{orum} &   \mathrm{is} &   \mathrm{as} &   \mathrm{arum} &   \mathrm{os} \\ 
&n &\Blue{1.732}  & 1.225  & 1.061  & 1.061  & 1.155  & 0.737  & 0.567  & 0.577  & 0.316  & 0.246  & 0  & 0  & 0.577 \\
&g& 1.155  & \Blue{1.429}  & 1.061  & 0.884  & 1.155  & 0.442  & 0.567  & 0.577  & 0.949  & 0.246  & 0  & 0.577  & 0.577 \\
MSG&d& 1.155  & 1.021  & \Blue{1.414}  & 0.884  & 1.155  & 0.442  & 0.567  & 0.577  & 0.316  & 0.615  & 0  & 0  & 0.577 \\
&ac& 1.155  & 1.021  & 1.061  & \Blue{1.237}  & 1.155  & 0.590  & 0.378  & 1.155  & 0.316  & 0.246  & 0.577  & 0  & 1.155 \\
&ab &1.155  & 1.021  & \Blue{1.414}  & 0.884  & 1.155  & 0.590  & 0.378  & 0.577  & 0.316  & 0.615  & 0  & 0  & 0.577 \\
&v&1.155  & 1.225  & 1.061  & 1.061  & \Blue{1.732}  & 0.737  & 0.567  & 0.577  & 0.316  & 0.246  & 0  & 0  & 0.577 \\
&&&&&&&&&&&&&&\\
&n&1.155  & 0.612  & 0.707  & 0.884  & 0.577  & \Green{1.180}  & \Red{1.323}  & 1.155  & 0  & 0.246  & 0.577  & 0.577  & 0 \\
&g&0.577  & 0.816  & 0.707  & 0.707  & 0.577  & 0.885  & \Blue{1.323}  & 1.155  & 0.632  & 0.246  & 0.577  & 1.155  & 0 \\
FSG&d&0.577  & 0.408  & 1.061  & 0.707  & 0.577  & 0.885  & \Blue{1.323}  & 1.155  & 0  & 0.615  & 0.577  & 0.577  & 0 \\
&ac&0.577  & 0.408  & 0.707  & 1.061  & 0.577  & 1.032  & 1.134  & \Blue{1.732}  & 0  & 0.246  & 1.155  & 0.577  & 0.577 \\
&ab&0.577  & 0.408  & 1.061  & 0.707  & 0.577  & 1.032  & \Green{1.134}  & \Red{1.155}  & 0  & 0.615  & 0.577  & 0.577  & 0 \\
&v&0.577  & 0.612  & 0.707  & 0.884  & 1.155  & 1.180  & \Blue{1.323}  & 1.155  & 0  & 0.246  & 0.577  & 0.577  & 0 \\
&&&&&&&&&&&&&&\\
&n&1.155  & 0.816  & 1.061  & \Blue{1.414}  & 0.577  & 1.180  & 0.567  & 0.577  & 0.316  & 0.246  & 0  & 0  & 0 \\
&g&0.577  & \Green{1.021}  & 1.061  & \Red{1.237}  & 0.577  & 0.885  & 0.567  & 0.577  & 0.949  & 0.246  & 0  & 0.577  & 0 \\
NSG&d&0.577  & 0.612  & \Blue{1.414}  & 1.237  & 0.577  & 0.885  & 0.567  & 0.577  & 0.316  & 0.615  & 0  & 0  & 0 \\
&ac&0.577  & 0.612  & 1.061  & \Blue{1.591}  & 0.577  & 1.032  & 0.378  & 1.155  & 0.316  & 0.246  & 0.577  & 0  & 0.577 \\
&ab&0.577  & 0.612  & \Blue{1.414}  & 1.237  & 0.577  & 1.032  & 0.378  & 0.577  & 0.316  & 0.615  & 0  & 0  & 0 \\
&v&0.577  & 0.816  & 1.061  & \Blue{1.414}  & 1.155  & 1.180  & 0.567  & 0.577  & 0.316  & 0.246  & 0  & 0  & 0 \\
&&&&&&&&&&&&&&\\
&n&1.155  & \Blue{1.225}  & 0.354  & 0.354  & 0.577  & 0.737  & 0.567  & 0  & 0.949  & 0.985  & 0.577  & 0.577  & 1.155 \\
&g&0.577  & 1.429  & 0.354  & 0.177  & 0.577  & 0.442  & 0.567  & 0  & \Blue{1.581}  & 0.985  & 0.577  & 1.155  & 1.155 \\
MPL&d&0.577  & 1.021  & 0.707  & 0.177  & 0.577  & 0.442  & 0.567  & 0  & 0.949  & \Blue{1.354}  & 0.577  & 0.577  & 1.155 \\
&ac&0.577  & 1.021  & 0.354  & 0.530  & 0.577  & 0.590  & 0.378  & 0.577  & 0.949  & 0.985  & 1.155  & 0.577  & \Blue{1.732} \\
&ab&0.577  & 1.021  & 0.707  & 0.177  & 0.577  & 0.590  & 0.378  & 0  & 0.949  & \Blue{1.354}  & 0.577  & 0.577  & 1.155 \\
&v&0.577  & \Blue{1.225}  & 0.354  & 0.354  & 1.155  & 0.737  & 0.567  & 0  & 0.949  & 0.985  & 0.577  & 0.577  & 1.155 \\
&&&&&&&&&&&&&&\\
&n&0.577  & 0.612  & 0  & 0.177  & 0  & 1.180  & \Blue{1.323}  & 0.577  & 0.632  & 0.985  & 1.155  & 1.155  & 0.577 \\
&g&0  & 0.816  & 0  & 0  & 0  & 0.885  & 1.323  & 0.577  & 1.265  & 0.985  & 1.155  & \Blue{1.732}  & 0.577 \\
FPL&d&0  & 0.408  & 0.354  & 0  & 0  & 0.885  & 1.323  & 0.577  & 0.632  & \Blue{1.354}  & 1.155  & 1.155  & 0.577 \\
&ac&0  & 0.408  & 0  & 0.354  & 0  & 1.032  & 1.134  & 1.155  & 0.632  & 0.985  & \Blue{1.732}  & 1.155  & 1.155 \\
&ab&0  & 0.408  & 0.354  & 0  & 0  & 1.032  & 1.134  & 0.577  & 0.632  & \Blue{1.354}  & 1.155  & 1.155  & 0.577 \\
&v&0  & 0.612  & 0  & 0.177  & 0.577  & 1.180  & \Blue{1.323}  & 0.577  & 0.632  & 0.985  & 1.155  & 1.155  & 0.577 \\
&&&&&&&&&&&&&&\\
&n&0.577  & 0.816  & 0.354  & 0.707  & 0  & \Blue{1.180}  & 0.567  & 0  & 0.949  & 0.985  & 0.577  & 0.577  & 0.577 \\
&g&0  & 1.021  & 0.354  & 0.530  & 0  & 0.885  & 0.567  & 0  & \Blue{1.581}  & 0.985  & 0.577  & 1.155  & 0.577 \\
NPL&d&0  & 0.612  & 0.707  & 0.530  & 0  & 0.885  & 0.567  & 0  & 0.949  & \Blue{1.354}  & 0.577  & 0.577  & 0.577 \\
&ac&0  & 0.612  & 0.354  & 0.884  & 0  & \Green{1.032}  & 0.378  & 0.577  & 0.949  & 0.985  & \Red{1.155}  & 0.577  & 1.155 \\
&ab&0  & 0.612  & 0.707  & 0.530  & 0  & 1.032  & 0.378  & 0  & 0.949  & \Blue{1.354}  & 0.577  & 0.577  & 0.577 \\
&v&0  & 0.816  & 0.354  & 0.707  & 0.577  & \Blue{1.180}  & 0.567  & 0  & 0.949  & 0.985  & 0.577  & 0.577  & 0.577 \\
\end{array}\right]\)
\end{example}

 \subsection{Russian, 1 class}
 
 The Russian nominal declension classes have been much studied in modern studies of inflectional morphology. Here we present the paradigm of  a Russian noun that falls into the first of four  classes.\footnote{The literature on Russian nominal inflectional morphology is large. See \citet{corbett-fraser}. \citet[114]{corbett-fraser}: ``We have presented four declensional classes. This is not the traditional account; most descriptions recognize only three, treating zakon and v'ino as variants of a single declensional lcass (as in, for instance, \citet{Vinogradov}, \citet{Unbegaun} and \citet{Stankiewicz}.)''
 
Regarding formalism, \citet{corbett-fraser} reference \citet{Evans}, and also the ELU formalism of \citet{Russell} and Word Grammar formalism of \citet{Fraser}.

We adopt the analysis by which [\Bi] is derived from more basic [i] when it follows a non-back hard consonant; some treat [\Bi] as a distinct underlying vowel. Consonants fall into three distinct categories with regarding to palatalization (the hard/soft contrast). Most have both a hard and soft version, which is indicated as C and C'. /\v{c} \v{z} c/ are hard, and remain unchanged in environments where we would expect softening (notably before a suffix beginning with /e/). Two consonants, /\v{c}'/ and /\v{s}\v{c}'/, are soft in all positions (that is, morphemes containing these consonants appear in only one form). /k,g,x/ appear in soft form before /i/.}

 \medskip
 
 \begin{example}\label{ex:russnouns1}
\begin{tabular}{lll}\hline
&\multicolumn{2}{c}{\Gl{law} \textsc{class i}}   \\
 & \FV{singular} & \FV{plural}\\ \hline
\FV{nom} & zakon & zakoni  \\
\FV{gen} & zakona & zakonov  \\
\FV{acc} & zakon & zakoni\\
\FV{loc} & zakone & zakonax  \\
\FV{dat} & zakonu & zakonam  \\
\FV{inst} & zakonom & zakonam'i \\ \hline
\end{tabular}
\end{example}

\medskip

Applying smart initialization gives us the following counts:

\medskip

\begin{example}
\begin{tabular}{lllllllllll}
&   $\emptyset$ & a & e & u & om & y & ov & ax & am & ami \\ \hline
\FV{sg} & 2 & 1 & 1 & 1 & 1 & 0 & 0 & 0 & 0 & 0 \\
\FV{pl} & 0 & 0 & 0 & 0 & 0 & 2 & 1 & 1 & 1 & 1 \\
\FV{nom} & 1 & 0 & 0 & 0 & 0 & 1 & 0 & 0 & 0 & 0 \\
\FV{gen} & 0 & 1 & 0 & 0 & 0 & 0 & 1 & 0 & 0 & 0 \\
\FV{acc} & 1 & 0 & 0 & 0 & 0 & 1 & 0 & 0 & 0 & 0 \\
\FV{loc} & 0 & 0 & 1 & 0 & 0 & 0 & 0 & 1 & 0 & 0 \\
\FV{dat} & 0 & 0 & 0 & 1 & 0 & 0 & 0 & 0 & 1 & 0 \\
\FV{inst} & 0 & 0 & 0 & 0 & 1 & 0 & 0 & 0 & 0 & 1 \\
\end{tabular}
\end{example}

\medskip

Expressed algebraically:

\medskip

\begin{example}
\(\mathcal{B}=\left[\begin{array}{llllllllll}
\frac{2}{\sqrt{6}}  & \frac{1}{\sqrt{2}}  & \frac{1}{\sqrt{2}}  & \frac{1}{\sqrt{2}}  & \frac{1}{\sqrt{2}}  & 0  & 0  & 0  & 0  & 0 \\
0  & 0  & 0  & 0  & 0  & \frac{2}{\sqrt{6}}  & \frac{1}{\sqrt{2}}  & \frac{1}{\sqrt{2}}  & \frac{1}{\sqrt{2}}  & \frac{1}{\sqrt{2}} \\
\frac{1}{\sqrt{6}}  & 0  & 0  & 0  & 0  & \frac{1}{\sqrt{6}}  & 0  & 0  & 0  & 0 \\
0  & \frac{1}{\sqrt{2}}  & 0  & 0  & 0  & 0  & \frac{1}{\sqrt{2}}  & 0  & 0  & 0 \\
\frac{1}{\sqrt{6}}  & 0  & 0  & 0  & 0  & \frac{1}{\sqrt{6}}  & 0  & 0  & 0  & 0 \\
0  & 0  & \frac{1}{\sqrt{2}}  & 0  & 0  & 0  & 0  & \frac{1}{\sqrt{2}}  & 0  & 0 \\
0  & 0  & 0  & \frac{1}{\sqrt{2}}  & 0  & 0  & 0  & 0  & \frac{1}{\sqrt{2}}  & 0 \\
0  & 0  & 0  & 0  & \frac{1}{\sqrt{2}}  & 0  & 0  & 0  & 0  & \frac{1}{\sqrt{2}} \\
\end{array}\right]\)
\end{example}

\medskip

Matrix $\mathcal{B}$ expressed numerically:

\medskip

\begin{example}
\(\mathcal{B}=\left[\begin{array}{llllllllll}
0.816  & 0.707  & 0.707  & 0.707  & 0.707  & 0  & 0  & 0  & 0  & 0 \\
0  & 0  & 0  & 0  & 0  & 0.816  & 0.707  & 0.707  & 0.707  & 0.707 \\
0.408  & 0  & 0  & 0  & 0  & 0.408  & 0  & 0  & 0  & 0 \\
0  & 0.707  & 0  & 0  & 0  & 0  & 0.707  & 0  & 0  & 0 \\
0.408  & 0  & 0  & 0  & 0  & 0.408  & 0  & 0  & 0  & 0 \\
0  & 0  & 0.707  & 0  & 0  & 0  & 0  & 0.707  & 0  & 0 \\
0  & 0  & 0  & 0.707  & 0  & 0  & 0  & 0  & 0.707  & 0 \\
0  & 0  & 0  & 0  & 0.707  & 0  & 0  & 0  & 0  & 0.707 \\
\end{array}\right]\)
\end{example}

\bigskip

\begin{example}
\(\phi\times\mathcal{B}=\left[\begin{array}{lllllllllll}
  &\emptyset & a & e & u & om & y & ov & ax & am & ami \\ 
n.sg.&\Blue{1.225}  & 0.707  & 0.707  & 0.707  & 0.707  & 0.408  & 0  & 0  & 0  & 0 \\
gen.sg.&0.816  & \Blue{1.414}  & 0.707  & 0.707  & 0.707  & 0  & 0.707  & 0  & 0  & 0 \\
acc.sg.&\Blue{1.225}  & 0.707  & 0.707  & 0.707  & 0.707  & 0.408  & 0  & 0  & 0  & 0 \\
loc.sg.&0.816  & 0.707  & \Blue{1.414}  & 0.707  & 0.707  & 0  & 0  & 0.707  & 0  & 0 \\
dat.sg.&0.816  & 0.707  & 0.707  & \Blue{1.414}  & 0.707  & 0  & 0  & 0  & 0.707  & 0 \\
inst.sg.&0.816  & 0.707  & 0.707  & 0.707  & \Blue{1.414}  & 0  & 0  & 0  & 0  & 0.707 \\
nom.pl.&0.408  & 0  & 0  & 0  & 0  & \Blue{1.225}  & 0.707  & 0.707  & 0.707  & 0.707 \\
gen.pl.&0  & 0.707  & 0  & 0  & 0  & 0.816  & \Blue{1.414}  & 0.707  & 0.707  & 0.707 \\
acc.pl.&0.408  & 0  & 0  & 0  & 0  & \Blue{1.225}  & 0.707  & 0.707  & 0.707  & 0.707 \\
loc.pl.&0  & 0  & 0.707  & 0  & 0  & 0.816  & 0.707  & \Blue{1.414}  & 0.707  & 0.707 \\
dat.pl.&0  & 0  & 0  & 0.707  & 0  & 0.816  & 0.707  & 0.707  & \Blue{1.414}  & 0.707 \\
inst.pl.&0  & 0  & 0  & 0  & 0.707  & 0.816  & 0.707  & 0.707  & 0.707  & \Blue{1.414} \\
\end{array}\right]\)
\end{example}

We find that smart initialization works perfectly for this particular paradigm. We next look at what steps our model takes beyond smart initialization in order to deal with cases where it does not always predict the correct morpheme for a given paradigm position.
\section{Learning: the Delta Rule}\label{sec:deltarule}

\subsection{The role of learning in linguistic analysis}

In this section, we will extend our concern with learning, which is to say, with construction of an algorithm that maps from data to the parametric values of the grammar. 

Learning and learnability play an important role in the framework that we explore in this paper, and we would like to point out that in our view, the linguist should understand that success in understanding learning goes hand in hand with no longer accepting the idea that it is an {\em advantage} for a linguistic theory to rule out certain grammars, or that we should prefer theory A over theory B because the theory B permits some grammars or languages that theory A does not. To say that we do not accept those views seems like such heresy that we need to explain why a bit more. The reader is cetainly not obliged to accept any of our views on this subject, but it does inform the work that we present here.

There are four reasons why we {\em reject} the notion that the value of a linguistic proposal should be evaluated by how well it limits or restricts the class of possible human languages. The first is that the notion of ``more restricted theory''  was imported surreptiously into linguistics, inappropriately from Popper's conception of science and inappropriately as a place-holder for learnability. The second is that we have little evidence of what does not occur in grammar (either in occurring grammars or in non-occurring but possible grammars). The third (closely related to the second) is that we are still in an expansionary phase of linguistics, in which every successful piece of research involves the discovery of new organizational principles of grammar, and that no successful piece of research has ever succeeded by ruling out a set of grammars. The fourth is simply that an implicit, and illicit, connection has been made between reducing the number of knowable human languages and the difficulty of solving the problem of language learning, but that implicit connection does not stand the light of what we know about machine learning today. We have added some remarks on this in an appendix at the end of this paper (secction 6).

\subsection{Smart initialization: example from German}
We have employed what we call {\em smart initialization} in the way we assigned the vectors associated with each morpheme. Smart initialization amounts to simply using the observed frequencies of feature values, observed in each of the cases where a particular morpheme is observed, to directly inform the coordinates of the morpheme in the simplest way imaginable. This method will often produce values that work correctly, but there is no guarantee that it will, and on a number of occasions, it does not. We will explore here what learning needs to take place to correct the placement of the morphemes.\footnote{We would expect that smart initialization would constitute a diachronic attraction for a language, in the sense that all other things being equal, there would be a tendency for a language to shift towards the the system described by smart initialization. We hope to explore this soon.} 

Let's look at a case where smart initialization does not do the job. We looked at the case of person-number endings of the German verb in the present tense above, in section \ref{subsection:germanverb}. If we add the past tense forms of the weak verb in German, things do not work out as well for this method, and the 3rd person singular present is assigned the wrong suffix under smart initialization as we have presented it so far.\footnote{A weak verb is one whose stem does not change in the past, and which contains a past tense suffix -t- between the stem and the person-number suffix. The account is due to Jacob Grimm 1819. A number of common verbs, called ``mixed'' verbs, have distinct past tense stems, like the strong verbs, but take the suffixes of a weak verb in the past tense.} Consider the data in (\ref{ex:german-paradigm1}).

\begin{example}
\label{ex:german-paradigm1}
\begin{tabular}{lllll} \toprule
\Feature{number} & \FV{singular}  & \FV{plural} \\ \midrule 
\Feature{person}&\multicolumn{2}{c}{\FV{past}} \\
\FV{1st}	& lieb+t+e &lieb+t+en \\
\FV{2nd}	& lieb+t+(e)st &lieb+t+(e)t \\
\FV{3rd}	&  lieb+t+e &lieb+t+en \\ \bottomrule 
&\multicolumn{2}{c}{\FV{present}} \\
\FV{1st}	& lieb+e& lieb+en\\
\FV{2nd}	& lieb+st & lieb+t\\
\FV{3rd}	& lieb+t &lieb+en\\ \bottomrule 
\end{tabular}
\end{example}

\begin{example}
\begin{tabular}{llllll}
                &e       &st      &en      &t       \\ \toprule
\FV{past, 1, sg} &     1  &    -   &    -   &    -   \\ 
\FV{past, 2, sg} &    -   &     1  &    -   &    -   \\ 
\FV{past, 3, sg} &     1  &    -   &    -   &    -   \\ 
\FV{past, 1, pl} &    -   &    -   &     1  &    -   \\ 
\FV{past, 2, pl} &    -   &    -   &    -   &     1  \\ 
\FV{past, 3, pl} &    -   &    -   &     1  &    -   \\ 
\FV{present, 1, sg} &     1  &    -   &    -   &    -   \\ 
\FV{present, 2, sg} &    -   &     1  &    -   &    -   \\ 
\FV{present, 3, sg} &    -   &    -   &    -   &     1  \\ 
\FV{present, 1, pl} &    -   &    -   &     1  &    -   \\ 
\FV{present, 2, pl} &    -   &    -   &    -   &     1  \\ 
\FV{present, 3, pl} &    -   &    -   &     1  &    -   \\ \bottomrule
\end{tabular}
\end{example}

\vspace{.2in}

\begin{example}
\PHI =  
\begin{tabular}{lccccccc}\toprule
                &\FV{past}    &\FV{present} &\FV{1}       &\FV{2}       &\FV{3}       &\FV{sg}      &\FV{pl}      \\ \midrule 
\FV{past, 1, sg} &     1  &    -   &     1  &    -   &    -   &     1  &    -   \\ 
\FV{past, 2, sg} &     1  &    -   &    -   &     1  &    -   &     1  &    -   \\ 
\FV{past, 3, sg} &     1  &    -   &    -   &    -   &     1  &     1  &    -   \\ 
\FV{past, 1, pl} &     1  &    -   &     1  &    -   &    -   &    -   &     1  \\ 
\FV{past, 2, pl} &     1  &    -   &    -   &     1  &    -   &    -   &     1  \\ 
\FV{past, 3, pl} &     1  &    -   &    -   &    -   &     1  &    -   &     1  \\ 
\FV{present, 1, sg} &    -   &     1  &     1  &    -   &    -   &     1  &    -   \\ 
\FV{present, 2, sg} &    -   &     1  &    -   &     1  &    -   &     1  &    -   \\ 
\FV{present, 3, sg} &    -   &     1  &    -   &    -   &     1  &     1  &    -   \\ 
\FV{present, 1, pl} &    -   &     1  &     1  &    -   &    -   &    -   &     1  \\ 
\FV{present, 2, pl} &    -   &     1  &    -   &     1  &    -   &    -   &     1  \\ 
\FV{present, 3, pl} &    -   &     1  &    -   &    -   &     1  &    -   &     1  \\ \bottomrule 
\end{tabular}
\end{example}

\vspace{.2in}

Count array:

\begin{example}
\begin{tabular}{llllll} \toprule
                &e       &st      &en      &t       \\ \midrule
\FV{past}        &     2  &     1  &     2  &     1  \\ 
\FV{present}     &     1  &     1  &     2  &     2  \\ 
\FV{1}           &     2  &    -   &     2  &    -   \\ 
\FV{2}           &    -   &     2  &    -   &     2  \\ 
\FV{3}           &     1  &    -   &     2  &     1  \\ 
\FV{sg}          &     3  &     2  &    -   &     1  \\ 
\FV{pl}          &    -   &    -   &     4  &     2  \\ \bottomrule
\end{tabular}
\end{example}

\vspace{.2in}

\begin{example} 
\begin{tabular}{llllllll} \toprule
 		& $\bra{e}$  & $\bra{st}$& $\bra{en}$ & $\bra{t}$    \\ \midrule
\FV{past} & $\frac{2}{\sqrt{19}}$  &  $\frac{1}{\sqrt{10}}$ &   $\frac{2}{4\sqrt{2}}$ & $\frac{1}{\sqrt{15}}$  \\ 
\FV{present} & $\frac{1}{\sqrt{19}}$ & $\frac{1}{\sqrt{10}}$ & $\frac{2}{4\sqrt{2}}$ & $\frac{2}{\sqrt{15}}$ \\ 
\FV{1st} & $\frac{2}{\sqrt{19}}$ & $\frac{0}{\sqrt{10}}$ & $\frac{2}{4\sqrt{2}}$ &$\frac{0}{\sqrt{15}}$ \\ 
\FV{2nd} &  $\frac{0}{\sqrt{19}}$ & $\frac{2}{\sqrt{10}}$ &  $\frac{0}{4\sqrt{2}}$ &$\frac{2}{\sqrt{15}}$  \\ 
\FV{3rd} & $\frac{1}{\sqrt{19}}$ & $\frac{0}{\sqrt{10}}$ &  $\frac{2}{4\sqrt{2}}$ &  $\frac{1}{\sqrt{15}}$         \\ 
\FV{sg} &   $\frac{3}{\sqrt{19}}$ &$\frac{2}{\sqrt{10}}$ &  $\frac{0}{4\sqrt{2}}$ &$\frac{1}{\sqrt{15}}$  \\
\FV{pl} &  $\frac{0}{\sqrt{19}}$ & $\frac{0}{\sqrt{10}}$ &  $\frac{4}{4\sqrt{2}}$ & $\frac{2}{\sqrt{15}}$ \\ \bottomrule 
\end{tabular} = \BB.
\end{example}

\vspace{0.2in}

Or approximately:

\begin{example} 
\begin{tabular}{llllll}\toprule
                  &e       &st      &en      &t       \\ \midrule
\FV{past}            &  0.46  &  0.32  &  0.35  &  0.26  \\ 
\FV{present}         &  0.23  &  0.32  &  0.35  &  0.52  \\ 
\FV{1}               &  0.46  &    -   &  0.35  &    -   \\ 
\FV{2}               &    -   &  0.63  &    -   &  0.52  \\ 
\FV{3}               &  0.23  &    -   &  0.35  &  0.26  \\ 
\FV{sg}              &  0.69  &  0.63  &    -   &  0.26  \\ 
\FV{pl}              &    -   &    -   &  0.71  &  0.52  \\ \bottomrule
\end{tabular} $\approx$ \BB
\end{example}

\vspace{.2in}

\PHI $\times$ \BB: Competition matrix 

\nopagebreak
\begin{example}
\begin{tabular}{llllll}\toprule
                  &e       &st      &en      &t       \\ 
\FV{past, 1, sg}     & 1.61*  &  0.95  &  0.71  &  0.52  \\ 
\FV{past, 2, sg}     &  1.15  & 1.58*  &  0.35  &  1.03  \\ 
\FV{past, 3, sg}     & 1.38*  &  0.95  &  0.71  &  0.77  \\ 
\FV{past, 1, pl}     &  0.92  &  0.32  & 1.41*  &  0.77  \\ 
\FV{past, 2, pl}     &  0.46  &  0.95  &  1.06  & 1.29*  \\ 
\FV{past, 3, pl}     &  0.69  &  0.32  & 1.41*  &  1.03  \\ 
\FV{present, 1, sg}  & 1.38*  &  0.95  &  0.71  &  0.77  \\ 
\FV{present, 2, sg}  &  0.92  & 1.58*  &  0.35  &  1.29  \\ 
\FV{present, 3, sg}  & 1.15*  &  0.95  &  0.71  &  1.03  \\ 
\FV{present, 1, pl}  &  0.69  &  0.32  & 1.41*  &  1.03  \\ 
\FV{present, 2, pl}  &  0.23  &  0.95  &  1.06  & 1.55*  \\ 
\FV{present, 3, pl}  &  0.46  &  0.32  & 1.41*  &  1.29  \\ \bottomrule
\end{tabular}
\end{example}
\vspace{.2in}

\vspace{.2in}

\begin{example}
\(\mathcal{B}=\left[\begin{array}{lllllll}
0.459  & 0.229  & 0.459  & 0  & 0.229  & 0.688  & 0 \\
0.316  & 0.316  & 0  & 0.632  & 0  & 0.632  & 0 \\
0.354  & 0.354  & 0.354  & 0  & 0.354  & 0  & 0.707 \\
0.258  & 0.516  & 0  & 0.516  & 0.258  & 0.258  & 0.516 \\
\end{array}\right]\)
\end{example}

We repeat here the values of $\PHI \times \BB$ in a table, with the maximum values for each row in blue.

\begin{example} \label{ex:GermanActivations}
 $\Phi \times \mathcal{B}$=
\begin{tabular}{lllll} 
suffix & -e & -st & -en & -t \\ \midrule
1st sg past &\Blue{1.606}  & 0.949  & 0.707  & 0.516 \\
2nd sg past &1.147  & \Blue{1.581}  & 0.354  & 1.033 \\
3rd sg past &\Blue{1.376}  & 0.949  & 0.707  & 0.775 \\
1st pl past &0.918  & 0.316  & \Blue{1.414}  & 0.775 \\
2nd pl past &0.459  & 0.949  & 1.061  & \Blue{1.291} \\
3rd pl past &0.688  & 0.316  & \Blue{1.414}  & 1.033 \\
1st sg pres &\Blue{1.376}  & 0.949  & 0.707  & 0.775 \\
2nd sg pres &0.918  & \Blue{1.581}  & 0.354  & 1.291 \\
3rd sg pres &\Red{1.147}  & 0.949  & 0.707  & \Green{1.033} \\
1st pl pres &0.688  & 0.316  & \Blue{1.414}  & 1.033 \\
2nd pl pres &0.229  & 0.949  & 1.061  & \Blue{1.549} \\
3rd pl pres &0.459  & 0.316  & \Blue{1.414}  & 1.291 \\
\end{tabular}
\end{example}

All the paradigm positions have the maximum value on the correct morpheme except the 3rd singular present, where the incorrect winner is marked in red. The intended winner, {\em -t} loses by 0.114.
The winner is calculated to be {\em e} (shown in red) whereas the correct suffix is {\em t} (shown in green).  What does it mean that the system did not obtain the correct results, even though in some sense we fed it all the right answers? 

Let's address a question about the learning of inflectional morphology that is quite basic, and has an impact on how we analyze anybody's theory and their account of any particular language. When we analyze a morphology, we typically do it with a full set of specifications of the paradigms of the language. Given a full set of paradigms, along with one or two examples for each grouping, we expect of our analysis that it can extend the correct results to unseen words. But we also typically expect that the analysis of the data that we are given will be simpler than (merely) repeating all of the given data. To take a simple case, when we analyze the present tense of the verb in English, we do not specify the form for each of the six person/number combinations; we typically say what  the 3rd sg present form is, and then say that all of the other present forms take a different form.  

But an account of morphology that takes learning seriously must also do something else: it must be able to provide an analysis of an {\em incomplete} set of data, making predictions about the forms that it has not yet been given. The more successful a theory is in this regard --- that is, in inferring the correct answers on the basis of incomplete data---the better the theory is.

\subsubsection{Delta rule}
	
When there is a lack of agreement between the ceorrect choice of affix and the predicted choice of affix (and at this point, such a lack of agreement must be the result of our so-called smart initialization not being quite smart enough!), we want our system to be modified automatically in order to no longer make that error. 

Weights that will determine the correct suffix for every feature-value combination can be learned through a simple gradient descent algorithm, known as the Delta Rule, which goes back to the perceptron learning rule. The intuition that lies behind it is that if we can identify an input that leads to an incorrect output (input $i$ to output $j$, let's say), we should change the weight from $i$ to $j$ in proportion to the strength of the signal that went to $i$ and in proportion to the difference between the calculated output and the desired output. 
 
We choose a small stepsize $\eta$ for each iteration of the algorithm in which we modify each suffix vector $\mu_j$ according to the following formula.

\begin{example}\label{ex:delta2}
\(\bra{\mu_j} := \bra{\mu_j} - \eta(\hat{\mu_j}  - \Phi_i \bra{\mu_j}) (-\Phi_i)^T\)
\end{example}

\noindent  where $\mu_j$ is the $j^{th}$ suffix, $\bra{\mu_j}$ is its representation in feature-value space (here, $R^7$),   $\hat{\mu}$ is the suffix's paradigm representation (here, in $R^{12}$).\footnote{The rule can be derived by calculating the partial derivative of the loss function \(L=\frac{1}{2}(\hat{\mu{j}}-\Phi_i\mu_j)^2\) with respect to $\mu_j$. A modified version of this algorithm will update a vector $\mu_j$ only when it is the incorrect choice for a given feature combination.} 
After each modification of $\mu_j$, the vector is re-normalized.
\subsection{German}
Applying the Delta Rule to the German verb suffix vectors, with a stepsize $\eta = 0.1$ and modifying the vectors only for paradigm positions that have the incorrect winner, we get correct values after just one iteration, as shown below.

\begin{example}
\(\mu=\left[\begin{array}{llll}
0.521  & 0.344  & 0.370  & 0.259 \\
0.130  & 0.241  & 0.296  & 0.515 \\
0.521  & 0  & 0.370  & 0 \\
0  & 0.687  & 0  & 0.518 \\
0.130  & -0.103  & 0.296  & 0.256 \\
0.651  & 0.584  & -0.074  & 0.256 \\
0  & 0  & 0.739  & 0.518 \\
\end{array}\right]\)
\end{example}

\bigskip

\begin{example}
\(\Phi\times\mu=\left[\begin{array}{lllll}
 & -e & -st & -en & -t \\ 
1st \ sg \ pst &\Blue{1.692}  & 0.928  & 0.665  & 0.515 \\
2nd \ sg \ pst &1.172  & \Blue{1.615}  & 0.296  & 1.033 \\
3rd \ sg \ pst &\Blue{1.302}  & 0.825  & 0.591  & 0.771 \\
1st \ pl \ pst & 1.042  & 0.344  & \Blue{1.478}  & 0.777 \\
2nd \ pl \ pst &0.521  & 1.031  & 1.109  & \Blue{1.295} \\
3rd \ pl \ pst &0.651  & 0.241  & \Blue{1.405}  & 1.033 \\
1st \ sg \ pres &\Blue{1.302}  & 0.825  & 0.591  & 0.771 \\
2nd \ sg \ pres & 0.781  & \Blue{1.512}  & 0.222  & 1.289 \\
3rd \ sg \ pres &0.911  & 0.722  & 0.517  & \Blue{1.026} \\
1st \ pl \ pres & 0.651  & 0.241  & \Blue{1.405}  & 1.033 \\
2nd \ pl \ pres &0.130  & 0.928  & 1.035  & \Blue{1.551} \\
3rd \ pl \ pres &0.260  & 0.137  & \Blue{1.331}  & 1.289 \\
\end{array}\right]\)
\end{example}
\newpage
\subsection{Latin} 
The values of $\Phi\times\mathcal{B}$ for the Latin adjective are repeated below from (\ref{ex:LatinActivations}) as (\ref{ex:LatinActivations2}).

\begin{example}\label{ex:LatinActivations2}
\(\Phi\times\mathcal{B}=\left[\begin{array}{lllllllllllllll}
&&\mathrm{us} &    \mathrm{i} &   \mathrm{o} &   \mathrm{um} &   \mathrm{e} &   \mathrm{a} &   \mathrm{ae} &   \mathrm{am} &   \mathrm{orum} &   \mathrm{is} &   \mathrm{as} &   \mathrm{arum} &   \mathrm{os} \\ 
&n &\Blue{1.732}  & 1.225  & 1.061  & 1.061  & 1.155  & 0.737  & 0.567  & 0.577  & 0.316  & 0.246  & 0  & 0  & 0.577 \\
&g& 1.155  & \Blue{1.429}  & 1.061  & 0.884  & 1.155  & 0.442  & 0.567  & 0.577  & 0.949  & 0.246  & 0  & 0.577  & 0.577 \\
MSG&d& 1.155  & 1.021  & \Blue{1.414}  & 0.884  & 1.155  & 0.442  & 0.567  & 0.577  & 0.316  & 0.615  & 0  & 0  & 0.577 \\
&ac& 1.155  & 1.021  & 1.061  & \Blue{1.237}  & 1.155  & 0.590  & 0.378  & 1.155  & 0.316  & 0.246  & 0.577  & 0  & 1.155 \\
&ab &1.155  & 1.021  & \Blue{1.414}  & 0.884  & 1.155  & 0.590  & 0.378  & 0.577  & 0.316  & 0.615  & 0  & 0  & 0.577 \\
&v&1.155  & 1.225  & 1.061  & 1.061  & \Blue{1.732}  & 0.737  & 0.567  & 0.577  & 0.316  & 0.246  & 0  & 0  & 0.577 \\
&&&&&&&&&&&&&&\\
&n&1.155  & 0.612  & 0.707  & 0.884  & 0.577  & \Green{1.180}  & \Red{1.323}  & 1.155  & 0  & 0.246  & 0.577  & 0.577  & 0 \\
&g&0.577  & 0.816  & 0.707  & 0.707  & 0.577  & 0.885  & \Blue{1.323}  & 1.155  & 0.632  & 0.246  & 0.577  & 1.155  & 0 \\
FSG&d&0.577  & 0.408  & 1.061  & 0.707  & 0.577  & 0.885  & \Blue{1.323}  & 1.155  & 0  & 0.615  & 0.577  & 0.577  & 0 \\
&ac&0.577  & 0.408  & 0.707  & 1.061  & 0.577  & 1.032  & 1.134  & \Blue{1.732}  & 0  & 0.246  & 1.155  & 0.577  & 0.577 \\
&ab&0.577  & 0.408  & 1.061  & 0.707  & 0.577  & 1.032  & \Green{1.134}  & \Red{1.155}  & 0  & 0.615  & 0.577  & 0.577  & 0 \\ 
&v&0.577  & 0.612  & 0.707  & 0.884  & 1.155  & 1.180  & \Blue{1.323}  & 1.155  & 0  & 0.246  & 0.577  & 0.577  & 0 \\
&&&&&&&&&&&&&&\\
&n&1.155  & 0.816  & 1.061  & \Blue{1.414}  & 0.577  & 1.180  & 0.567  & 0.577  & 0.316  & 0.246  & 0  & 0  & 0 \\
&g&0.577  & \Green{1.021}  & 1.061  & \Red{1.237}  & 0.577  & 0.885  & 0.567  & 0.577  & 0.949  & 0.246  & 0  & 0.577  & 0 \\
NSG&d&0.577  & 0.612  & \Blue{1.414}  & 1.237  & 0.577  & 0.885  & 0.567  & 0.577  & 0.316  & 0.615  & 0  & 0  & 0 \\
&ac&0.577  & 0.612  & 1.061  & \Blue{1.591}  & 0.577  & 1.032  & 0.378  & 1.155  & 0.316  & 0.246  & 0.577  & 0  & 0.577 \\
&ab&0.577  & 0.612  & \Blue{1.414}  & 1.237  & 0.577  & 1.032  & 0.378  & 0.577  & 0.316  & 0.615  & 0  & 0  & 0 \\
&v&0.577  & 0.816  & 1.061  & \Blue{1.414}  & 1.155  & 1.180  & 0.567  & 0.577  & 0.316  & 0.246  & 0  & 0  & 0 \\
&&&&&&&&&&&&&&\\
&n&1.155  & \Blue{1.225}  & 0.354  & 0.354  & 0.577  & 0.737  & 0.567  & 0  & 0.949  & 0.985  & 0.577  & 0.577  & 1.155 \\
&g&0.577  & 1.429  & 0.354  & 0.177  & 0.577  & 0.442  & 0.567  & 0  & \Blue{1.581}  & 0.985  & 0.577  & 1.155  & 1.155 \\
MPL&d&0.577  & 1.021  & 0.707  & 0.177  & 0.577  & 0.442  & 0.567  & 0  & 0.949  & \Blue{1.354}  & 0.577  & 0.577  & 1.155 \\
&ac&0.577  & 1.021  & 0.354  & 0.530  & 0.577  & 0.590  & 0.378  & 0.577  & 0.949  & 0.985  & 1.155  & 0.577  & \Blue{1.732} \\
&ab&0.577  & 1.021  & 0.707  & 0.177  & 0.577  & 0.590  & 0.378  & 0  & 0.949  & \Blue{1.354}  & 0.577  & 0.577  & 1.155 \\
&v&0.577  & \Blue{1.225}  & 0.354  & 0.354  & 1.155  & 0.737  & 0.567  & 0  & 0.949  & 0.985  & 0.577  & 0.577  & 1.155 \\
&&&&&&&&&&&&&&\\
&n&0.577  & 0.612  & 0  & 0.177  & 0  & 1.180  & \Blue{1.323}  & 0.577  & 0.632  & 0.985  & 1.155  & 1.155  & 0.577 \\
&g&0  & 0.816  & 0  & 0  & 0  & 0.885  & 1.323  & 0.577  & 1.265  & 0.985  & 1.155  & \Blue{1.732}  & 0.577 \\
FPL&d&0  & 0.408  & 0.354  & 0  & 0  & 0.885  & 1.323  & 0.577  & 0.632  & \Blue{1.354}  & 1.155  & 1.155  & 0.577 \\
&ac&0  & 0.408  & 0  & 0.354  & 0  & 1.032  & 1.134  & 1.155  & 0.632  & 0.985  & \Blue{1.732}  & 1.155  & 1.155 \\
&ab&0  & 0.408  & 0.354  & 0  & 0  & 1.032  & 1.134  & 0.577  & 0.632  & \Blue{1.354}  & 1.155  & 1.155  & 0.577 \\
&v&0  & 0.612  & 0  & 0.177  & 0.577  & 1.180  & \Blue{1.323}  & 0.577  & 0.632  & 0.985  & 1.155  & 1.155  & 0.577 \\
&&&&&&&&&&&&&&\\
&n&0.577  & 0.816  & 0.354  & 0.707  & 0  & \Blue{1.180}  & 0.567  & 0  & 0.949  & 0.985  & 0.577  & 0.577  & 0.577 \\
&g&0  & 1.021  & 0.354  & 0.530  & 0  & 0.885  & 0.567  & 0  & \Blue{1.581}  & 0.985  & 0.577  & 1.155  & 0.577 \\
NPL&d&0  & 0.612  & 0.707  & 0.530  & 0  & 0.885  & 0.567  & 0  & 0.949  & \Blue{1.354}  & 0.577  & 0.577  & 0.577 \\
&ac&0  & 0.612  & 0.354  & 0.884  & 0  & \Green{1.032}  & 0.378  & 0.577  & 0.949  & 0.985  & \Red{1.155}  & 0.577  & 1.155 \\
&ab&0  & 0.612  & 0.707  & 0.530  & 0  & 1.032  & 0.378  & 0  & 0.949  & \Blue{1.354}  & 0.577  & 0.577  & 0.577 \\
&v&0  & 0.816  & 0.354  & 0.707  & 0.577  & \Blue{1.180}  & 0.567  & 0  & 0.949  & 0.985  & 0.577  & 0.577  & 0.577 \\
\end{array}\right]\)
\end{example}
 
Recall that there were four positions (shown above in red) in which smart initialization chose the wrong form. 

In six iterations of the Delta Rule with stepsize $\eta=0.1$ we end up with the correct forms for all 36 positions. Here are the final values of suffix vectors:

\newpage
\begin{example}
\(\mathcal{B}=\left[\begin{array}{rrrrrrrrrrrrr}
\mathrm{us} &    \mathrm{i} &   \mathrm{o} &   \mathrm{um} &   \mathrm{e} &   \mathrm{a} &   \mathrm{ae} &   \mathrm{am} &   \mathrm{orum} &   \mathrm{is} &   \mathrm{as} &   \mathrm{arum} &   \mathrm{os} \\ 

0.48  & 0.33  & 0.44  & 0.59  & 0.39  & 0.36  & 0.10  & 0.31  & -0.25  & -0.16  & -0.23  & -0.27  & -0.16 \\
-0.26  & 0.20  & -0.14  & -0.19  & -0.17  & 0.28  & 0.23  & -0.11  & 0.40  & 0.51  & 0.30  & 0.29  & 0.21 \\
0.54  & 0.59  & 0.35  & 0.25  & 0.52  & -0.25  & -0.25  & -0.02  & 0.18  & 0.04  & -0.19  & -0.21  & 0.47 \\
-0.23  & -0.26  & -0.32  & -0.27  & -0.23  & 0.43  & 0.78  & 0.45  & -0.12  & 0.18  & 0.51  & 0.44  & -0.10 \\
-0.08  & 0.20  & 0.27  & 0.43  & -0.06  & 0.45  & -0.20  & -0.23  & 0.08  & 0.14  & -0.24  & -0.21  & -0.32 \\
0.55  & 0.19  & -0.15  & 0.11  & -0.21  & 0.27  & 0.03  & -0.13  & -0.21  & -0.24  & -0.16  & -0.18  & -0.30 \\
-0.15  & 0.48  & -0.21  & -0.28  & -0.14  & -0.19  & 0.35  & -0.05  & 0.79  & -0.09  & 0.04  & 0.67  & -0.06 \\
0.10  & -0.05  & 0.50  & 0.10  & 0.10  & -0.18  & 0.09  & -0.08  & -0.12  & 0.52  & -0.24  & -0.23  & -0.10 \\
-0.10  & -0.27  & -0.13  & 0.39  & -0.08  & 0.18  & -0.07  & 0.75  & -0.19  & -0.22  & 0.63  & -0.05  & 0.68 \\
-0.07  & -0.05  & 0.39  & -0.10  & -0.08  & 0.22  & -0.23  & -0.17  & 0  & 0.51  & -0.08  & -0.07  & 0 \\
-0.12  & 0.23  & -0.09  & 0.18  & 0.64  & 0.33  & 0.17  & -0.12  & -0.12  & -0.13  & -0.12  & -0.12  & -0.17 \\
\end{array}\right]\)
\end{example}

\medskip

And here are the final values for suffixes projected onto paradigm positions.

\medskip
\begin{example}
\(\Phi\times\mathcal{B}=\left[\begin{array}{rrrrrrrrrrrrr}
\mathrm{us} &    \mathrm{i} &   \mathrm{o} &   \mathrm{um} &   \mathrm{e} &   \mathrm{a} &   \mathrm{ae} &   \mathrm{am} &   \mathrm{orum} &   \mathrm{is} &   \mathrm{as} &   \mathrm{arum} &   \mathrm{os} \\ 
\Blue{1.56}  & 1.10  & 0.64  & 0.95  & 0.70  & 0.38  & -0.12  & 0.15  & -0.28  & -0.36  & -0.58  & -0.65  & 0.02 \\
0.87  & \Blue{1.40}  & 0.58  & 0.56  & 0.78  & -0.09  & 0.20  & 0.23  & 0.72  & -0.21  & -0.38  & 0.20  & 0.25 \\
1.11  & 0.87  & \Blue{1.29}  & 0.94  & 1.01  & -0.08  & -0.05  & 0.20  & -0.18  & 0.41  & -0.66  & -0.70  & 0.22 \\
0.92  & 0.65  & 0.65  & \Blue{1.23}  & 0.83  & 0.29  & -0.22  & 1.04  & -0.26  & -0.34  & 0.21  & -0.53  & 1.00 \\
0.95  & 0.87  & \Blue{1.17}  & 0.74  & 0.84  & 0.33  & -0.38  & 0.12  & -0.07  & 0.39  & -0.50  & -0.55  & 0.31 \\
0.90  & 1.14  & 0.70  & 1.02  & \Blue{1.55}  & 0.44  & 0.03  & 0.17  & -0.19  & -0.25  & -0.54  & -0.59  & 0.14 \\
0.79  & 0.26  & -0.02  & 0.43  & -0.05  & \Blue{1.06}  & 0.91  & 0.63  & -0.58  & -0.22  & 0.11  & -0.01  & -0.55 \\
0.10  & 0.55  & -0.09  & 0.04  & 0.03  & 0.59  & \Blue{1.23}  & 0.71  & 0.42  & -0.07  & 0.32  & 0.85  & -0.32 \\
0.34  & 0.02  & 0.63  & 0.42  & 0.26  & 0.60  & \Blue{0.98}  & 0.68  & -0.48  & 0.55  & 0.04  & -0.05  & -0.35 \\
0.15  & -0.20  & -0.01  & 0.71  & 0.08  & 0.97  & 0.81  & \Blue{1.51}  & -0.56  & -0.20  & 0.91  & 0.12  & 0.43 \\
0.18  & 0.03  & 0.51  & 0.22  & 0.09  & \Blue{1.01}  & 0.65  & 0.59  & -0.37  & 0.53  & 0.20  & 0.10  & -0.25 \\
0.13  & 0.30  & 0.03  & 0.50  & 0.80  & \Blue{1.12}  & 1.06  & 0.65  & -0.49  & -0.11  & 0.16  & 0.06  & -0.43 \\
0.95  & 0.71  & 0.56  & \Blue{1.14}  & 0.12  & 1.08  & -0.07  & -0.05  & -0.39  & -0.26  & -0.63  & -0.66  & -0.78 \\
0.25  & \Blue{1.01}  & 0.50  & 0.75  & 0.19  & 0.62  & 0.24  & 0.03  & 0.61  & -0.11  & -0.43  & 0.19  & -0.54 \\
0.50  & 0.48  & \Blue{1.21}  & 1.12  & 0.42  & 0.63  & -0.01  & -0.00  & -0.29  & 0.51  & -0.71  & -0.71  & -0.58 \\
0.30  & 0.26  & 0.58  & \Blue{1.41}  & 0.25  & 0.99  & -0.18  & 0.83  & -0.36  & -0.24  & 0.17  & -0.53  & 0.20 \\
0.33  & 0.48  & \Blue{1.10}  & 0.93  & 0.25  & 1.03  & -0.33  & -0.09  & -0.17  & 0.49  & -0.55  & -0.55  & -0.48 \\
0.28  & 0.75  & 0.62  & \Blue{1.21}  & 0.96  & 1.14  & 0.07  & -0.04  & -0.29  & -0.15  & -0.59  & -0.60  & -0.65 \\
0.83  & \Blue{0.98}  & 0.06  & 0.17  & 0.14  & 0.30  & 0.02  & -0.27  & 0.37  & 0.30  & -0.06  & -0.09  & 0.38 \\
0.13  & 1.28  & -0.00  & -0.22  & 0.22  & -0.16  & 0.34  & -0.19  & \Blue{1.37}  & 0.45  & 0.15  & 0.76  & 0.61 \\
0.38  & 0.75  & 0.71  & 0.16  & 0.45  & -0.15  & 0.08  & -0.22  & 0.47  & \Blue{1.07}  & -0.13  & -0.14  & 0.58 \\
0.18  & 0.52  & 0.07  & 0.45  & 0.27  & 0.21  & -0.09  & 0.62  & 0.39  & 0.33  & 0.74  & 0.04  & \Blue{1.36} \\
0.22  & 0.75  & 0.59  & -0.04  & 0.28  & 0.25  & -0.24  & -0.30  & 0.58  & \Blue{1.06}  & 0.03  & 0.02  & 0.68 \\
0.16  & \Blue{1.02}  & 0.12  & 0.24  & 0.99  & 0.36  & 0.16  & -0.25  & 0.46  & 0.42  & -0.02  & -0.03  & 0.50 \\
0.06  & 0.13  & -0.60  & -0.34  & -0.61  & 0.98  & \Blue{1.05}  & 0.21  & 0.07  & 0.44  & 0.64  & 0.56  & -0.19 \\
-0.64  & 0.43  & -0.67  & -0.73  & -0.53  & 0.52  & 1.36  & 0.29  & 1.07  & 0.59  & 0.84  & \Blue{1.41}  & 0.05 \\
-0.39  & -0.10  & 0.05  & -0.36  & -0.30  & 0.53  & 1.11  & 0.26  & 0.17  & \Blue{1.21}  & 0.57  & 0.51  & 0.01 \\
-0.59  & -0.33  & -0.59  & -0.07  & -0.48  & 0.89  & 0.94  & 1.10  & 0.09  & 0.47  & \Blue{1.44}  & 0.69  & 0.79 \\
-0.55  & -0.10  & -0.07  & -0.55  & -0.47  & 0.93  & 0.79  & 0.18  & 0.28  & \Blue{1.19}  & 0.73  & 0.67  & 0.11 \\
-0.61  & 0.17  & -0.55  & -0.27  & 0.24  & 1.05  & \Blue{1.19}  & 0.23  & 0.16  & 0.55  & 0.68  & 0.62  & -0.07 \\
0.21  & 0.59  & -0.02  & 0.36  & -0.44  & \Blue{1.01}  & 0.06  & -0.47  & 0.26  & 0.40  & -0.11  & -0.09  & -0.41 \\
-0.48  & 0.88  & -0.08  & -0.03  & -0.37  & 0.54  & 0.38  & -0.39  & \Blue{1.26}  & 0.55  & 0.10  & 0.76  & -0.18 \\
-0.23  & 0.35  & 0.63  & 0.35  & -0.13  & 0.55  & 0.12  & -0.42  & 0.36  & \Blue{1.17}  & -0.18  & -0.14  & -0.21 \\
-0.43  & 0.13  & -0.00  & 0.64  & -0.31  & \Blue{0.92}  & -0.05  & 0.41  & 0.29  & 0.43  & 0.69  & 0.03  & 0.57 \\
-0.40  & 0.36  & 0.52  & 0.15  & -0.31  & 0.95  & -0.20  & -0.51  & 0.48  & \Blue{1.15}  & -0.02  & 0.01  & -0.12 \\
-0.45  & 0.63  & 0.04  & 0.43  & 0.40  & \Blue{1.07}  & 0.20  & -0.46  & 0.36  & 0.52  & -0.06  & -0.03  & -0.29 \\
\end{array}\right]\)
\end{example}

To summarize this section: the Delta Rule is an error-driven machine learning tool that enables us to simulate the way a learner might proceed beyond smart initialization of feature values for a set of exponents in a paradigm to arrive at a set of values that will find the correct morpheme for each morphosyntactic position. We next look at ways in which our model can deal with the way stems and affixes interact in cases where a learner needs to select both the correct stem and the correct affix for a given paradigm position.

\section{Morpheme concatenation as vector sum}
 
We have so far considered only the selection of a single morpheme in all of our computations, but in the more general case, we wish to be able to select multiple morphemes---which is to say, the correct form in a word paradigm may be polymorphemic. For example, if a stem morpheme has two allomorphs, and a suffix must also be chosen, then it is a {\em pair}, the stem and suffix, which must be selected. The central notion of geometrical morphology provides a prediction for these cases: the (stem, affix) pair provides a vector sum (the vector sum of the vector representing the stem and the vector representing the affix), and that pair is chosen whose vector sum is closest to the target position M. If a stem $\mu_t$ is realized from a set of allomorphs $\mathcal{T}$ and its suffix is selected from a set of inflectional suffixes $\mathcal{F}$, then to realize the position M in its inflectional paradigm, the morphology selects one stem allomorph $\hat{\mu}_{\text{stem}}$ in $\mathcal{T}$ and one affix $\hat{\mu}_{\text{affix}}$ in  $\mathcal{F}$: 

\[  
(\hat{\mu}_{\text{stem}}, \hat{\mu}_{\text{affix}}) = 
arg min_{\mu_{\text{stem}}  \in \mathcal{T},\mu_{\text{affix}} \in \mathcal{F} } \,  
distance (M, 
    \vec{\mu}_{\text{stem}} + \vec{\mu}_{\text{affix}} )
    \]

This characteristic allows the geometric morphological model to have its cake and eat it too, so to speak: the model is equally comfortable with a paradigmatic position being realized as a single morpheme or as a set of morphemes. The simplicity of concatenation is permitted, but not required.

\subsection{German plurals}
We shall first illustrate this with one of the simplest possible examples: the plural suffix in German. German has a number of different plural suffixes whose choice depends on the stem with which the suffix occurs. The singular is unmarked, so we shall continue to indicate that suffix as $\emptyset$, remaining agnostic about whether this is a null morpheme or simply the lack of a morpheme. The following are some examples of plural suffixes in German.\footnote{Where umlaut is part of the suffix, we analyze umlaut as a floating $[-back]$ feature, and abstract away here from the question of exactly how and where this feature is realized. Following tradition, we write this as a typographically floating diaresis.}

\bigskip

\begin{table}[ht]
\centering
\begin{tabular}{llll}
Noun & Pl.\ suffix & Pl.\ form& Gloss \\ \hline
Kind & er & Kinder & \Gl{child}\\
Glas & \"{} + er & Gl\"aser & \Gl{glass}\\
Fenster & $\emptyset$ & Fenster & \Gl{window}\\
Mutter & \"{} & M\"utter & \Gl{mother}\\
Auto & s & Autos & \Gl{automobile}\\
\end{tabular}
\caption{Some German plurals}
\end{table}
 
\bigskip

We only have two feature values in feature-value space: singular and plural. In this space, all the stems $\mathcal{T}$ and affixes $\mathcal{F}$ are represented by two dimensional vectors. Let us visualize the \FV{singular} dimension as the vertical axis and the \FV{plural} dimension as the horizontal axis. Since all the vectors are of unit length, we can specify any morpheme simply by an angle $\theta_i$, the angle between the position of the vector and the plural axis. For a given choice of noun stem, its  plural is the affix for which the vector sum of the stem vector and the affix vector are closest to, i.e., make the smallest angle with, the plural axis.  Similarly, the affix that occurs with a given stem on the singular will be the one for which the angle of the stem plus affix is closest to $\frac{\pi}{2}$ or $90^o$.

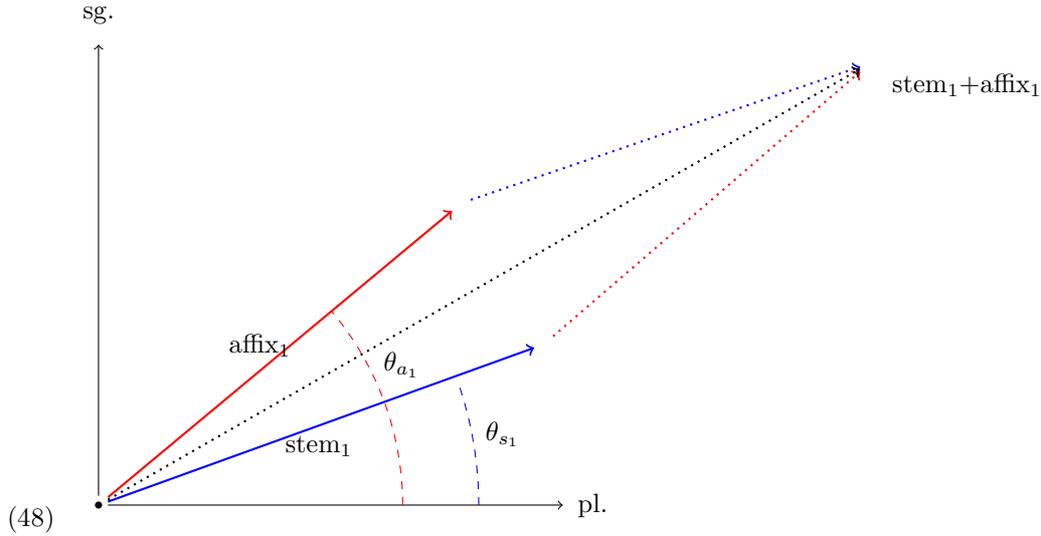
\bigskip

\begin{example}
\begin{tikzpicture}
\filldraw[fill=black] (0,0) circle (0.04cm);
\node (origin) at (0,0) {};
\node (pl) at (canvas polar cs:angle=0, radius=6.23cm) {};
\node (sg) at (canvas polar cs:angle=90, radius=6.23cm) {};
\node (plabel) at (canvas polar cs:angle=0, radius=6.5cm) {pl.};
\node (slabel) at (canvas polar cs:angle=90, radius=6.5cm) {sg.};
\path[->] (origin) edge (pl);
\path[->] (origin) edge (sg);
\node (stem1) at (canvas polar cs:angle=20, radius=6.23cm) {};
\node (aff1) at (canvas polar cs:angle=40, radius=6.23cm) {};
\path[->, blue, thick] (origin) edge (stem1);
\path[->, red, thick] (origin) edge (aff1);
\node (s1label) at (canvas polar cs:angle=15, radius=3cm) {stem$_1$};
\node (a1label) at (canvas polar cs:angle=45, radius=3cm) {affix$_1$};
\node (sum) at (canvas polar cs:angle=30, radius=11.71cm) {};
\path[->, blue, thick, dotted] (aff1) edge (sum);
\path[->, red, thick, dotted] (stem1) edge (sum);
\path[->, black, thick, dotted] (origin) edge (sum);
\node (sumlabel) at (canvas polar cs:angle=26, radius=12.71cm) {stem$_1$+affix$_1$};
\draw[blue, dashed] (5, 0) arc (0:19:5);
\draw[red, dashed] (4, 0) arc (0:40:4);
\node (th1label) at (canvas polar cs:angle=10, radius=5.4cm) {$\theta_{s_1}$};
\node (af1label) at (canvas polar cs:angle=25, radius=4.4cm) {$\theta_{a_1}$};
\end{tikzpicture}
\end{example}
\bigskip

Our model can learn feature values for stems and affixes so that the correct affixes will occur with the correct stems through the following algorithm.
\medskip

\begin{enumerate}
\item Initialize feature values for the stems and affixes as randomly chosen angles between $\frac{\pi}{2}$ and $-\frac{\pi}{2}$.
\item Set the stepsize used in learning to 0.01 radians, and set a small margin of required separation, such as 0.05 radians.
\item Repeat each of the following steps for each of a series of iterations until no adjustments need to be made.
\begin{enumerate}
\item For each stem:
\begin{enumerate}
\item For each affix:
\begin{enumerate}
\item If that combination is correct for the singular: if any other affix with the same stem results in an angle closer to $90^o$ than this combination does, move both the stem and the correct affix closer to $90^o$ and the incorrect one farther from $90^o$, each by the stepsize.
\item Do as in (A) for the plural forms, though with respect to the angle $0^o$.
\end{enumerate}
\end{enumerate}
\end{enumerate}
\end{enumerate}

With a stepsize of $\eta=0.01$ and margin $\epsilon=0.05$ one run found, after 22 iterations, the following feature values  expressed as angles in degrees from the plural axis.

\bigskip

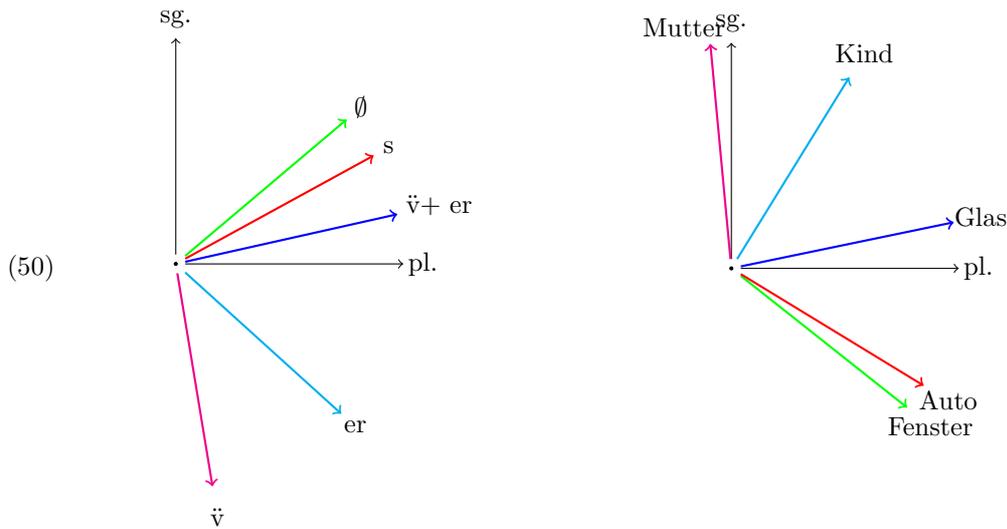
\begin{example}
\begin{tabular}{llll}
Stem or affix &  Angle & \FV{plural} & \FV{singular} \\ \hline

Fenster      & -38.593 & 0.782 & -0.624 \\ 
Auto         & -31.568 & 0.852 & -0.524 \\ 
Glas         & 11.800 & 0.979 & 0.204 \\ 
Kind         & 58.535 & 0.522 & 0.853 \\ 
Mutter       & 95.310 & -0.093 & 0.996 \\ 
$\emptyset$  & 40.430 & 0.761 & 0.649 \\ 
s            & 28.909 & 0.875 & 0.483 \\ 
\"{}-er & 12.729 & 0.975 & 0.220 \\ 
er           & -42.441 & 0.738 & -0.675 \\ 
\"{}     & -80.694 & 0.162 & -0.987 \\ 
\end{tabular}
\end{example}
\bigskip

\begin{example}
\begin{minipage}{.4\textwidth}
\centering%
\begin{tikzpicture}[scale=0.5]
\filldraw[fill=black] (0,0) circle (0.04cm);
\node (origin) at (0,0) {};
\node (pl) at (canvas polar cs:angle=0, radius=6.23cm) {};
\node (sg) at (canvas polar cs:angle=90, radius=6.23cm) {};
\node (plabel) at (canvas polar cs:angle=0, radius=6.5cm) {pl.};
\node (slabel) at (canvas polar cs:angle=90, radius=6.5cm) {sg.};
\path[->] (origin) edge (pl);
\path[->] (origin) edge (sg);
\node (null) at (canvas polar cs:angle=40.43, radius=6.23cm) {};
\node (nulllabel) at (canvas polar cs:angle=40.43, radius=6.4cm) {$\emptyset$};
\path[->, green, thick] (origin) edge (null);
\node (s) at (canvas polar cs:angle=28.9, radius=6.23cm) {};
\node (slabel) at (canvas polar cs:angle=28.9, radius=6.4cm) {s};
\path[->, red, thick] (origin) edge (s);
\node (uer) at (canvas polar cs:angle=12.73, radius=6.23cm) {};
\node (uerlabel) at (canvas polar cs:angle=12.73, radius=7.1cm) {\"{v}+ er};
\path[->, blue, thick] (origin) edge (uer);
\node (er) at (canvas polar cs:angle=-42.4, radius=6.23cm) {};
\node (erlabel) at (canvas polar cs:angle=-42.4, radius=6.4cm) {er};
\path[->, cyan, thick] (origin) edge (er);
\node (u) at (canvas polar cs:angle=-80.7, radius=6.23cm) {};
\node (ulabel) at (canvas polar cs:angle=-80.7, radius=6.8cm) {\"{v}};
\path[->, magenta, thick] (origin) edge (u);
\end{tikzpicture}%
\end{minipage}%
\hspace{1cm}%
\begin{minipage}{.4\textwidth}
\begin{tikzpicture}[scale=0.5]
\filldraw[fill=black] (0,0) circle (0.04cm);
\node (origin) at (0,0) {};
\node (pl) at (canvas polar cs:angle=0, radius=6.23cm) {};
\node (sg) at (canvas polar cs:angle=90, radius=6.23cm) {};
\node (plabel) at (canvas polar cs:angle=0, radius=6.5cm) {pl.};
\node (slabel) at (canvas polar cs:angle=90, radius=6.5cm) {sg.};
\path[->] (origin) edge (pl);
\path[->] (origin) edge (sg);
\node (mutter) at (canvas polar cs:angle=95.31, radius=6.23cm) {};
\node (mutterlabel) at (canvas polar cs:angle=101, radius=6.5cm) {Mutter};
\path[->, magenta, thick] (origin) edge (mutter);
\node (fenster) at (canvas polar cs:angle=-38.6, radius=6.23cm) {};
\node (fensterlabel) at (canvas polar cs:angle=-38.6, radius=6.7cm) {Fenster};
\path[->, green, thick] (origin) edge (fenster);
\node (auto) at (canvas polar cs:angle=-31.57, radius=6.23cm) {};
\node (autolabel) at (canvas polar cs:angle=-31.57, radius=6.7cm) {Auto};
\path[->, red, thick] (origin) edge (auto);
\node (glas) at (canvas polar cs:angle=11.8, radius=6.23cm) {};
\node (glaslabel) at (canvas polar cs:angle=11.8, radius=6.7cm) {Glas};
\path[->, blue, thick] (origin) edge (glas);
\node (kind) at (canvas polar cs:angle=58.54, radius=6.23cm) {};
\node (kindlabel) at (canvas polar cs:angle=58.54, radius=6.7cm) {Kind};
\path[->, cyan, thick] (origin) edge (kind);
\node (ulabel) at (canvas polar cs:angle=-80.7, radius=6.8cm) {}; 
\end{tikzpicture}
\end{minipage}
\end{example}

It is noteworthy that the vector for $\emptyset$ has ended up closer to the singular axis than any other affix. This ensures that it will be the affix that occurs with each stem in the singular. And because stem {\em Fenster} is the only one that occurs with the affix $\emptyset$ in the plural as well as the singular, it must have the smallest value (in this case negative) value for singular among all the stems, so that when combined with affix $\emptyset$, it will be closer to the plural axis than when it  combines with any other affixes.

\bigskip
It is important to note that when stems come from different lexemes, as is the case here, we take it that the learner has other ways of determining which lexeme (and therefore which stem) is the one in question. It will be the stem the chooses the correct affix for a given morphosyntactic combination; not the affix that chooses the stem. To put the matter simply, the choice of stem is extragrammatical, and the choice of affix is grammatical.

\bigskip

For ease of exposition, we show how just one stem chooses the correct affix for plural. We can see that the vector sum {\em Auto} + {\em s} comes closest to having its endpoint on the plural axis. The diagram on the left shows the stem vector and all of the affixes that it chooses among, while on the right we see how the two vectors, one for  {\em Auto} and one for the suffix {\em  s} sum to be closest to the plural axis.

 \bigskip

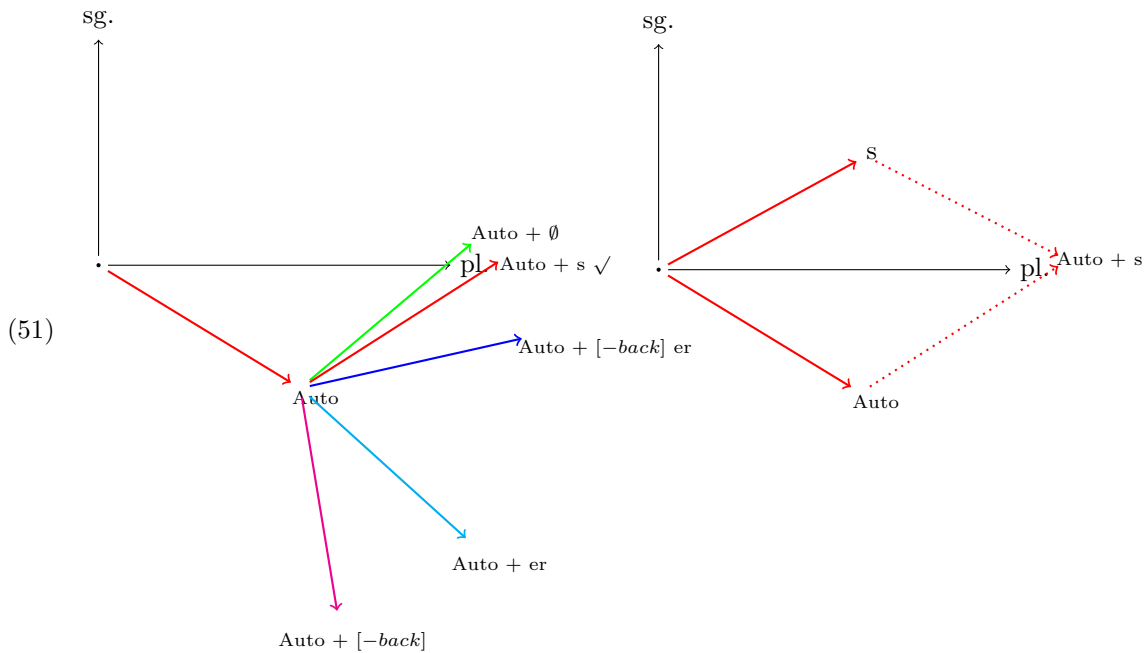
\begin{example}
\begin{minipage}{.4\textwidth}
\centering%
\begin{tikzpicture}[scale=0.5]
\filldraw[fill=black] (0,0) circle (0.04cm);
\node (origin) at (0,0) {};
\node (pl) at (canvas polar cs:angle=0, radius=9.5cm) {};
\node (sg) at (canvas polar cs:angle=90, radius=6.23cm) {};
\node (plabel) at (canvas polar cs:angle=0, radius=9.9cm) {pl.};
\node (slabel) at (canvas polar cs:angle=90, radius=6.5cm) {sg.};
\path[->] (origin) edge (pl);
\path[->] (origin) edge (sg);
\node (auto) at (canvas polar cs:angle=-31.57, radius=6.23cm) {};
\node (autolabel) at (canvas polar cs:angle=-31.57, radius=6.7cm) {\scriptsize Auto};
\path[->, red, thick] (origin) edge (auto);
\node (null) at (canvas polar cs:angle=4.43, radius=10.083cm) {};
\node (nulllabel) at (canvas polar cs:angle=4.43, radius=11cm) {\scriptsize Auto + $\emptyset$};
\path[->, green, thick] (auto) edge (null);
\node (s) at (canvas polar cs:angle=1.33, radius=10.765cm) {};
\node (slabel) at (canvas polar cs:angle=0, radius=12.0cm) {\scriptsize Auto + s $\surd$};
\path[->, red, thick] (auto) edge (s);
\node (uer) at (canvas polar cs:angle=-9.42, radius=11.54cm) {};
\node (uerlabel) at (canvas polar cs:angle=-9.42, radius=13.5cm) {\scriptsize Auto + $[-back]$ er};
\path[->, blue, thick] (auto) edge (uer);
\node (er) at (canvas polar cs:angle=-36.984, radius=12.403cm) {};
\node (erlabel) at (canvas polar cs:angle=-36.98, radius=13.2cm) {\scriptsize Auto + er};
\path[->, cyan, thick] (auto) edge (er);
\node (u) at (canvas polar cs:angle=-56.137, radius=11.332cm) {};
\node (ulabel) at (canvas polar cs:angle=-56.13, radius=12cm) {\scriptsize Auto + $[-back]$};
\path[->, magenta, thick] (auto) edge (u);
\end{tikzpicture}
\end{minipage}%
\hspace{1cm}%
\begin{minipage}{.4\textwidth}
\begin{tikzpicture}[scale=0.5]
\filldraw[fill=black] (0,0) circle (0.04cm);
\node (origin) at (0,0) {};
\node (pl) at (canvas polar cs:angle=0, radius=9.5cm) {};
\node (sg) at (canvas polar cs:angle=90, radius=6.23cm) {};
\node (plabel) at (canvas polar cs:angle=0, radius=9.9cm) {pl.};
\node (slabel) at (canvas polar cs:angle=90, radius=6.5cm) {sg.};
\path[->] (origin) edge (pl);
\path[->] (origin) edge (sg);
\node (auto) at (canvas polar cs:angle=-31.57, radius=6.23cm) {};
\node (autolabel) at (canvas polar cs:angle=-31.57, radius=6.7cm) {\scriptsize Auto};
\path[->, red, thick] (origin) edge (auto);
\node (s) at (canvas polar cs:angle=28.9, radius=6.23cm) {};
\node (slabel) at (canvas polar cs:angle=28.9, radius=6.4cm) {s};
\path[->, red, thick] (origin) edge (s);
\node (autos) at (canvas polar cs:angle=1.33, radius=10.765cm) {};
\node (autoslabel) at (canvas polar cs:angle=1.33, radius=11.6cm) {\scriptsize Auto + s};
\path[->, red, thick, dotted] (auto) edge (autos);
\path[->, red, thick, dotted] (s) edge (autos);
\node (u) at (canvas polar cs:angle=-56.137, radius=12cm) {}; 
\end{tikzpicture}
\end{minipage}%
\end{example}

\subsection{Spanish verbal classes}
\label{sec:spanish}
Spanish, like other Romance languages, is often treated as having three inflectional classes for verbs, labeled by their infinitival suffix: -{\em ar} verbs, -{\em er} verbs, and -{\em ir} verbs. The three different vowels of these suffixes may well be analyzed as a {\em theme vowel}, a morpheme separate from the root and from the inflectional affixes. But for purposes of illustrating vector geometry, we will analyze Spanish more simply, and adopt the analysis that there are three distinct inflectional classes among the verbs, and each class selects its own set of inflectional suffixes.\footnote{If we analyzed the system with a third morpheme, a theme vowel between the stem and the suffix, the system would be slightly more complex, but would not look very different.}

\medskip
\begin{example}
\begin{minipage}{.4\textwidth}
\begin{tabular}{lll}\hline
Class -ar & \Gl{sing} & \Gl{call} \\
infinitive: & cantar & llamar \\ \hline
\FV{1st sg.}  & canto & llamo \\
\FV{2nd sg.}  & cantas & llamas \\
\FV{3rd sg.}  & canta & llama \\
\FV{1st pl.}  & cantamos & llamamos \\
\FV{2nd pl.}  & cant\'ais & llam\'ais \\
\FV{3rd pl.}  & cantan & llaman \\ \hline
\end{tabular}
\end{minipage}
\begin{minipage}{.3\textwidth} 
\begin{tabular}{lll}\hline
Class -er & \Gl{eat} & \Gl{fear} \\
infinitive & comer & temer \\ \hline
\FV{1st sg.}  & como & temo \\
\FV{2nd sg.}  & comes & temes \\
\FV{3rd sg.}  & come & teme \\
\FV{1st pl.}  & comemos & tememos \\
\FV{2nd pl.}  & com\'eis & tem\'eis \\
\FV{3rd pl.}  & comen & temen \\ \hline
\end{tabular}
\end{minipage}
\medskip

\begin{tabular}{lll}\hline
Class -ir & \Gl{open} & \Gl{live} \\
infinitive & abrir & vivir \\ \hline
\FV{1st sg.}  & abro & vivo \\
\FV{2nd sg.}  & abres & vives \\
\FV{3rd sg.}  & abre & vive \\
\FV{1st pl.}  & abrimos & vivimos \\
\FV{2nd pl.}  & abr\'is & viv\'is \\
\FV{3rd pl.}  & abren & viven \\ \hline
\end{tabular}
\end{example}
\bigskip

From these data we can abstract the following inflectional classes of suffixes:

\bigskip

\medskip

\begin{example}
\begin{minipage}{.3\textwidth} 
\begin{tabular}{ll}\toprule
\multicolumn{2}{c}{Class 1} \\
infinitive & ar \\ \midrule
1st sg. & o \\
2nd sg. & as \\
3rd sg. & a \\
1st pl. & amos \\
2nd pl. & \'ais \\
3rd pl. & an \\ \hline
\end{tabular}
\end{minipage}
\begin{minipage}{.3\textwidth} 
\begin{tabular}{ll}\toprule
\multicolumn{2}{c}{Class 2} \\
infinitive & er \\ \midrule
1st sg. & o \\
2nd sg. & es \\
3rd sg. & e \\
1st pl. & emos \\
2nd pl. & \'eis \\
3rd pl. & en \\ \hline
\end{tabular}
\end{minipage}
\begin{minipage}{.3\textwidth}  
\begin{tabular}{ll}\toprule
\multicolumn{2}{c}{Class 3} \\
infinitive & ir \\ \midrule
1st sg. & o \\
2nd sg. & es \\
3rd sg. & e \\
1st pl. & imos \\
2nd pl. & \'is \\
3rd pl. & en \\ \hline
\end{tabular}
\end{minipage}
\end{example}

\bigskip

Let's consider the selection of the morpheme {\em -as} for the {\em ar}-class 2nd person singular form, and of {\em -es} for the {\em -er}-class 2nd person singular form. Smart initialization is sketched in in (\ref{spanish-verb-FVs}): 

\begin{example}
\label{spanish-verb-FVs}
\begin{minipage}{.4\textwidth} 
\begin{tabular}{llllll}\toprule
                    & -o       &-as      &-es      &        \\ \midrule
\FV{present, 1, sg} &     1  &    -   &    -   &        \\ 
\FV{present, 2, sg} &    -   &     1  &    1   &        \\ 
\FV{present, 3, sg} &    -   &    -   &    -   &       \\ 
\FV{present, 1, pl} &    -   &    -   &   -   &        \\ 
\FV{present, 2, pl} &    -   &    -   &    -   &       \\ 
\FV{present, 3, pl} &    -   &    -   &   -     &       \\  \bottomrule
\end{tabular}
\end{minipage} 
\begin{minipage}{.4\textwidth} 
\begin{tabular}{llllllll}\toprule
 		    & -o & -as & -es &    \\ \midrule
\FV{present}& 1  & 1 & 1   \\
\FV{1st}    & 1  & 0 & 0    \\
\FV{2nd}    & 0  & 1 & 1   \\
\FV{3rd}    & 0  & 0 & 0    \\
\FV{sg}     & 1  & 1 & 1    \\
\FV{pl}     & 0  & 0 & 0   \\ \bottomrule
\end{tabular}
\end{minipage}
\end{example}

This example illustrates that smart initialization has nothing useful to say about alternative (or competing) systems, such as the three different inflectional patterns we are considering. Smart initialization will place each of the corresponding suffixes in the different inflectional patterns in the same position in FV-space, though the fact of the matter is that they appear in complementary distribution. The geometrical model requires that the suffixes {\em not} be in the same position, just as an {\em -ar} stem must be in a different position from an {\em -er} stem. Consider the positioning as in \BB \ in (\ref{spanish-verb-FVs}).

\medskip

However, different geometric positions for different inflectional patterns {\em do} allow for choice of affix to be resolved geometrically, in the way that we have presented here. Consider vectors as in the following table, presented graphically just below. For convenience, we identify vectors in radial coordinates (where at this point only the angle is significant, as we are operating in a 2-dimensional space).

\begin{table}[ht]
\centering
\begin{tabular}{rl} \toprule
\multicolumn{2}{c}{Positions of morphemes in radians}\\ \midrule
{\em cant} & -0.18875 \\ 
{\em com} & 1.6188 \\  
{\em -o} & 1.04273 \\  
{\em -as} & 0.17836 \\  
{\em -es} & - 0.15520 \\  \bottomrule
\end{tabular}
\caption{Position for Spanish morphemes in {\em canto, cantas, comes}, etc.}
\end{table}

\bigskip

The following graph shows how these values are located in the two-dimensional subspace of first and second person feature values. We can see that the {\em o} suffix will occur with either stem for the first person since it is the closest to that axis.

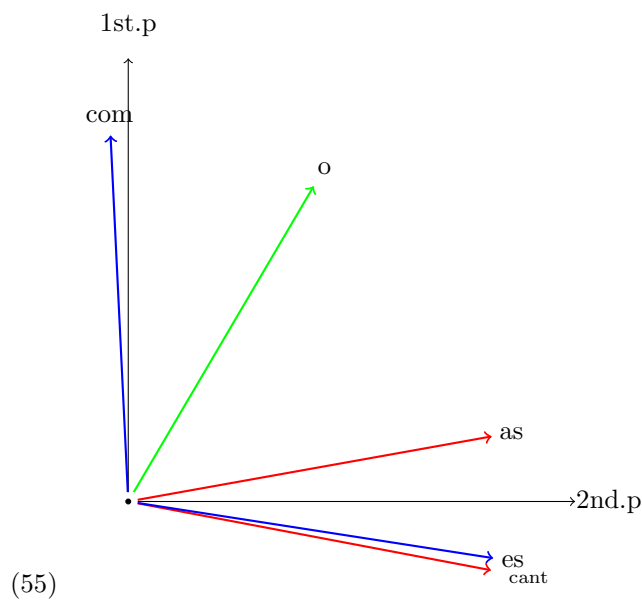
\begin{example}
\begin{tikzpicture}[scale=0.8]
\filldraw[fill=black] (0,0) circle (0.04cm);
\node (origin) at (0,0) {};
\node (2p) at (canvas polar cs:angle=0, radius=7.5cm) {};
\node (1p) at (canvas polar cs:angle=90, radius=7.5cm) {};
\node (plabel) at (canvas polar cs:angle=0, radius=7.9cm) {2nd.p};
\node (slabel) at (canvas polar cs:angle=90, radius=7.9cm) {1st.p};
\path[->] (origin) edge (2p);
\path[->] (origin) edge (1p);
\node (cant) at (canvas polar cs:angle=-10.815, radius=6.23cm) {};
\node (cantlabel) at (canvas polar cs:angle=-10.815, radius=6.7cm) {\scriptsize cant};
\path[->, red, thick] (origin) edge (cant);
\node (com) at (canvas polar cs:angle=92.75, radius=6.23cm) {};
\node (comlabel) at (canvas polar cs:angle=92.75, radius=6.4cm) {com};
\path[->, blue, thick] (origin) edge (com);
\node (o) at (canvas polar cs:angle=59.744, radius=6.23cm) {};
\node (olabel) at (canvas polar cs:angle=59.744, radius=6.4cm) {o};
\path[->, green, thick] (origin) edge (o);
\node (as) at (canvas polar cs:angle=10.236, radius=6.23cm) {};
\node (aslabel) at (canvas polar cs:angle=10.236, radius=6.4cm) {as};
\path[->, red, thick] (origin) edge (as);
\node (es) at (canvas polar cs:angle=-8.892, radius=6.23cm) {};
\node (eslabel) at (canvas polar cs:angle=-8.892, radius=6.4cm) {es};
\path[->, blue, thick] (origin) edge (es);
\end{tikzpicture}
\end{example}

\bigskip

This geometry also provides a direct account for how the stem {\em cant-} chooses affix {\em -as} over {\em -es}, while stem {\em com-} chooses affix {\em -es} over {\em as} in the second person. The vector sum of {\em com+es} is closer to the 2nd person axis than {\em com+as} and the vector sum of {\em cant+as} is closer to the second person axis than {\em cant+es}.

\bigskip

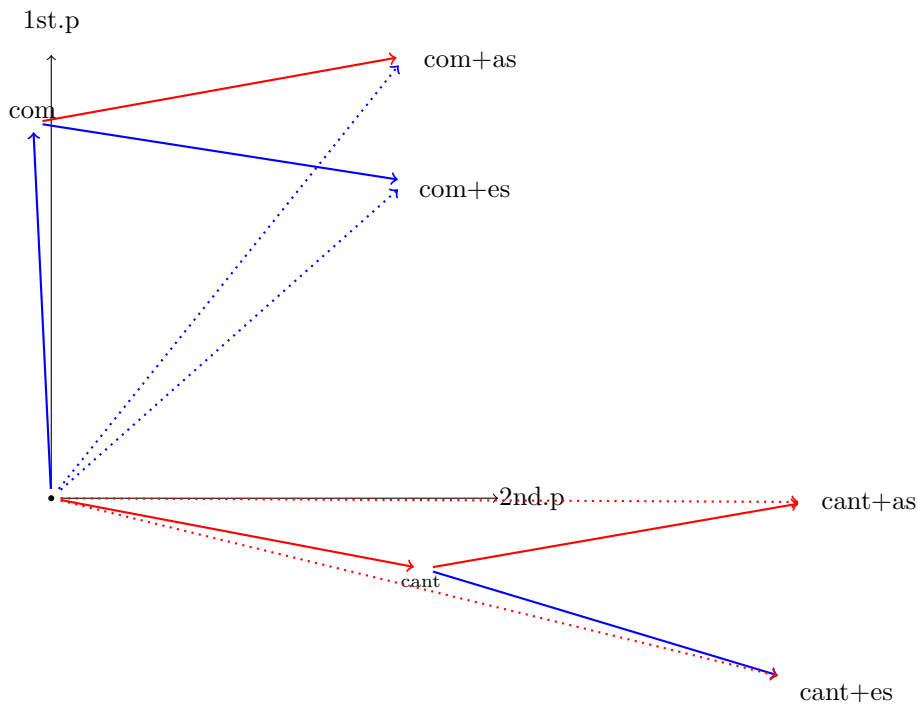
\begin{example}
\begin{tikzpicture}[scale=0.8]
\filldraw[fill=black] (0,0) circle (0.04cm);
\node (origin) at (0,0) {};
\node (2p) at (canvas polar cs:angle=0, radius=7.5cm) {};
\node (1p) at (canvas polar cs:angle=90, radius=7.5cm) {};
\node (plabel) at (canvas polar cs:angle=0, radius=7.9cm) {2nd.p};
\node (slabel) at (canvas polar cs:angle=90, radius=7.9cm) {1st.p};
\path[->] (origin) edge (2p);
\path[->] (origin) edge (1p);
\node (cant) at (canvas polar cs:angle=-10.815, radius=6.23cm) {};
\node (cantlabel) at (canvas polar cs:angle=-12.815, radius=6.23cm) {\scriptsize cant};
\path[->, red, thick] (origin) edge (cant);
\node (cantas) at (canvas polar cs:angle=-.289, radius=12.438cm) {};
\node (cantaslabel) at (canvas polar cs:angle=-.289, radius=13.438cm) {cant+as};
\path[->, red, thick] (cant) edge (cantas);
\node (cantes) at (canvas polar cs:angle=-13.85, radius=12.46cm) {};
\node (canteslabel) at (canvas polar cs:angle=-13.85, radius=13.46cm) {cant+es};
\path[->, blue, thick] (cant) edge (cantes);
\node (com) at (canvas polar cs:angle=92.75, radius=6.23cm) {};
\node (comlabel) at (canvas polar cs:angle=92.75, radius=6.4cm) {com};
\path[->, blue, thick] (origin) edge (com);
\node (comas) at (canvas polar cs:angle=51.49, radius=9.368cm) {};
\node (comaslabel) at (canvas polar cs:angle=46.49, radius=10cm) {com+as};
\path[->, red, thick] (com) edge (comas);
\node (comes) at (canvas polar cs:angle=41.92, radius=7.87cm) {};
\node (comeslabel) at (canvas polar cs:angle=36.92, radius=8.5cm) {com+es};
\path[->, blue, thick] (com) edge (comes);
\path[->, blue, thick, dotted] (origin) edge (comes);
\path[->, blue, thick, dotted] (origin) edge (comas);
\path[->, red, thick, dotted] (origin) edge (cantes);
\path[->, red, thick, dotted] (origin) edge (cantas);
\end{tikzpicture}
\end{example}


\section{Multiple patterns of inflection within a language}

\subsection{General discussion}

A central part of the task of analyzing inflectional morphologies is the analysis of different inflectional classes within a single language. That is, it is often the case that the specific choice of inflectional affixes is not fixed once and for all for  lexical stems in a given category; instead we often find that there may be as many as several dozen different patterns. These patterns often show striking similarities, all the while maintaining their differences. We will explore the analysis of such systems in some detail.

One of the fundamental tasks that we face, when we explore the inflectional morphology of a system with multiple inflectional patterns, is how to make sense of the fact that a language may have a large number of patterns that are similar but which do not all collapse into one another. Somehow the grammar---so to speak---has little trouble keeping the systems apart, though at the same time we know that the splitting and collapsing of these categories is a common and significant part of the historical change that linguists identify when they study the diachrony of such languages.

Our hypothesis is one that derives from our focus on geometry. We suggest that the set of inflectional vectors maintains a rigid relative structure across these different patterns, but that what they share is a common set of morphemes, fixed in relation to one another, that are rotated in various ways.

The careful reader may see that the exploration in the previous section regarding three inflectional patterns in Spanish did not involve rotations; why are rotations needed here? We have some reasons to believe that rotations can be grammatically controlled (that is, that grammatical information can ``trigger'' a rotation of the morpheme vector cluster), some of which we will discuss below. But at this point we are not sure whether the kinds of generalizations we looked at in Spanish should treated as we did, or with the application of rotations.

Nuer is a widely spoken Western Nilotic language in South Sudan and Ethiopia. Nouns are marked for case with three cases (nominative, genitive, locative) and two numbers singular and plural, and there are quite a number of different inflectional classes of nouns.

We have profited from the work of Matthew Baerman on Nuer, and the earlier work by Frank 1999, and from discussions of the material with our colleague Karlos Arregi.

As discussed in detail by \citet{Baerman:2012}, the paradigms of number and case suffixes on nouns in Nuer vary among at least sixteen different classes, with similar but not quite identical patterns occurring among the classes. The following table from \citet{Baerman:2012}, adapted from \citet{Frank:1999}, illustrates the complexity of variation among these classes.

\begin{example}\label{ex:classcounts}
-\begin{tabular}{llllllllllllll} \toprule
& I & II & III & IV & V & VI & VII & VIII & IX & X & XI & XII & XIII\\ \midrule
\FV{nom sg} & \nul & \nul & \nul& \nul& \nul& \nul& \nul& \nul& \nul& \nul& \nul& \nul& \nul\\
\FV{gen sg} & \nul & k\"a  & k\"a & \nul & \nul & \nul & k\"a & \nul & k\"a & \nul & k\"a  & k\"a & \nul\\
\FV{loc sg} & \nul & k\"a  & k\"a & \nul & \nul  & k\"a & \nul & \nul & \nul & k\"a  & k\"a & k\"a & \nul \\
\FV{nom pl} & \nul & \nul & ni & ni & \nul & \nul & ni & \nul & \nul & ni & \nul & \nul & \nul\\
\FV{gen pl} & ni & ni & ni & ni & \nul & ni & ni & ni & ni & ni & ni & \nul & \nul\\
\FV{loc pl} & ni & ni & ni & ni & \nul & ni & ni & \nul & ni & ni & \nul & \nul & ni\\ \bottomrule
\# of lexemes & 61 & 52 & 45 & 23 & 11& 10 & 9 &8 &5 & 3 & 2 & 2 &2\\
&&&&&&&&&&&&\\
& \scriptsize XIV & \scriptsize XV & \scriptsize XVI & \scriptsize XVII & \scriptsize XVIII &\scriptsize XIX &\scriptsize XX &\scriptsize XXI &\scriptsize XXII &\scriptsize XXIII &\scriptsize XXIV&\scriptsize XXV \\ \midrule
\FV{nom sg} & \nul & \nul & \nul& \nul& \nul& \nul& \nul& \nul& \nul& \nul& \nul& \nul\\
\FV{gen sg} & k\"a  & k\"a & \nul & \"a & \"a & k\"a & k\"a & \nul & \"a & \nul & k\"a  & k\"a \\
\FV{loc sg} & k\"a  & \nul & k\"a & \"a & \"a & \"a & \"a & \"a & k\"a  & \"a & k\"a &  k\"a\\
\FV{nom pl} & \nul & \nul & \nul & ni  & \nul & ni & \nul & \nul & \nul & ni & ni & \nul\\
\FV{gen pl} & \nul & ni & \nul & ni& ni & ni & ni & ni  & ni& ni  & \nul & k\"a \\
\FV{loc pl} & \nul & \nul & ni & ni & ni & ni & ni & ni & ni & ni & ni& ni & \scc{total:}\\
\# of lexemes & 1 & 1 & 1 & 4 & 2 & 2 & 2 & 1 & 1 & 2 & 1 & 1 & 236 \\ \bottomrule
\end{tabular}

\end{example}

\vspace{.4in}

In our discussion here, we abstract away here from consideration of suffix {\em \"a}, which occurs only among 13 out of 252 lexemes, and focus on the first sixteen classes in the table.

\medskip

As Baerman remarks, the paradigms look deceptively simple, with only three suffixes occurring among the majority of classes.

\begin{quote}
On the face of it this is a very simple system. But consider how these suffixes are distributed in the paradigms of individual nouns, some examples of which are given in Table 3. With some lexemes the suffixes are restricted to a single morphosyntactic value; with others they are {\sc syncretic}---that is, they combine two or more distinct morphosyntactic values in a single form. For example, -kä is used for the genitive singular of `potato', but for the genitive and locative singular of `bump'. While variation between syncretic and nonsyncretic distribution of morphological formatives is to be found in many languages, the sorts of patterns found in the Nuer paradigms are not ones that current models of morphology are well equipped to describe. \citep[468]{Baerman:2012} 
\end{quote}

The challenge presented by Nuer is to account for the paradigmatic variation among classes through the simplest possible model of grammar. In the model we present here,  what syncretism there is takes on a very different hue.  {\em Every} suffix carries some value for every morphosyntactic feature by the very nature of the model. Which suffix is realized for a particular combination is determined by competition among suffixes with respect to their projection on a vector for that feature-value combination. What is referred to as `blocking' in other models arises then  naturally in this model as a result of the competition we have explored in this paper. 

\medskip

How, then, does the grammar account for paradigmatic variation among classes? Our proposal is that there is a rigid configuration of vectors for the suffixes which remains constant across classes and which varies only by rotations that are applied to the vectors, with each rotation applying equally to each vector in the case of Nuer. These rotations preserve both the lengths of the vectors and the angles of separation between them. All the learner needs to derive a given class is the base configuration of vectors plus the rotation applied to that configuration that takes it to its position for that class. 

The rotations that we describe just below appear to us to be the central dynamic element of the geometrical model that we have outlined. Earlier we saw the importance of being able to ask which element from a set of vectors is closest to a particular point; now we add the possibility that a global dynamic may influence the behavior of the system as a whole.

\subsection{Rotations: Deriving inflection classes with rotations}

We can choose base weights for all the classes by applying the smart initialization heuristic that we employed earlier, adding up the number of times each suffix occurs for a given feature in the table in (\ref{ex:classcounts}) and weighting the number by the number of lexemes that represent the class in which we are counting. Using all and only the classes that have at least 3 lexemes in them, this initialization gives us the following counts.

\vspace{.1in}

\begin{example}\label{ex:weightedtotals}
\begin{tabular}{clllll} \toprule
 		&$\nul$   & ni & k\"a  \\ \midrule
\FV{sg}	& 460 & 0 & 234   \\
\FV{pl}	 & 177 & 510 & 0   \\
\FV{nom}	& 374 & 80 & 0   \\
\FV{gen}  & 127	& 218 & 119 \\						
\FV{loc}  & 126 & 212 &114 \\ \bottomrule

\end{tabular}
\end{example}

\vspace{.1in}

Normalizing these, we obtain the following:

\vspace{.1in}

 \begin{example}\label{ex:normalizedtotals}
\begin{tabular}{clllll} \toprule
 		  &$\bra{\nul}$     & $\bra{ni}$ & $\bra{k\ddot{a}}$  \\ \midrule
\FV{sg}	  & .714 & 0.0   & .817   \\
\FV{pl}	  & .275 & .851 & 0.0   \\
\FV{nom}  & .580 & .133  & 0.0   \\
\FV{gen}  & .197	& .364 & .416 \\				
\FV{loc}  & .195 & .354 & .398 \\ \bottomrule

\end{tabular}
\end{example}

\vspace{.1in}

These weights for suffix vectors result in the following values, and by selecting the maximal element in each row, we obtain the paradigm in the table below, which is the paradigm for Class 3.

\vspace{.1in}

\begin{example}\label{ex:class3activ}
\begin{tabular}{llll} \toprule
 & \nul & ni & k\"{a} \\ \midrule
\FV{nom sg}  & \Blue{1.294} & 0.133  & 0.817  \\
\FV{gen sg} & 0.911  & 0.364  & \Blue{1.233} \\
\FV{loc sg} & 0.909  & 0.354  & \Blue{1.215} \\
\FV{nom pl} & 0.855  & \Blue{0.984} & 0.000  \\
\FV{gen pl} & 0.472  & \Blue{1.215} & 0.416  \\
\FV{loc pl} & 0.470  & \Blue{1.205} & 0.398  \\ \bottomrule
\end{tabular}
\end{example}

\vspace{.1in}

Class 3 only has the third highest number of lexemes, but we think of the vectors in (\ref{ex:normalizedtotals}) as representing a central position in the space in which the vectors occur, derived from a superficial weighted average of positions. We take this set of vectors to be a set of base positions from which the other classes can be derived by rotations.

\subsection{A learning algorithm for rotations}\label{sec:nueralgorithm}

There are a number of ways to implement the decision to utilize rotations, and we describe one such way, recognizing that many of the specific choices we have made here will prove to have been arbitrary in ways that we do not yet understand.

The following algorithm was able to produce weights for suffixes that give the correct paradigm for each of classes 1 through 16 by applying, for each class, the same rotation to each of the three suffixes, starting from the class 3 base configuration given above. Here are the main features of the algorithm we used:

\begin{enumerate}
\item Approximate a transformation through all dimensions in the direction of the
activations we want by breaking down each transformation into six 2D rotations.
\item Each rotation moves the vectors in a direction determined by the
ideal state we want for a particular morphosyntactic combination.
\item Do this for each combination regardless of whether it already has the correct
suffix with maximum activation, based on the current set of weights.\footnote{The reason for this is that a rotation that is intended to affect just one feature combination will also have an effect on other combinations whose features are in the subspace in which we are rotating.}
\item We can weight the amount that we want to affect each combination according
to how far the intended winner is from the actual winner at a given point:
   \begin{itemize}
   \item If the intended winner is far behind the actual winner, make a more
strongly weighted change in a direction that would increase the activation of the intended winner for that morphosyntactic combination and
decrease the activation for the wrong winner.
   \item If the intended winner is already winning, increase its activation for that
morphosyntactic combination by a weaker amount.
   \end{itemize}
\end{enumerate}

If the independent variable is the activation of the real winner minus the activation of the intended winner, then it is natural to use a sigmoid function, which has a low value when the independent variable is less than 0 (the intended winner is already winning) and goes up sharply to a higher value when the independent variable is greater than 0 (the intended winner is losing.) Squaring the function seems to produce an even more satisfactory result.\footnote{Anyone familiar with neural nets and their recent incarnation as deep learning models will see natural connections here; it would be hard to say what has influenced what, since the ideas that describe representations on neural nets are fundamentally geometric in nature.} See (\ref{ex:logisticequation}) and (\ref{ex:logisticgraph}).

Here, \(\eta_i\) is the factor by which we multiply our rotational increment on each sub-iteration, \(\hat{a}_i\) is the activation of the current real winner and \(a_i\) that activation of the intended winner for morphosyntactic combination $i$. 

\begin{example}\label{ex:logisticequation}
\(\eta_i = \frac{1}{\left[1 + e^{-2(\hat{a_i}-a_i)}\right]^2}\)
\end{example}

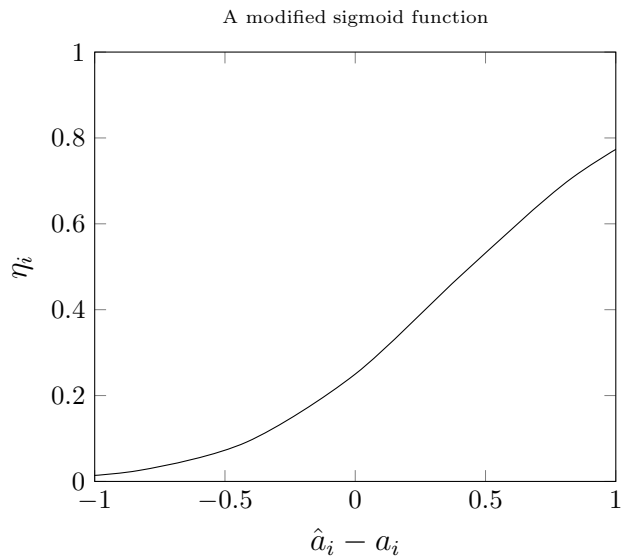
\begin{example}\label{ex:logisticgraph}
\begin{tikzpicture}[scale=1.0]
\begin{axis}[
      title=\scriptsize A modified sigmoid function,
       xlabel={\large $\hat{a}_i-a_i$},
      ylabel={\large $\eta_i$ },
      xmin = -1.0,
      ymin = 0,
      xmax = 1,
      ymax = 1,
 ]
    \addplot [smooth, no markers] {1/(1 + exp(-2*x))^2};
    \end{axis};
\end{tikzpicture}
\end{example}

On each iteration, there are six sub-iterations, each of which considers, in turn,
one of the six morphosyntactic combinations. The algorithm chooses two feature values in which to rotate the suffix vectors:

\begin{enumerate}
\item The feature for which the intended winner's weight for that feature exceeds that of its closest rival by the maximum amount. We are going to rotate away from this axis, since here is where the intended winner can most afford to lose some weight.
\item One of the two features, randomly chosen, that comprise the combination we are concerned with for this subiteration. We are going to rotate {\em towards} this axis.
\end{enumerate}

The following 2D graph shows how this rotation works for one example. We establish a 2-dimensional space in which the $x$-axis is a feature in which the intended winner has greater value than its competitors and the $y$-axis is the dimension in which we want the intended winner to gain. In the figure just below, these are \FV{nominative} and \FV{genitive}, with \nul \ the intended winner and \textit{k\"a} the false winner. Arrowed lines in solid colors indicate the original positions of vectors. Dotted lines in the same color indicate the ending position of a vector after rotation. We can see that a positive (i.e.\ counter-clockwise) rotation of the vectors will increase the projection of \nul \ on the genitive axis, as desired, and decrease the projection of {\em k\"a} on the genitive axis. The fact that \nul \ loses some of its projection on the nominative axis does not matter. The nominative axis was chosen to rotate away from  since \nul \ can afford to lose value there.

\begin{example}
\begin{tikzpicture}[scale=1.3]
\begin{axis}[
      axis equal,
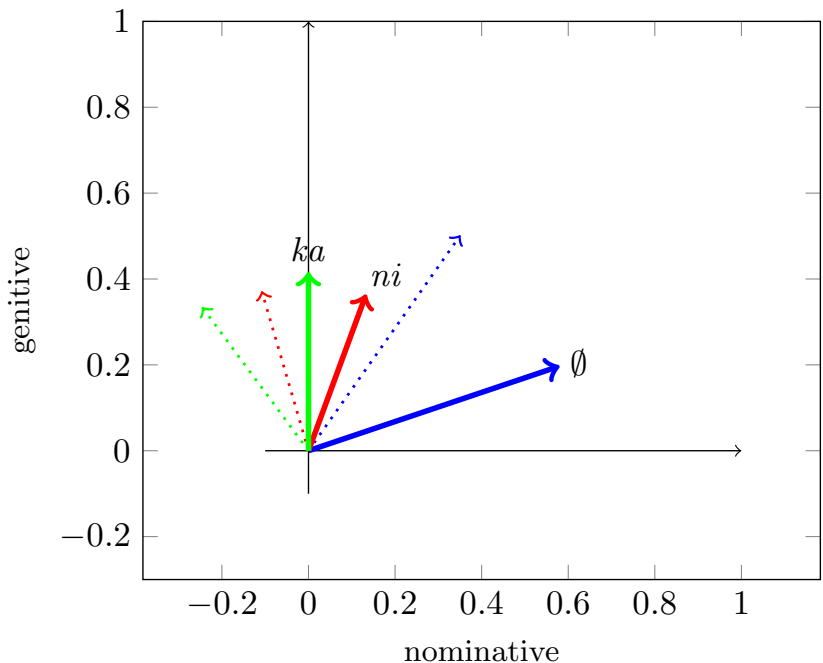
      title=\scriptsize Graph of suffixes $\emptyset$ ni and k\"a in 2 dimensions: \FV{nominative} and  \FV{genitive}.,
       xlabel={\small nominative },
      ylabel={\small genitive },
      xmin = -0.2,
      ymin = -0.3,
      xmax = 1,
      ymax = 1,
 ]
\addplot[->,black, no markers, thin] coordinates
   {(0,-0.1) (0,1)};
\addplot[->,black, no markers, thin] coordinates
   {(-0.1,0) (1,0)};
\addplot[->, blue , no markers, ultra thick] coordinates
   {(0,0) ( 0.580 , 0.197 )};
\addplot[->, blue , no markers, thick, dotted] coordinates
   {(0,0) ( 0.351 , 0.502 )};
\node at ( 0.610 , 0.205 ) {\textit{ $\emptyset$ }};
\addplot[->, red , no markers, ultra thick] coordinates
   {(0,0) ( 0.133 , 0.364 )};
\addplot[->, red , no markers, thick, dotted] coordinates
   {(0,0) ( -0.108 , 0.372 )};
\node at ( 0.18 , 0.41 ) {\textit{ni}};
\addplot[->, green , no markers, ultra thick] coordinates
   {(0,0) ( 0.0 , 0.416 )};
\addplot[->, green , no markers, thick, dotted] coordinates
   {(0,0) ( -0.246 , 0.335 )};
\node at ( 0.0 , 0.47 ) {\textit{ka}};
    \end{axis};
\end{tikzpicture}
\end{example}

\subsection{Results of learning algorithm to derive 16 classes}

%

In the following table, all classes were derived by rotation from the base position of vectors given above in (\ref{ex:normalizedtotals}) which corresponds to Class 3. The term `distance' refers to the number of cells in the paradigm at which the class in question differs from Class 3. The term `smallest margin' refers to the smallest difference in activation among all the cells in the paradigm for that class between a winning suffix and its closest competitor (i.e.\ projection onto the vector of a morphosyntactic feature combination). The algorithm was run 100 times for each class, with the results averaged.

\bigskip

\begin{example}
\begin{tabular}{lcclr}\toprule
Class & \#lexemes & Distance\ from  & Smallest & \#iterations \\
&& base pos'n & margin &\\\midrule
1 & 61 & 3 & 0.034 & 8.73 \\
2 & 52 & 1 & 0.033 & 1.48 \\
4 & 23 & 2 & 0.061 & 2.00 \\
5 & 11 & 5 & 0.044 & 4.86 \\
6 & 10 & 2 & 0.027 & 7.36 \\
7 & 9 & 1  & 0.046 & 2.83 \\
8 & 8 & 3 &  0.033 & 10.03 \\
9 & 5 & 2 &  0.024 & 6.69 \\
10 & 3 & 1 & 0.069 & 2.98 \\
11 & 2 & 2 & 0.048 & 5.22 \\
12 & 2 & 2 & 0.046 & 5.56 \\
13 & 2 & 3 & 0.026 & 9.72 \\
14 & 1 & 3 & 0.057 & 3.50 \\
15 & 1 & 3 & 0.040 & 7.91 \\
16 & 1 & 3 & 0.037 & 7.62 \\ \bottomrule
\end{tabular}
\end{example}

\subsection{On Baerman's ``variable defaults''} 

Baerman's analysis\footnote{\citep[482]{Baerman:2012}} of Nuer suffixes proposes to capture certain statistical preferences among the configurations that occur in Nuer through a set of ``variable defaults'' that are subject to a set of implications. Some of the defaults he proposes are the following.

\begin{enumerate}
\item By default, genitive and locative singular are {\em k\"a}. 
\item By default, genitive and locative plural are {\em ni}. 
\item  By default, nominative plural is \nul. 
\item If the nominative plural is {\em ni}, this entails {\em ni} in the other plural cases. 
\item By default, genitive and locative are identical.
\end{enumerate}

Most of the tendencies that Baerman observes will fall out naturally from our model without the need to posit default rules. Consider again the base values for suffix vectors we proposed in (\ref{ex:normalizedtotals}), repeated here as (\ref{ex:normalizedtotals2}).

\begin{example}\label{ex:normalizedtotals2}
\begin{tabular}{clllll} \toprule
 		  &$\bra{\nul}$     & $\bra{ni}$ & $\bra{k\ddot{a}}$  \\ \midrule
\FV{sg}	  & .714 & 0.0   & .817   \\
\FV{pl}	  & .275 & .851 & 0.0   \\
\FV{nom}  & .580 & .133  & 0.0   \\
\FV{gen}  & .197	& .364 & .416 \\				
\FV{loc}  & .195 & .354 & .398 \\ \bottomrule
\end{tabular}
\end{example}

\begin{enumerate}
\item Baerman's default rule 1 follows from the fact that the projections of vector {\em k\"a} on genitive singular\ and locative singular are \(.817+.416 = 1.233\) and \(.817+.398 = 1.215\) respectively, far greater than that of the other two suffixes. Consequently, the vectors would have to be rotated to a great extent in order to change this tendency.
\item Baerman's default rule 2 follows in a similar fashion, where vector {\em ni} has projections of \(.851+.364=1.215\) and \(.851+.354 = 1.205\) on vectors genitive plural\ and locative plural.
\item His rule 3 is not reflected as strongly in the values we have proposed, but neither is the presence of {\em ni} in nom.pl.\ as strongly attested among the classes.
\item The tendency towards rule 4 can be accounted for as follows. If the nominative plural  is {\em ni}, as it is in our base class, then to maintain {\em ni} in the nominative plural we need to avoid rotations that reduce the strong weight of $.852$ in the plural for {\em ni}, since its weight in the nominative is less strong and cannot be depended on to maintain a strong projection for {\em ni} on the nom.pl.\ vector. Maintaining a strong weight in the plural for {\em ni} will tend to make it the winner for other combinations in the plural.
\item The tendency for genitive and locative to be identical is enforced by the values we have proposed for them among the three vectors, which are very close for all of them. 
\end{enumerate}

\subsection{Deponent verbs: an example from Latin}

We have explored now how relations between different inflectional classes can be accounted for through rotations of a complete set of vectors for morphemes in the space of feature values.  We would like to explore briefly the way in which the deponent verbs of Latin also call to mind a treatment employing the notion of rotation of vectors. 

The case of deponent verbs in Latin involves verbs whose morphology is in a formal sense regular, but out of sync with the expected grammatical specification. In the present indicative active, a deponent verb appears with the suffixes that otherwise occur on a passive verb (and these verbs have no passive forms). Viewed from a high enough altitude, what this case illustrates is the way in which a subpart of a grammar can be maintained in its substance but shifted {\em en bloc} with respect to another set of grammatical features that are active in the language. It is systems of this sort that we believe a geometric morphology will be best able to deal with. 

The following data, taken from \citet[198]{Stump:2016} compare the paradigms of a non-deponent and a deponent first conjugation verb.

\bigskip

\begin{minipage}[t]{.45\textwidth}
\begin{tabular}{lll}\toprule
&\textsc{par\=are} &\Gl{prepare}\\ \midrule
Active & 1sg & {\em par\=o}\\
& 2sg & {\em par\=as}\\
& 3sg & {\em parat}\\
& 1pl & {\em par\=amus}\\
& 2pl & {\em par\=atis}\\
& 3pl & {\em parant}\\ \hline
Passive & 1sg & {\em paror}\\
& 2sg & {\em par\=aris}\\
& 3sg & {\em par\=atur}\\
& 1pl & {\em par\=amur}\\
& 2pl & {\em par\=amini}\\
& 3pl & {\em parantur}\\ \bottomrule
\end{tabular}
\end{minipage}
\begin{minipage}[t]{.45\textwidth}
\begin{tabular}{lll}\toprule
&\textsc{c\=on\=ari} &\Gl{try}\\ \midrule
Active & 1sg & {\em c\=onor}\\
& 2sg & {\em c\=on\=aris}\\
& 3sg & {\em c\=on\=atur}\\
& 1pl & {\em c\=on\=amur}\\
& 2pl & {\em c\=on\=amini}\\
& 3pl & {\em c\=onantur}\\ \hline
Passive && (none) \\ \bottomrule
\end{tabular}
\end{minipage} 
\bigskip

The active voice in the deponent verb is expressed with the passive voice suffixes and the passive voice does not occur with the deponent stem. The following matrices express the counts of each feature value for each affix in the paradigm of regular verb {\em par\=are}. As we did in the section on Spanish \S\ref{sec:spanish}, we shall not treat theme vowels as separate affixes, and consider suffixes such as {\em \=amur} as a single morpheme.

\begin{example} 
\label{deponent}
\[
\begin{blockarray}{lcccccc|cccccc}
& \multicolumn{6}{c}{Active suffixes} & \multicolumn{6}{c}{Passive suffixes} \\ 
& \mathrm{o} & \mathrm{\overline{a}s} & \mathrm{at} & \mathrm{\overline{a}mus} & \mathrm{\overline{a}tis} & \mathrm{ant} & \mathrm{or} & \mathrm{\overline{a}ris} & \mathrm{atur} & \mathrm{\overline{a}mur} & \mathrm{\overline{a}mini} & \mathrm{antur} \\
\begin{block} {l(cccccc|cccccc)}
\FV{singular} & 1  & 1 & 1 & 0 & 0 & 0 & 1 & 1 & 1 & 0 & 0 & 0\\ 
 \FV{plural}  & 0  & 0 & 0 & 1 & 1 & 1 & 0 & 0 & 0 & 1 & 1 & 1 \\
\FV{1st} &  1  & 0 & 0 &  1  & 0 & 0 &  1  & 0 & 0 &  1  & 0 & 0 \\ 
 \FV{(2nd} & 0  & 1 & 0 & 0  & 1 & 0 & 0  & 1 & 0 & 0  & 1 & 0 \\
 \FV{3rd} & 0  & 0 & 1& 0  & 0 & 1& 0  & 0 & 1& 0  & 0 & 1\\
\FV{active} &   1 & 1 & 1&   1 & 1 & 1 & 0 & 0 & 0 & 0 & 0 & 0 \\ 
\FV{passive} & 0 & 0 & 0 & 0 & 0 & 0 &   1 & 1 & 1&   1 & 1 & 1  \\
\end{block}
\end{blockarray}
 \]
\end{example} 

Normalizing the columns, we obtain the following values, expressed algebraically:

\begin{example} 
\label{deponentnorm}
\[
\begin{blockarray}{lcccccc|cccccc}
& \multicolumn{6}{c}{Active suffixes} & \multicolumn{6}{c}{Passive suffixes} \\
& \mathrm{o} & \mathrm{\overline{a}s} & \mathrm{at} & \mathrm{\overline{a}mus} & \mathrm{\overline{a}tis} & \mathrm{ant} & \mathrm{or} & \mathrm{\overline{a}ris} & \mathrm{atur} & \mathrm{\overline{a}mur} & \mathrm{\overline{a}mini} & \mathrm{antur} \\
\begin{block} {l(cccccc|cccccc)}
\FV{singular} & \frac{1}{\sqrt{3}}  & \frac{1}{\sqrt{3}} & \frac{1}{\sqrt{3}} & 0 & 0 & 0 & \frac{1}{\sqrt{3}} & \frac{1}{\sqrt{3}} & \frac{1}{\sqrt{3}} & 0 & 0 & 0\\ 
 \FV{plural}  & 0  & 0 & 0 & \frac{1}{\sqrt{3}} & \frac{1}{\sqrt{3}} & \frac{1}{\sqrt{3}} & ) & 0 & 0 & \frac{1}{\sqrt{3}} & \frac{1}{\sqrt{3}} & \frac{1}{\sqrt{3}} \\
\FV{1st} &  \frac{1}{\sqrt{3}}  & 0 & 0 &  \frac{1}{\sqrt{3}}  & 0 & 0 &  \frac{1}{\sqrt{3}}  & 0 & 0 &  \frac{1}{\sqrt{3}}  & 0 & 0 \\ 
 \FV{2nd} & 0  & \frac{1}{\sqrt{3}} & 0 & 0  & \frac{1}{\sqrt{3}} & 0 & 0  & \frac{1}{\sqrt{3}} & 0 & 0  & \frac{1}{\sqrt{3}} & 0 \\
 \FV{3rd} & 0  & 0 & \frac{1}{\sqrt{3}}& 0  & 0 & \frac{1}{\sqrt{3}}& 0  & 0 & \frac{1}{\sqrt{3}}& 0  & 0 & \frac{1}{\sqrt{3}}\\
\FV{active} &   \frac{1}{\sqrt{3}} & \frac{1}{\sqrt{3}} & \frac{1}{\sqrt{3}}&   \frac{1}{\sqrt{3}} & \frac{1}{\sqrt{3}} & \frac{1}{\sqrt{3}} & 0 & 0 & 0 & 0 & 0 & 0 \\ 
\FV{passive} & 0 & 0 & 0 & 0 & 0 & 0 &   \frac{1}{\sqrt{3}} & \frac{1}{\sqrt{3}} & \frac{1}{\sqrt{3}}&   \frac{1}{\sqrt{3}} & \frac{1}{\sqrt{3}} & \frac{1}{\sqrt{3}}  \\
\end{block}
\end{blockarray}
 \]
\end{example} 

\ldots and expressed numerically:

\begin{example} 
\label{deponentnorm2}
\[
\begin{blockarray}{lcccccc|cccccc}
& \multicolumn{6}{c}{Active suffixes} & \multicolumn{6}{c}{Passive suffixes} \\
& \mathrm{o} & \mathrm{\overline{a}s} & \mathrm{at} & \mathrm{\overline{a}mus} & \mathrm{\overline{a}tis} & \mathrm{ant} & \mathrm{or} & \mathrm{\overline{a}ris} & \mathrm{atur} & \mathrm{\overline{a}mur} & \mathrm{\overline{a}mini} & \mathrm{antur} \\
\begin{block} {l(cccccc|cccccc)}
\FV{singular} & 0.577  & 0.577 & 0.577 & 0 & 0 & 0 & 0.577 & 0.577 & 0.577 & 0 & 0 & 0\\ 
 \FV{plural}  & 0  & 0 & 0 & 0.577 & 0.577 & 0.577 & 0 & 0 & 0 & 0.577 & 0.577 & 0.577 \\
\FV{1st} &  0.577  & 0 & 0 &  0.577  & 0 & 0 &  0.577  & 0 & 0 &  0.577  & 0 & 0 \\ 
 \FV{2nd} & 0  & 0.577 & 0 & 0  & 0.577 & 0 & 0  & 0.577 & 0 & 0  & 0.577 & 0 \\
 \FV{3rd} & 0  & 0 & 0.577& 0  & 0 & 0.577& 0  & 0 & 0.577& 0  & 0 & 0.577\\
\FV{active} &   0.577 & 0.577 & 0.577&   0.577 & 0.577 & 0.577 & 0 & 0 & 0 & 0 & 0 & 0 \\ 
\FV{passive} & 0 & 0 & 0 & 0 & 0 & 0 &   0.577 & 0.577 & 0.577&   0.577 & 0.577 & 0.577  \\
\end{block}
\end{blockarray}
 \]
\end{example} 

We can see that there is an exact correspondence between the feature values of the active affixes and the passive affixes except for the active and passive values themselves. If we look at the 2D subspace of the active and passive feature values, all the active affixes will be in one position and all the passive affixes in another, as shown in the following graph.

\bigskip

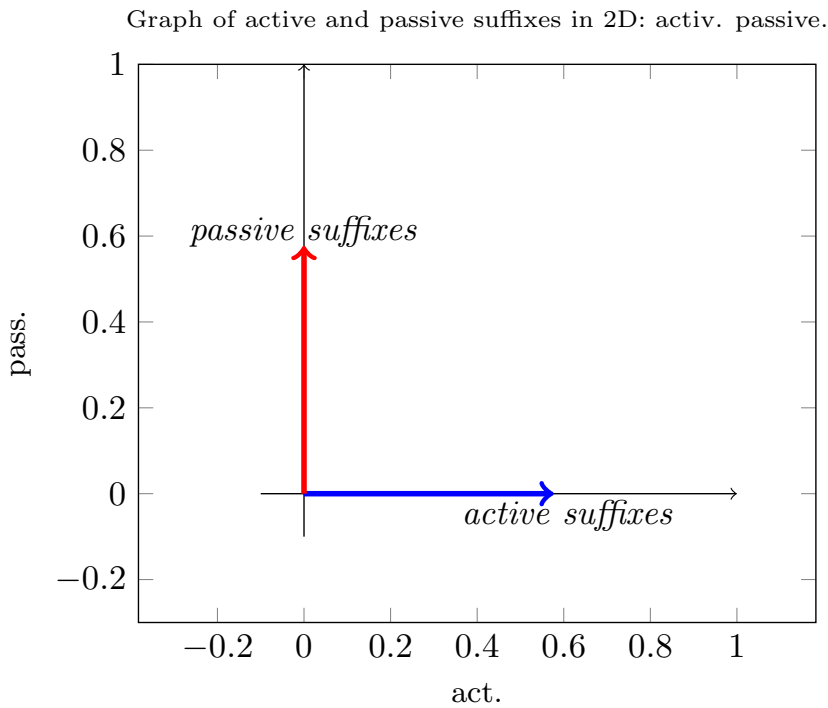
\begin{example}
\begin{tikzpicture}[scale=1.3]
\begin{axis}[
      axis equal,
      title=\scriptsize Graph of active and passive suffixes  in 2D: activ. passive.,
       xlabel={\small act. },
      ylabel={\small pass. },
      xmin = -0.2,
      ymin = -0.3,
      xmax = 1,
      ymax = 1,
 ]
\addplot[->,black, no markers, thin] coordinates
   {(0,-0.1) (0,1)};
\addplot[->,black, no markers, thin] coordinates
   {(-0.1,0) (1,0)};
\addplot[->, blue , no markers, ultra thick] coordinates
   {(0,0) ( 0.577 , 0 )};
\node at ( 0.610 , -0.05 ) {\textit{active suffixes}};
\addplot[->, red , no markers, ultra thick] coordinates
   {(0,0) ( 0 , 0.577 )};
\node at ( 0, 0.610) {\textit{passive suffixes}};
    \end{axis};
\end{tikzpicture}
\end{example}

If we apply a rotation of a three-quarter turn counter-clockwise to all the affixes, just in that 2D subspace, we will end up with the following positions for the two sets of vectors.

\begin{example}
\begin{tikzpicture}[scale=1.3]
\begin{axis}[
      axis equal,
      title=\scriptsize Graph of active and passive suffixes  in 2D: activ. passive.,
       xlabel={\small act. },
      ylabel={\small pass. },
      xmin = -0.2,
      ymin = -0.7,
      xmax = 1,
      ymax = 1,
 ]
\addplot[->,black, no markers, thin] coordinates
   {(0,-0.1) (0,1)};
\addplot[->,black, no markers, thin] coordinates
   {(-0.1,0) (1,0)};
\addplot[->, red , no markers, ultra thick] coordinates
   {(0,0) ( 0.577 , 0 )};
\node at ( 0.610 , -0.05 ) {\textit{passive suffixes}};
\addplot[->, blue , no markers, ultra thick] coordinates
   {(0,0) ( 0 , -0.577 )};
\node at ( 0, -0.610) {\textit{active suffixes}};
    \end{axis};
\end{tikzpicture}
\end{example}

The affixes that normally occur in the passive paradigm now have the exact values that the active affixes had and the active suffixes have moved to negative territory. This will result in the passive suffixes being chosen for feature-value combinations involving active voice -- exactly what we see for deponent verbs. In general, a rotation in a two-dimensional subspace (active, passive)  is expressed as follows, for a rotation through an angle of $\theta$:

\begin{example} 
\label{sincos}
\[
\begin{blockarray}{lcc}
& \FV{active} & \FV{passive}  \\
\begin{block} {l(cc)}
\FV{active} & \cos\theta  & -\sin\theta \\
 \FV{passive}  & \sin\theta  & \cos\theta \\
\end{block}
\end{blockarray}
 \]
\end{example} 

In this case, the counter-clockwise rotation of a three-quarter turn is  $\frac{3\pi}{2}$ radians ($270^o$), whose sine and cosine are $-1$ and 0 respectively. This gives us the following rotation matrix, which, applied to the feature-value set for a regular verb will result in the set for a deponent verb. Cells whose values depart from those in an identity matrix are colored blue.

\begin{example}
\(Rotation Matrix=\left[\begin{array}{lllllrl}
1  & 0  & 0  & 0  & 0  & 0  & 0 \\
0  & 1  & 0  & 0  & 0  & 0  & 0 \\
0  & 0  & 1  & 0  & 0  & 0  & 0 \\
0  & 0  & 0  & 1  & 0  & 0  & 0 \\
0  & 0  & 0  & 0  & 1  & 0  & 0 \\
0  & 0  & 0  & 0  & 0  & \cellcolor{blue!30} 0  & \cellcolor{blue!30} 1 \\
0  & 0  & 0  & 0  & 0  & \cellcolor{blue!30} -1  & \cellcolor{blue!30} 0 \\
\end{array}\right]\)
\end{example}

Here is how the matrix of feature values looks after applying this rotation.  

\begin{example} 
\label{deponentnormrotated}
\[
\begin{blockarray}{lcccccc|cccccc}
& \multicolumn{6}{c}{Active suffixes} & \multicolumn{6}{c}{Passive suffixes} \\
& \mathrm{o} & \mathrm{\overline{a}s} & \mathrm{at} & \mathrm{\overline{a}mus} & \mathrm{\overline{a}tis} & \mathrm{ant} & \mathrm{or} & \mathrm{\overline{a}ris} & \mathrm{atur} & \mathrm{\overline{a}mur} & \mathrm{\overline{a}mini} & \mathrm{antur} \\
\begin{block} {l(cccccc|cccccc)}
\FV{singular} &  0.58  &  0.58 &  0.58 &  0 &  0 &  0 &  0.58 &  0.58 &  0.58 &  0 &  0 &  0\\ 
 \FV{plural}  &  0  &  0 &  0 &  0.58 &  0.58 &  0.58 &  0 &  0 &  0 &  0.58 &  0.58 &  0.58 \\
\FV{1st} &   0.58  &  0 &  0 &   0.58  &  0 &  0 &   0.58  &  0 &  0 &   0.58  &  0 &  0 \\ 
 \FV{2nd} &  0  &  0.58 &  0 &  0  &  0.58 &  0 &  0  &  0.58 &  0 &  0  &  0.58 &  0 \\
 \FV{3rd} &  0  &  0 &  0.58&  0  &  0 &  0.58&  0  &  0 & 0.58& 0  & 0 & 0.58\\
\FV{active} &   0 & 0 & 0&   0 & 0 & 0 & 0.58 & 0.58 & 0.58 & 0.58 & 0.58 & 0.58 \\ 
\FV{passive} & -0.58 & -0.58 & -0.58 & -0.58 & -0.58 & -0.58 &   0 & 0 & 0&   0 & 0 & 0  \\
\end{block}
\end{blockarray}
 \]
\end{example} 

The feature values for the passive suffixes (right half of the matrix) are now exactly what they are for the active suffixes without the rotation.
\section{Conclusions}

This paper illustrates some of the initial results that arise from treating the problems of inflectional morphology from a geometrical point of view. From our perspective, the advantages of a geometrical perspective are three in number:

\begin{enumerate}
\item Despite the initial novelty of thinking of morphemes as vectors in a space of dimensionality greater than 3, it allows for a visually intuitive way of seeing how the structural information of morphemes interact with one another---how the inflectional information of a stem interacts with the information of a neighboring affix, for example. This kind of interaction includes a natural geometric account of why more specific morphemes typically dominate over more general morphemes: it is because they are closer to the target. 

\item There is a large set of learning algorithms in machine learning that are easily applicable to models grounded in geometry. 

\item The analysis is distinctly less derivational than what is found in analyses in some other approaches. We certainly recognize the importance of including several distinct representations in the analysis of a given word or utterance, and in that sense we are perfectly comfortable with the notion of a derivation in the abstract. But derivational accounts run certain risks, in our opinion. It has not been our intent in this paper to contrast our account with others, but we have tried to develop an account in which we avoid two things that are natural in a derivational context: (1) the use of rules that have been called {\em impoverishment} rules or feature-deletion, and (2) a style of explanation that employs a sort of abstract topography which aims to offer a linguistic explanation. The first we avoid because we are not sure that such rules are formally, or theoretically, coherent\footnote{At risk of over-simplifying, we take it that when we develop a theory, we employ objects, relations, and functions of various sorts. Features and feature values are distinct sorts of entities, and features {\em may} be understood as functions from objects to a set of feature values. Functions are not the sort of entity that delete (or are deleted) by virtue of their nature of mapping from one domain to another co-domain. It seems to us that care was employed in developing the theory of autosegmental phonology so as not to risk inadvertent theoretical incoherence.} The second is a style of explanation that we are uncomfortable with, which takes pedagogical metaphors as if they were meant in some sense literally, such as offering an explanation of a phenomenon by the saying that one computation occurs {\em here}, and another occurs {\em there} (one in the lexicon, say, and one in the syntax) and that it is this imaginary distance that provides an explanation of their ignorance of each other.

\end{enumerate}

The validity of this approach will depend on whether it can be extended to the range of phenomena known to morphologists, and we invite our reader to join in that exploration.

\section{Some remarks about learnability}

\subsubsection{Falsifiability and learnability}

Karl Popper is often associated with the view that an important goal for philosophy is to provide a means for determining which human enterprises are sciences and which are not, and the view that a scientific theory must provide explicit means of proving itself wrong. Popper was concerned that such fields as Freudian psychology and Marxist economics were not scientific, and the justification for this belief lay in the fact that no one was able to specify observations that would prove either theory wrong. Regardless of what one thinks of this solution to what Popper called the problem of demarcation, it is a simple error---a misunderstanding---to think that something like it can be used as a measure for declaring one theory {\em more} scientific or {\em less} scientific than another. Popper's solution was not intended (and should not be understood as) an apriori evaluation of a scientific theory's desirability.

\subsubsection{Ignorance of non-existence}

The second reason that we should avoid preferring theories that exclude grammars over those that do not is that as a discipline, we do not have measures in place for evaluating claims about the existence or non-existence of particular phenomena across the range of human languages. Broad claims are not infrequently made by linguists motivated to rule out linguistic phenomena, and the only means to check the validity of their claims is the hope (if that is the proper word) that someone who reads their work and knows a counterexample is motivated enough to inform them of it. Many linguists operate by the principle that a linguist has the freedom to make a claim (as if being granted permission by the late Sir Karl Popper) by virtue of the fact that someone else might do the work to show them wrong. This is not a reasonable way to run a science.

\subsubsection{Expansionary phase} The third reason is tightly connected the second: linguistics as a field is still in an expansionary stage, in the sense that any linguist who spends a year or two studying a language will inevitably discover a new phenomenon that is new and interesting. A  student who studies a language and only finds phenomena that match perfectly what has been described in the literature that they are  exposed to is not a linguist; we are certain that they have not looked hard enough. 

Closely related to that, too, is the fact that there have been no important publications in the field to date whose primary contribution is the elimination of a certain aspect of formal grammar. All important works either expand our universe of known linguistic phenomena, or they provide simpler and more insightful accounts of what we already were aware of. 

\subsubsection{Learning is not random selection} It was perhaps once reasonable to think that the discovery of limitations on the class of possible human languages would shed light on how language is learned, but that time is long gone, and has been gone since the beginning of machine learning in the 1980s. We now know a great deal about learning, especially in the computational context, and we know about many ways in which structure can be inferred from data.

\section{Appendix: summary of important variables mentioned}
We summarize our use of different spaces so far in the following tables. Note that {\em number of entries} is in a linguistic sense the number of degrees of freedom in the system; it is the number of distinct entries in the entire description of the paradigm.  

\vspace{.3in}
\begin{table}[h]
\begin{tabular}{lllll} \toprule
Parameter                                       & Symbol & Illustrated  \\ \midrule
Number of morphemes                             & NumMorph    & \# columns in \BB{}  \\
Number of feature values in paradigm space      & NumFeaVal  & \# rows in \BB{} or columns in \PHI{}.               \\
Number of inflectional features & InflDimen  										 \\
Number of positions in paradigm & NumParaPos  & \# rows in \PHI{} \\ \bottomrule 
\end{tabular}
\caption{Important parameters}
\end{table}
\vspace{.3in}

\begin{table}
\begin{tabular}{llll} \toprule
	  & Number of dimensions &  Number of feature values & number of entries   \\ \midrule
{\bf PS: }{\em paradigm space} &  number of inflectional  & sum of possible values & product of possible  \\  
   & features (InflDimen) & of inflectional features & values  \\ 
   && (NumFeaVal) \\

English weak verbs  & 3 & 7 & 12 (each a morpheme).\\ 
& & & \\ \midrule

{\bf VS: }{\em feature value space} & number of feature   & 2 in each dimension (0,1) & number of morphemes $\times$\\ 
                          & values in feature space & (trivially) &    number of feature-values       \\ 
        & NumFeaVal \\
English weak verbs   & 7 &   2 in each dimension for  & 3 $\times$ 7 (each a real number). \\ 
    & &paradigm points  & NumMorph $\times$ NumFeaVal. \\ \bottomrule
 					 
\end{tabular}
\caption{Paradigm space versus Feature-value space}
\end{table}

\bibliographystyle{plainnat}
\bibliography{main}{}

\begin{thebibliography}{13}
\providecommand{\natexlab}[1]{#1}
\providecommand{\url}[1]{\texttt{#1}}
\expandafter\ifx\csname urlstyle\endcsname\relax
  \providecommand{\doi}[1]{doi: #1}\else
  \providecommand{\doi}{doi: \begingroup \urlstyle{rm}\Url}\fi

\bibitem[Aaronson(2013)]{aaronson}
Scott Aaronson.
\newblock \emph{{Q}uantum {C}omputing {S}ince {D}emocritus}.
\newblock Cambridge University Press, 2013.

\bibitem[Baerman(2012)]{Baerman:2012}
Matthew Baerman.
\newblock {P}aradigmatic chaos in {N}uer.
\newblock \emph{Language}, 88\penalty0 (3):\penalty0 467--494, 2012.

\bibitem[Corbett and Fraser(1993)]{corbett-fraser}
Greville~G.\ Corbett and Andrew Fraser.
\newblock {N}etwork {M}orphology: {A} {DATR} account of {R}ussian nominal
  inflection.
\newblock \emph{Journal of Linguistics}, 29\penalty0 (1):\penalty0 113--142,
  1993.

\bibitem[Evans and Gazdar(1989)]{Evans}
Roger Evans and Gerald Gazdar.
\newblock {A}n introduction to the {S}ussex {P}rolog {DATR} system.
  ({C}ognitive {S}cience {R}esearch {R}eport {CSRP} 139).
\newblock In Roger Evans and Gerald Gazdar, editors, \emph{{T}he {DATR}
  {P}apers}, pages 63--71. University of Sussex, 1989.

\bibitem[Frank(1999)]{Frank:1999}
Wright~Jay Frank.
\newblock {N}uer noun morphology.
\newblock Master's thesis, State University of New York, Buffalo, 1999.

\bibitem[Fraser and Hudson(1992)]{Fraser}
Norman~M.\ Fraser and Richard Hudson.
\newblock {I}nheritance in {W}ord {G}rammar.
\newblock In \emph{{C}omputational {L}inguistics}, volume~18, pages 133--158,
  1992.

\bibitem[Goldsmith(1994)]{Goldsmith:1994}
John Goldsmith.
\newblock {G}rammar within a neural net.
\newblock In \emph{{T}he {R}eality of {L}inguistic {R}ules}, pages 95--113.
  John Benjamins, 1994.

\bibitem[Goldsmith and Rosen()]{gr}
John Goldsmith and Eric Rosen.
\newblock {L}earning morphosyntactic categories and features for inflectional
  paradigms.
\newblock Submitted to the 2017 annual meeting of the Association for
  Computational Linguistics.

\bibitem[Russell et~al.(1992)Russell, Ballim, Carroll, and
  Warwick-Armstrong]{Russell}
Graham Russell, Afzal Ballim, John Carroll, and Susan Warwick-Armstrong.
\newblock {A} practical approach to multiple-default inheritance for
  unification-based lexicons.
\newblock In \emph{{C}omputational {L}inguistics}, volume~18, pages 311--337,
  1992.

\bibitem[Stankiewicz(1968)]{Stankiewicz}
Edward Stankiewicz.
\newblock \emph{{D}eclension and gradation of {R}ussian substantives in
  contemporary standard {R}ussian}.
\newblock Mouton, The Hague, 1968.

\bibitem[Stump(2016)]{Stump:2016}
Gregory Stump.
\newblock \emph{{I}nflectional Paradigms, {C}ontent and {F}orm at the
  {S}yntax-{M}orphology {I}nterface}.
\newblock Cambridge University Press, Cambridge, 2016.

\bibitem[Unbegaun(1957)]{Unbegaun}
Boris~O.\ Unbegaun.
\newblock \emph{{R}ussian {G}rammar}.
\newblock Oxford University Press, Oxford, 1957.

\bibitem[Vinogradov et~al.(1952)Vinogradov, Istrina, and
  Barxudarov]{Vinogradov}
V.~V. Vinogradov, E.~S. Istrina, and S.~G. Barxudarov, editors.
\newblock \emph{{G}rammatika russkogo jazyka, vol.\ I: {F}onetika i
  morfologija}.
\newblock ANSSR, Moscow, 1952.

\end{thebibliography}

\end{document}